\documentclass[a4paper,fleqn]{cas-dc}

\usepackage[authoryear]{natbib}
\usepackage{amsmath,amssymb}
\usepackage{graphicx}
\usepackage{booktabs}
\usepackage{array}
\usepackage{url}
\usepackage{xcolor}
\usepackage{tikz}
\usetikzlibrary{positioning,arrows.meta,fit,backgrounds}
\newcolumntype{P}[1]{>{\raggedright\arraybackslash}p{#1}}

\begin{document}
\let\WriteBookmarks\relax
\def\floatpagepagefraction{1}
\def\textpagefraction{.001}

\shorttitle{Protocol Before Progress: Leakage-Aware AIS Evaluation}
\shortauthors{Z.~Raisi and V.~Nazarzehi Had}

\title[mode = title]{Protocol before progress: Leakage-aware evaluation of AIS-based vessel trajectory prediction}
\tnotemark[1]
\tnotetext[1]{This is the accepted manuscript of the article published as: Z. Raisi, V.~M. Nazarzehi Had, Protocol before progress: Leakage-aware evaluation of AIS-based vessel trajectory prediction, \textit{Results in Engineering} 32 (2026) 114381, \url{https://doi.org/10.1016/j.rineng.2026.114381}. Please cite the published version.}

\author[1]{Zobeir Raisi}[orcid=0000-0002-1591-4492]
\cormark[1]
\ead{zobeir.raisi@cmu.ac.ir}
\credit{Conceptualization, Methodology, Software, Validation, Formal analysis, Investigation, Data curation, Writing -- original draft, Writing -- review \& editing, Visualization, Project administration}

\author[1]{Vali Mohammad Nazarzehi Had}[orcid=0000-0003-3261-6320]
\ead{v.nazarzehi@cmu.ac.ir}
\credit{Validation, Supervision, Writing -- review \& editing}

\affiliation[1]{organization={Marine Engineering Faculty, Chabahar Maritime University},
            addressline={Shahid Rigi Blvd.},
            city={Chabahar},
            postcode={99717778631},
            country={Iran}}

\cortext[1]{Corresponding author}

\begin{abstract}
Vessel trajectory prediction from Automatic Identification System (AIS) data supports collision avoidance and vessel traffic services, and new deep models routinely report lower error. How much of such a reduction comes from the model, the evaluation protocol or the problem formulation is rarely measured. We build a leakage-aware evaluation protocol that separates leakage (vessel-sharing, chronological and causal-history splits) from distribution shift (spatial and vessel-type hold-outs) and apply it to one month each of Danish and US Gulf Coast AIS. We evaluate TrAISformer (7.4\,M parameters), GATransformer (1.8\,M), AISFormer- and MGFormer-inspired reconstructions (8.3\,M and 0.71\,M), and three controls: TrAISformer with relative position tokens (7.6\,M), an LSTM on displacements (1.1\,M) and a compact encoder (0.43\,M). A best-of-16 oracle decoder lowers TrAISformer's error by a factor of 2.1--3.2, about three fifths of which a noise-perturbed constant-velocity predictor also obtains. Sharing vessels between training and test lowers the absolute-position models' one-hour error by 23--30\% (62--67\% at 57\,M parameters), but by at most 9\% at any size with relative position tokens. Region hold-outs raise the absolute models' error to 5--8 times that of constant velocity, while the relative variant stays within 1.2 times, and spatial blocks without unseen position bins point to location memorization. Errors measured with shared vessels or an oracle decoder therefore overstate the accuracy that collision-avoidance and traffic-monitoring systems can expect for new vessels or waters, and the coordinate frame belongs to the protocol. We release the splits, code and reporting recommendations.
\end{abstract}


\maketitle
\section{Introduction}\label{sec:intro}

The Automatic Identification System (AIS) periodically broadcasts a vessel's
position, speed and course, and these data underpin trajectory
prediction~\citep{capobianco2021rnn, nguyen2021traisformer, yu2025aisformer}
and long-range gap imputation~\citep{mgformer2026}, which serve collision
avoidance, vessel traffic services and port operations~\citep{riveiro2018maritime}.
Deep-learning methods, particularly transformers, now dominate both tasks, and
recent reviews document a steady stream of models reporting lower prediction
error~\citep{li2023shiptrajreview, xie2025survey, guo2026shiptrajreview}. When a
new model reports a lower error, how much of the improvement belongs to the
model? A reported number has at least three sources. The first is the
\emph{architecture}: an attention mechanism, a graph encoder, a positional
encoding. The second is the \emph{problem formulation}: what the model is given
and asked to produce, for example whether a position is an absolute cell of a
fixed region or an offset from the vessel's last report, and whether a
prediction is a single trajectory or the best of several samples. The third is
the \emph{evaluation protocol}: which windows the model trains and is tested
on. Comparisons in the literature usually attribute a gain to the first, while
the second and third change with it.

The protocol concern is not specific to maritime data. Splitting a dataset by
report or by track, rather than by vessel, lets the same ship's habits appear on
both sides of the split, so every number measured against it is
inflated~\citep{kaufman2011leakage, roberts2017blockcv}. A survey of
machine-learning-based science across seventeen fields traced overoptimistic
results in 294 papers to leakage of this kind~\citep{kapoor2023leakage}, and
audits that reimplement published systems under a common protocol have
repeatedly found reported gains shrinking or vanishing~\citep{dacrema2019progress,
henderson2018drl, melis2018evaluation, musgrave2020metric, oliver2018realistic};
practical checklists now exist~\citep{lones2024pitfalls}. To the best of our
knowledge, maritime deep learning has not had such an audit. A 2025 review of
transformer-based AIS modeling found no discussion of vessel-, time- or
region-level leakage in the literature it covered~\citep{xie2025survey}, and the
one public AIS benchmark built to standardize evaluation uses vessel-disjoint
splits but does not quantify how much other splits change reported
error~\citep{ma2026envship}. The decoding protocol compounds the problem:
probabilistic predictors are often reported with a best-of-$N$ oracle figure
alongside, or instead of, a single-shot error, without asking what the oracle is
worth to a predictor without a learned trajectory model.

Four terms are used throughout and defined formally in
Section~\ref{sec:framework}. A \emph{split manifest} is a released assignment
of every prediction window to training, validation or test. \emph{Leakage}
means that the training windows carry information about particular test windows
that the scenario the error is meant to describe (here, prediction for a vessel
the model was not trained on) would not provide; it lowers error.
\emph{Distribution shift} means that the test windows come from another area,
period or vessel type without such shared information; it measures transfer
and raises error. An \emph{oracle} decoder draws several predictions and scores
the one closest to the ground truth, which a deployed system cannot do; a
\emph{single-shot} error fixes one prediction per window before the truth is
seen.

A single dataset cannot settle whether such effects belong to the protocol or to
the data. We therefore run the same protocol, code and hyper-parameters on two
corpora with different geography and traffic: one month of Danish national AIS,
whose prediction windows are dominated by cargo traffic with substantial
fishing, tanker and ferry traffic, and one month of US Gulf Coast AIS from the
Houston--Galveston region~\citep{noaa_marinecadastre}, a working port and
waterway dominated by tug and towing traffic (Section~\ref{sec:data}). On both,
we evaluate four published approaches (TrAISformer, GATransformer, and
AISFormer- and MGFormer-inspired reconstructions) and three controls, two of
which change only how position is represented.

The protocol supplies much of a published number, and the problem formulation
decides how much. A best-of-16 oracle lowers TrAISformer's error by a factor of
2.1 to 3.2, and about three fifths of that factor, in log terms, also accrues
to a constant-velocity predictor with added noise. A split that shares vessels
lowers the one-hour error of the two absolute-position models by 23--30\%, but
that of TrAISformer with relative position tokens by only 9\% (Danish) and 3\%
(US); the gap grows with model size to 62--67\% for absolute tokens and stays
nearly flat for relative ones. A region hold-out raises the absolute models' error to
five to eight times that of constant velocity on both corpora, while the
relative variant stays near it, and held-out spatial blocks whose position bins
all occur in training show that the absolute models fail on unfamiliar
locations, not only on untrained bins. A chronological split, by contrast,
changes TrAISformer's normalized error by only 1--5\% on both corpora. Against these
effects, the architectural mechanisms the source papers emphasize are
second-order: GATransformer's graph attention gives no measurable benefit on
either corpus while its engineered waterway feature is worth 12--22\%, and
AISFormer's frequency-aware attention does not improve on TrAISformer.

Our contributions are as follows:
\begingroup\sloppy
\begin{enumerate}[1.]
\item A leakage-aware evaluation protocol for AIS trajectory prediction that
separates leakage (random-window, chronological with and without purging, and a
causal-history design) from distribution shift (longitude, latitude and
spatial-block hold-outs, composition matching, and a vessel-type hold-out),
applied to two corpora with different geography and traffic
(Section~\ref{sec:framework}).
\item A measurement of what a best-of-$N$ oracle decoder is worth, against the
model's own greedy decoding and against a noise-perturbed constant-velocity predictor,
for both absolute and relative position tokens (Section~\ref{sec:tfresult}).
\item Evidence that the coordinate frame is a protocol-level factor: the
vessel-sharing gain, its growth with model capacity, and the degradation under a
region hold-out all belong to absolute-position models, and controls that change
only the position representation remove most of them
(Sections~\ref{sec:leakage} and~\ref{sec:shift}).
\item An audit of four published approaches that separates what each mechanism
contributes from what the protocol supplies, including two leakage channels no
split manifest closes: the traffic around a held-out vessel, and auxiliary
structures fitted before splitting (Section~\ref{sec:components}).
\item Released split manifests for both corpora, preprocessing pipelines and all
model implementations~\citep{raisi2026code}, with a table of recommended
reporting practice (Section~\ref{sec:conclusion}).
\end{enumerate}
\endgroup

Fig.~\ref{fig:framework} summarizes the study. Section~\ref{sec:related} reviews
related work and Section~\ref{sec:data} describes the two corpora.
Section~\ref{sec:framework} defines the evaluation framework and
Section~\ref{sec:method} the models. Sections~\ref{sec:tfresult}
to~\ref{sec:shift} report, in turn, the decoding protocol, the component audits,
leakage and distribution shift. Sections~\ref{sec:discussion}
and~\ref{sec:conclusion} discuss and conclude.

\begin{figure*}
\centering
\begin{tikzpicture}[
  font=\footnotesize,
  box/.style={draw=black!60, rounded corners=2pt, fill=#1, align=center, inner sep=3pt, text width=2.75cm, minimum height=0.85cm},
  box/.default=black!3,
  wide/.style={box=#1, text width=3.8cm},
  wide/.default=black!3,
  narrow/.style={box=#1, text width=3.2cm},
  narrow/.default=black!3,
  res/.style={box=#1, text width=2.79cm, minimum height=0.5cm},
  res/.default=black!3,
  frame/.style={draw=black!45, dashed, rounded corners=4pt, inner sep=4pt},
  head/.style={font=\footnotesize\bfseries, anchor=base},
  arr/.style={-{Stealth[length=5pt]}, black!60, line width=0.7pt},
  node distance=0.2cm and 0.3cm]

\node[box, anchor=north] (dma) at (0,0) {Danish AIS\\(DMA, May 2026)};
\node[box, below=of dma] (noaa) {US AIS\\(NOAA, June 2023)};
\node[box, below=0.35cm of noaa] (prep) {Filter, split at 30-min gaps, remove jitter, resample to 10\,min};
\node[box, below=0.35cm of prep] (pool) {Window pool\\18 context +\\18 target steps};

\node[wide, anchor=north] (ref) at (4.12,0) {\textbf{Reference}\\vessel-disjoint: no test vessel in training};
\node[wide=orange!12, below=of ref] (leak) {\textbf{Leakage tests}\\\emph{relax the reference}\\random window, chronological (purged), causal history};
\node[wide=green!10, below=of leak] (shift) {\textbf{Shift tests}\\\emph{withhold area or fleet}\\longitude, latitude, spatial blocks, composition-matched, vessel type (Danish)};

\node[wide, anchor=north] (pub) at (8.9,0) {\textbf{Published comparators}\\TrAISformer, GATransformer,\\\mbox{AISFormer-inspired},\\\mbox{MGFormer-inspired} imputer};
\node[wide, below=of pub] (ctl) {\textbf{Controls}\\TrAISformer with relative tokens, LSTM on displacements, compact encoder};
\node[wide, below=of ctl] (base) {\textbf{Baselines}\\constant velocity (CV), noise-perturbed CV, linear interpolation};

\node[narrow=blue!8, anchor=north] (dec) at (13.39,0) {\textbf{Decoding}\\single-shot; \mbox{best-of-16} oracle for token models and noise-perturbed CV};
\node[narrow, below=of dec] (pred) {\textbf{Prediction}\\error at 1, 2 and 3\,h;\\ratio to CV};
\node[narrow, below=of pred] (imp) {\textbf{Imputation}\\masked-step error;\\ratio to interpolation};

\begin{scope}[on background layer]
\node[frame, fit=(ref)(leak)(shift)] (gman) {};
\node[frame, fit=(pub)(ctl)(base)] (gmod) {};
\node[frame, fit=(dec)(pred)(imp)] (gsc) {};
\end{scope}
\node[fit=(pool)(gman)(gmod)(gsc), inner sep=0pt] (all) {};

\node[res=blue!8, anchor=north west] (r6) at ([yshift=-0.6cm]all.south -| ref.west) {\S\ref*{sec:tfresult} decoding protocol};
\node[res, right=0.3cm of r6] (r7) {\S\ref*{sec:components} component audits};
\node[res=orange!12, right=0.3cm of r7] (r8) {\S\ref*{sec:leakage} leakage};
\node[res=green!10, right=0.3cm of r8] (r9) {\S\ref*{sec:shift} distribution shift};
\begin{scope}[on background layer]
\node[frame, fit=(r6)(r9)] (gres) {};
\end{scope}

\foreach \n/\t in {dma/Data, gman/Split manifests, gmod/Models, gsc/Scoring}
  \node[head] at (\n.north |- 0,0.42) {\t};
\node[head, anchor=east] at ([xshift=-0.3cm]gres.west) {Results};

\draw[arr] (dma.west) -- ++(-0.2,0) |- (prep.west);
\draw[black!60, line width=0.7pt] (noaa.west) -- ++(-0.2,0);
\draw[arr] (prep) -- (pool);
\draw[arr] (pool.east) -- (pool.east -| gman.west);
\draw[arr] (gman.east |- gmod.center) -- (gmod.west |- gmod.center);
\draw[arr] (gmod.east |- gsc.center) -- (gsc.west |- gsc.center);
\draw[arr] (gsc.south) -- (gsc.south |- gres.north);
\end{tikzpicture}
\caption{Study workflow. Two AIS corpora are preprocessed into one window pool each. Split manifests cut each pool into training, validation and test windows: the vessel-disjoint reference, leakage tests that relax it, and distribution-shift tests that withhold part of the area or of the fleet. Each learned model is trained from scratch on the training windows of every manifest it is evaluated on. All models are scored single-shot, and the token models and the noise-perturbed constant-velocity (CV) control also under the best-of-16 oracle. Prediction errors are normalized by constant velocity and imputation errors by linear interpolation on the same test windows. Arrows join stages; boxes within one dashed frame are parallel, and colors link the manifests and the decoding protocol to the sections that report them.}
\label{fig:framework}
\end{figure*}

\section{Related Work}\label{sec:related}

This section reviews AIS trajectory prediction (Section~\ref{sec:rel-baselines}),
the AIS spatio-temporal and imputation models that share its mechanisms
(Section~\ref{sec:rel-timeseries}), the evaluation methodology of other fields
on which the protocol draws (Section~\ref{sec:rel-sampled}), and the benchmark
closest to this work (Section~\ref{sec:rel-bench}).

\subsection{AIS trajectory prediction}\label{sec:rel-baselines}

Recent reviews trace vessel trajectory prediction from kinematic and
machine-learning models to recurrent, convolutional, transformer and graph
networks~\citep{li2023shiptrajreview, xie2025survey, guo2026shiptrajreview}.
Recurrent encoder--decoders~\citep{capobianco2021rnn} established the
sequence-to-sequence formulation. TrAISformer~\citep{nguyen2021traisformer}, a
widely used transformer baseline, represents latitude, longitude, SOG and COG as
discretized tokens of a fixed region and predicts by sampling; its published
results use a chronological split in which the same vessels can appear in
training and test, and report, for each test trajectory, the best of 16 sampled
predictions. AISFormer~\citep{yu2025aisformer} keeps the same four-hot
representation and classification framing and changes the attention to a
frequency-aware design in the manner of FEDformer~\citep{zhou2022fedformer}, so
the pair isolates one mechanism. Graph-based models widen the input to the
surrounding traffic: GATransformer~\citep{yuan2025gatransformer} applies graph
attention over nearby vessels and adds an engineered distance to waterway
intersections, and STMGCN~\citep{liu2022stmgcntraj} builds several interaction
graphs from social force, time to closest point of approach and vessel size.
A further line pursues predictive uncertainty through diffusion and stochastic
differential equations~\citep{li2024diffutraj, fang2026sde, jin2025stgdpm}; for
such generative predictors a best-of-$N$ figure is the natural, and often only,
thing reported, which is where the decoding question of
Section~\ref{sec:tfresult} matters most. Most recently, large language models
have been adapted to the task: AIS-LLM~\citep{park2025aisllm} aligns a time-series
encoder with a frozen language model for prediction, anomaly detection and
collision risk, and ShipTraj-R1~\citep{zhan2026shiptrajr1} casts prediction as
text generation and fine-tunes the model with group relative policy
optimization. A movement-state segmentation method~\citep{guo2024segmentation}
decomposes tracks into stop, transit and maneuver phases as preprocessing.

Across these works the evaluation protocol varies and is often only partly
stated. Of the four systems we audit, only MGFormer states a vessel-level
partition (all trajectories of one vessel go to one subset); the others describe
chronological or random divisions of tracks, and none reports how its error
changes under another split. The problem formulation varies as well: the
token models read absolute cells of a fixed region, AIS-LLM reads min-max
normalized absolute coordinates, and other models read offsets relative to the
vessel. These
differences are rarely isolated from the architecture they come with, which
Sections~\ref{sec:leakage} and~\ref{sec:shift} do.

\subsection{AIS spatio-temporal modeling and gap imputation}\label{sec:rel-timeseries}

AIS data also feed models of traffic rather than single tracks. Maritime
traffic networks extracted from AIS waypoints support flow prediction with
multigraph convolution~\citep{liang2022stmgcnflow} and the estimation of
link-level travel-time distributions on a directed
network~\citep{liang2024traveltime}, and trajectory features alone suffice to
classify vessel type~\citep{liang2021mvffnet}, which is why a vessel-type
hold-out (Section~\ref{sec:shift-type}) tests a shift in motion and not only in
a label. The fitted networks are the same kind of fitted
structure that MGFormer~\citep{mgformer2026} uses for long-gap imputation: a
waypoint graph learned from trajectories, with a gated graph network and a
bidirectional transformer. MGFormer descends from general time-series
imputation, from recurrent decay models~\citep{che2018grud} and bidirectional
recurrent imputation~\citep{cao2018brits} to self-attention
imputation~\citep{du2023saits} and score-based diffusion~\citep{tashiro2021csdi}.
On the forecasting side, patch-based~\citep{nie2023patchtst} and
frequency-decomposed~\citep{zhou2022fedformer} transformers dominate recent
benchmarks, yet a simple linear baseline has been shown to beat many of
them~\citep{zeng2023dlinear}, the closest precedent for the audit we run. A
structure fitted from trajectories has its own training data, which a split
manifest does not govern; Section~\ref{sec:imputeresult} measures that
channel.

\subsection{Evaluation methodology outside AIS}\label{sec:rel-sampled}

Evaluation methodology has received little direct attention in AIS work, so
the protocol draws on fields where it has been studied. Leakage through
non-independent splits is documented across machine-learning-based
science~\citep{kaufman2011leakage, kapoor2023leakage}, blocked and chronological
hold-outs are the standard remedy for autocorrelated and non-stationary
data~\citep{roberts2017blockcv, cerqueira2020evaluating}, and checklists now
codify the pitfalls~\citep{lones2024pitfalls}. Audits that reimplement
published systems under one protocol have found reported gains shrinking or
vanishing in recommender systems~\citep{dacrema2019progress}, reinforcement
learning~\citep{henderson2018drl}, language modeling~\citep{melis2018evaluation},
metric learning~\citep{musgrave2020metric} and semi-supervised
learning~\citep{oliver2018realistic}, and seed variance can exceed the
differences a benchmark reports~\citep{bouthillier2021variance}. Best-of-$N$
scoring comes from pedestrian and driving forecasting, where a generative
predictor draws $N$ futures and is scored on the nearest to the ground truth:
Social GAN~\citep{gupta2018socialgan} reports it as a variety loss,
Trajectron++~\citep{salzmann2020trajectron} for multimodal forecasts, and
Argoverse~\citep{chang2019argoverse} standardized it as minADE and minFDE. It
measures coverage legitimately, and its side effects on the learned distribution
have been analyzed~\citep{thiede2019variety}, but those studies do not ask what
the oracle is worth against a predictor with no learned dynamics, the control
Section~\ref{sec:tfresult} supplies for vessels.

\subsection{Benchmarks and positioning}\label{sec:rel-bench}\label{sec:rel-summary}

EnvShip-Bench~\citep{ma2026envship} is the closest work: a public prediction
benchmark built from four regional AIS sources (the Danish Maritime Authority,
NOAA MarineCadastre, the Piraeus dataset and Norway's Kystverket), with released
splits and preprocessing. It resamples all trajectories to a fixed 20-second
grid and expresses them in vessel-centric local coordinates. Its splits are
vessel-disjoint, assigned by a hash of the Maritime Mobile Service Identity
(MMSI), and leave-one-source-out
experiments test transfer by holding out a whole source rather than a region
within one dataset. It standardizes evaluation; it does not report how much a
different split, a decoding oracle or the coordinate frame changes the reported
error. The general reproducibility audits above, in turn, measure such effects
but not on AIS data and not for its leakage channels. Table~\ref{tbl:positioning}
summarizes the difference. To our knowledge, no published study combines
vessel-, time- and region-level manifests within one AIS corpus with a
re-evaluation of published systems under them.

\begin{table}
\caption{Position of this study against the closest benchmark and the general reproducibility audits. ``n.r.'': not reported.}\label{tbl:positioning}
\centering\footnotesize\setlength{\tabcolsep}{3pt}
\begin{tabular}{@{}P{0.35\columnwidth}P{0.19\columnwidth}P{0.18\columnwidth}P{0.19\columnwidth}@{}}
\toprule
 & EnvShip-Bench & Repro.\ audits$^{a}$ & This study \\
\midrule
AIS data & yes & no & yes \\
Released splits & yes & varies & yes \\
Splits within one corpus & vessel & --- & vessel, time, region \\
Leakage measured (permissive vs strict split) & n.r. & some & yes \\
Distribution shift measured & across sources & --- & area, vessel type \\
Oracle decoding against a null predictor & n.r. & no & yes \\
Coordinate frame isolated & n.r. & no & yes \\
Leakage outside the split (neighbors, fitted structures) & n.r. & no & yes \\
Published systems re-evaluated & yes & yes & yes \\
\bottomrule
\end{tabular}
\par\smallskip\footnotesize\raggedright $^{a}$\,E.g.\ \citet{dacrema2019progress, musgrave2020metric, oliver2018realistic}.
\end{table}

\section{Data}\label{sec:data}

We use two corpora with the same spatial extent but different geography and
traffic, and run the identical pipeline, windowing and hyper-parameters on both
(not re-tuned for the second). Table~\ref{tbl:corpora} compares them.

\begin{table}
\caption{The two corpora. Both prediction regions span $2.5^{\circ}\times2.7^{\circ}$; preprocessing, windowing and hyper-parameters are identical. Type shares are over the vessel-disjoint prediction windows with a known vessel type.}\label{tbl:corpora}
\centering\footnotesize\setlength{\tabcolsep}{3pt}
\begin{tabular}{@{}P{0.34\columnwidth}P{0.3\columnwidth}P{0.3\columnwidth}@{}}
\toprule
 & Danish (DMA) & US (NOAA) \\
\midrule
Source & Danish Maritime Authority & NOAA MarineCadastre \\
Period & May 2026 (31 days) & June 2023 (30 days) \\
Prediction region & 55.5--58.0$^{\circ}$N, 10.3--13.0$^{\circ}$E & 28.0--30.5$^{\circ}$N, 96.0--93.3$^{\circ}$W \\
Raw reports & 652.4\,M (national feed) & 265.1\,M (cropped region) \\
Reports after preprocessing & 180.0\,M in 396,662 tracks & 8.1\,M in 101,662 tracks \\
Median report interval & 10\,s & 70\,s \\
Vessels in prediction windows & 3,146 & 2,021 \\
Vessel-disjoint windows (train/val/test) & 46,168 / 3,887 / 3,528 & 30,207 / 2,408 / 3,147 \\
Largest vessel types (windows) & Cargo 37\%, Fishing 18\%, Tanker 17\%, Passenger 16\% & Tug 69\%, Tanker 12\%, Fishing 6\%, Cargo 6\% \\
Windows with known type & 94.7\% & 81.5\% \\
\bottomrule
\end{tabular}
\end{table}

\begin{figure*}
\centering
\includegraphics[width=\textwidth]{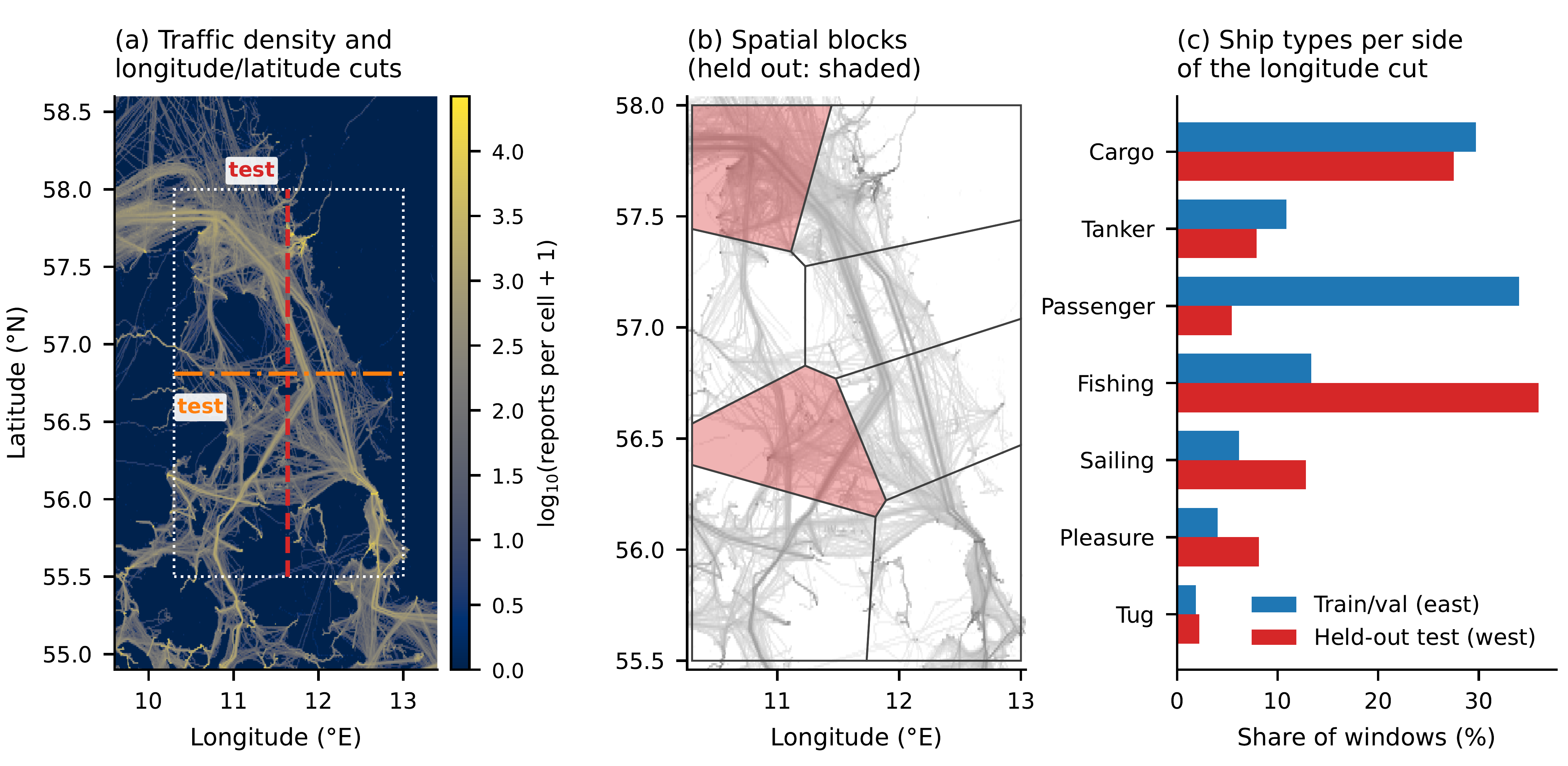}
\caption{Danish study area, maps at true aspect. (a) AIS traffic density
(log-scaled report counts per $0.01^{\circ}$ cell, the size of TrAISformer's
position bins, over four sample days spread across the month). The dotted box is
the 10.3--13.0$^{\circ}$E, 55.5--58.0$^{\circ}$N prediction region; the dashed
line is the longitude cut of the region manifests (median anchor longitude,
11.64$^{\circ}$E) and the dash-dotted line the latitude cut (56.81$^{\circ}$N),
each labeled on its held-out side. Coastlines and traffic separation schemes are
traced by the reports themselves; no basemap is overlaid. (b) The eight
$k$-means blocks of the spatial-block manifest; shaded blocks are held out.
(c) Ship-type composition of the longitude-cut windows on either side of the
cut. Types are each vessel's modal static AIS type over the month; one of the
seven classes is known for 94.8\% of the windows.}
\label{fig:study-area}
\end{figure*}

\textbf{Danish corpus.} The Danish Maritime Authority (DMA) publishes its
complete national AIS feed as daily archives~\citep{dma2026ais}; we use all 31
days of May 2026, 652.4~M raw position reports. The pipeline (i) restricts to
Class A/B vessels with a valid MMSI and coordinates, (ii) segments reports into
tracks at gaps longer than 30 minutes, and (iii) removes stationary-anchor GPS
jitter, which otherwise dominates raw counts with near-duplicate positions from
vessels at berth. It leaves 180.0~M reports in 396,662 tracks, with a native
median interval of 10~s between reports. The feed covers Danish national waters
and their approaches: the North Sea off the Jutland west coast, the Skagerrak and
Kattegat, the three Danish straits (Little Belt, Great Belt and the {\O}resund),
and the western Baltic. Deep-water routes carry
dense, regular transit traffic, while the straits compress it into narrow
channels crossed by ferry services, so a vessel's next hour is shaped strongly
by the channel it occupies. The prediction region, 55.5--58.0$^{\circ}$N and
10.3--13.0$^{\circ}$E (Fig.~\ref{fig:study-area}a), is the one TrAISformer was published on; its windows
are dominated by cargo traffic, with substantial fishing, tanker and ferry
traffic (Table~\ref{tbl:corpora}).

\begin{figure*}
\centering
\includegraphics[width=\textwidth]{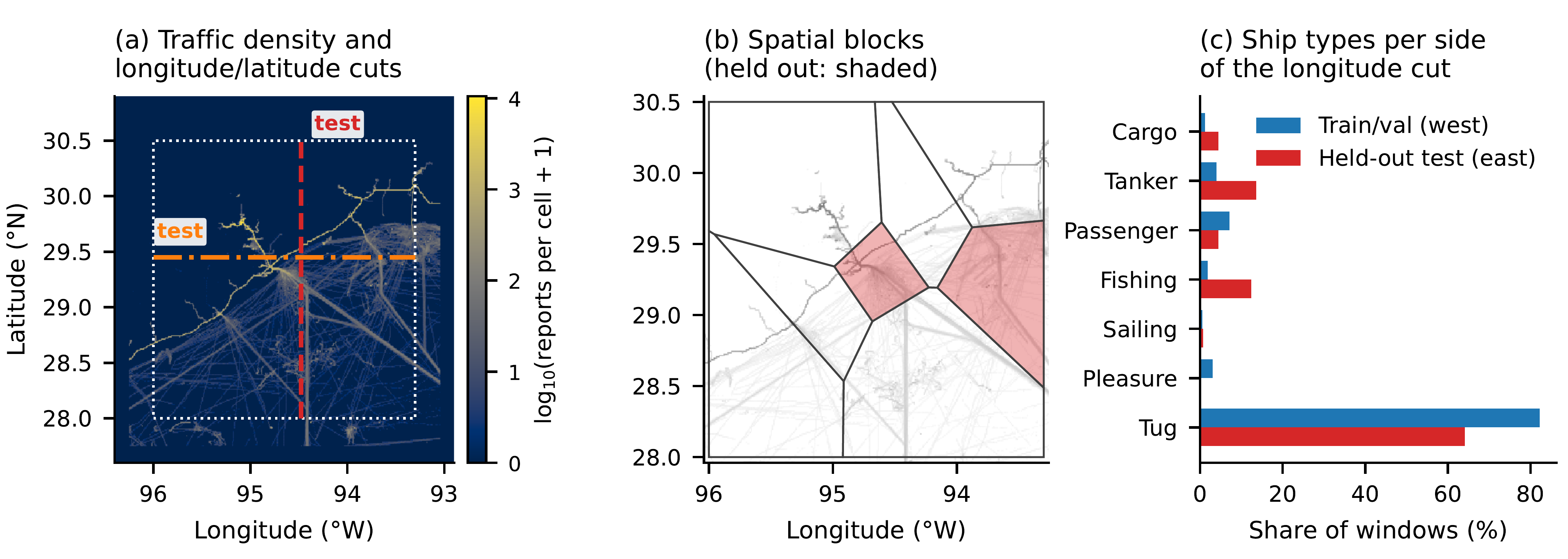}
\caption{US study area, same construction as Fig.~\ref{fig:study-area}.
(a) AIS traffic density over Houston--Galveston. The dotted box is the
28.0--30.5$^{\circ}$N, 96.0--93.3$^{\circ}$W prediction region, the dashed line
the longitude cut at the median anchor longitude (94.48$^{\circ}$W) and the
dash-dotted line the latitude cut (29.45$^{\circ}$N); unlike the Danish cuts,
the eastern and northern sides are held out here. (b) Spatial blocks; shaded
blocks are held out. (c) Ship-type composition on either side of the longitude
cut; one of the seven classes is known for 80.7\% of the windows. Tug and towing
traffic dominates both sides (82.2\% train/val, 64.1\% test), but the held-out
side carries a far larger share of Tanker (13.6\% versus 4.0\%) and Fishing
(12.4\% versus 1.9\%) traffic, so this manifest also tests generalization
across a genuine composition shift, in a different direction from the Danish
one (there, the held-out side has \emph{less} Passenger and \emph{more} Fishing
traffic than train/val; here, it has more Cargo, Tanker and Fishing traffic and
less Tug, Passenger and Pleasure traffic).}
\label{fig:study-area-noaa}
\end{figure*}

\textbf{US corpus.} NOAA's MarineCadastre archive~\citep{noaa_marinecadastre}
provides US AIS broadcast points; we use June 2023 cropped to the
Houston--Galveston region (28.0--30.5$^{\circ}$N, 96.0--93.3$^{\circ}$W), the
same $2.5^{\circ}\times2.7^{\circ}$ extent as the Danish prediction region, so
that both corpora present the same input geometry. The same pipeline reduces
265.1~M raw reports to 8.1~M reports in 101,662 tracks. Houston--Galveston is a
working port and waterway rather than a passenger corridor: tug and towing
traffic on the Gulf Intracoastal Waterway and tanker traffic to the Houston Ship
Channel terminals dominate, and passenger traffic is rare
(Fig.~\ref{fig:study-area-noaa}).

\begin{figure}
\centering
\includegraphics[width=\columnwidth]{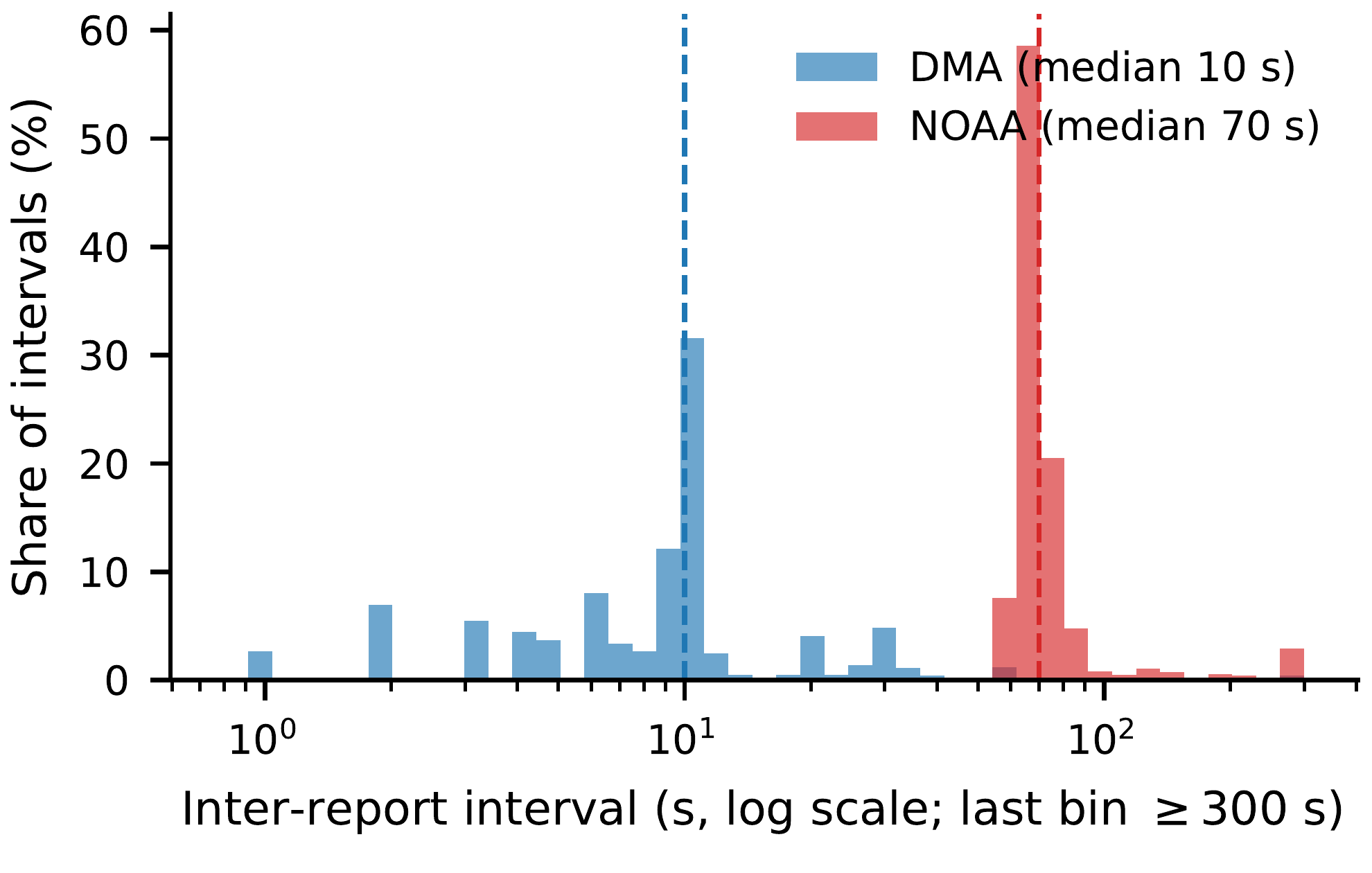}
\caption{Inter-report intervals on one sample day of each corpus (the first day
of its month), as the share of intervals in each log-spaced bin. The DMA feed's
native reporting interval (median 10~s) is an order of magnitude finer than NOAA
MarineCadastre's (median 70~s on the Houston--Galveston region), which enforces
a minimum reporting interval.}
\label{fig:interval-histogram}
\end{figure}

Every other manifest of Section~\ref{sec:splits} is built with the same rules
on both corpora. The held-out sides of the region manifests differ in traffic
composition, not only in area (panels c of Figs.~\ref{fig:study-area}
and~\ref{fig:study-area-noaa}): under the Danish longitude cut, the held-out
western side carries about one-sixth the training side's share of Passenger
windows (5.4\% against 34.0\%) and more than twice its share of Fishing windows
(35.9\% against 13.3\%), while on the US corpus the held-out eastern side carries
more Cargo, Tanker and Fishing traffic and less Tug traffic. Section~\ref{sec:shift} measures how
much of the region effect this composition shift explains.

\textbf{Reporting interval.} NOAA's native reporting interval is an order of
magnitude coarser than the DMA's (median 70~s against 10~s on one sample
day of each; Fig.~\ref{fig:interval-histogram}), consistent with the minimum interval
NOAA enforces. Both are much finer than the 10-minute grid every model here is
trained on (Section~\ref{sec:windowing}), so the difference does not reach the
models.

\textbf{Vessel-type labels.} Traffic-composition labels come from each vessel's
static AIS type field: text categories on DMA and numeric NAIS VesselType codes
on NOAA, both collapsed into the same seven classes (Cargo, Tanker, Passenger,
Fishing, Sailing, Pleasure, Tug) after removing rare and undefined categories,
with a per-vessel modal label where a track's reported type is inconsistent. On
NOAA, codes 31 and 32 (``Towing'') are merged into Tug, the same functional role
as code 52 in this region's traffic. The labels are used only to describe and
reweight the traffic (Section~\ref{sec:shift}), never as model input.

The main experiments are run on both corpora. Five smaller analyses are Danish
only and are marked where they appear: the vessel-type hold-out, the TrAISformer
decoder and loss ablation, and three GATransformer analyses (the
temporal-encoder and iterative-decoding ablations, the neighbor-provenance arms
and the neighborhood sweep). Each corpus covers one month;
Section~\ref{sec:disc-limits} discusses the limits this sets.

\section{Evaluation framework}\label{sec:framework}

This section defines the quantities that every later section measures: the
windows and tokens the models read (Section~\ref{sec:windowing}), the split
manifests that decide which windows a model trains and is tested on
(Section~\ref{sec:splits}), the decoding protocols
(Section~\ref{sec:decoding}), the metrics (Section~\ref{sec:metrics}), and the
rule for structures fitted from data before training
(Section~\ref{sec:fitting}).

\subsection{Windows, tokens and notation}\label{sec:windowing}

An AIS \emph{report} $r=(t,\varphi,\lambda,v,\psi)$ gives the timestamp,
latitude, longitude, speed over ground (SOG) and course over ground (COG)
broadcast by one vessel, which is identified by its MMSI. The reports of a vessel are segmented into \emph{voyages} at
gaps longer than 30~minutes. Within the prediction region of each corpus
(55.5--58.0$^{\circ}$N, 10.3--13.0$^{\circ}$E for the Danish corpus;
28.0--30.5$^{\circ}$N, 96.0--93.3$^{\circ}$W for the US corpus), voyages with
at least 20 reports are kept and resampled onto a regular grid of step
$\Delta=10$~minutes, giving the sequence $x_k=(\varphi_k,\lambda_k,v_k,\psi_k)$.
This is the dataset specification of TrAISformer~\citep{nguyen2021traisformer}.
We build the windows to that specification, instead of adapting TrAISformer to
other windows, so that its representation and decoder run as designed, and
every model in this paper reads the same windows.

A \emph{window} anchored at grid step $a$ consists of a context and a target,
\begin{equation}
X_a=(x_{a-L+1},\dots,x_a),\qquad Y_a=(x_{a+1},\dots,x_{a+H}),
\label{eq:window}
\end{equation}
with $L=H=18$ steps (3~h each). The last context step $x_a$ is the window's
\emph{anchor}, and its position $p_a=(\varphi_a,\lambda_a)$ is the origin of
every relative representation below. A window therefore spans
$(L+H-1)\Delta\approx 6$~h of one voyage. Training windows are drawn every
$s_{\mathrm{tr}}=6$ steps (1~h), so neighboring training windows of a voyage
overlap; validation and test windows are drawn every $s_{\mathrm{ev}}=18$ steps
(3~h), so no two evaluated contexts share a report, and neither do two evaluated
targets. The target of one evaluated window does, however, coincide with the
context of the next window of the same voyage. This overlap lies within one
subset and is not leakage, but it means that evaluated windows are not
independent samples, which is one reason why uncertainty is reported over
training seeds rather than over windows (Section~\ref{sec:metrics}). A model predicts the
positions $\hat p_{a+h}$ at horizons $h\in\{6,12,18\}$ steps (1, 2 and 3~h).
The \emph{window pool} $\mathcal{W}$ of a corpus is the set of all windows of
all kept voyages.

The TrAISformer-family models read each step as a \emph{four-hot token}, the
concatenation of the one-hot codes of its binned attributes,
\begin{equation}
e(x_k)=\big[\mathbf{1}_{b_\varphi(\varphi_k)},\,\mathbf{1}_{b_\lambda(\lambda_k)},\,
\mathbf{1}_{b_v(v_k)},\,\mathbf{1}_{b_\psi(\psi_k)}\big].
\label{eq:fourhot}
\end{equation}
In the published, \emph{absolute} representation, $b_\varphi$ and $b_\lambda$
index $0.01^{\circ}$ cells of the fixed region (250 latitude and 270 longitude
bins), $b_v$ 1-knot bins (30) and $b_\psi$ $5^{\circ}$ bins (72). The position
codes therefore state \emph{where} a vessel is. The \emph{relative}
representation of Section~\ref{sec:relcontrols} replaces them by bins of the
east and north offset $u_k$ of $p_k$ from the anchor $p_a$, in 0.8~km cells over
$\pm180$~km (450 bins per axis), which state its displacement from the anchor
but not its location.
The regression models (GATransformer, the LSTM and the compact encoder of
Section~\ref{sec:method}) read continuous features instead of tokens.

\subsection{Split manifests: leakage and distribution shift}\label{sec:splits}

A \emph{split manifest} $\mathcal{M}=(\mathcal{I}_{\mathrm{tr}},
\mathcal{I}_{\mathrm{va}},\mathcal{I}_{\mathrm{te}})$ is a partition of a
subset of $\mathcal{W}$ into training, validation and test windows, released as
index files so that every experiment can be rebuilt exactly. All manifests of a
corpus are cut from the same window pool; as a check, rebuilding the
vessel-disjoint, random-window, chronological and longitude-cut manifests from
the pool reproduces them array-identically.

A split can make a reported error unrepresentative in two different ways, and
we keep them apart. \emph{Leakage} means that the training windows carry
information about particular test windows that the scenario the error is meant
to describe would not provide. The scenario here is prediction for a vessel the
model was not trained on, and the information is the same vessel's other
windows, including windows that share reports with a test window. Leakage
biases error downward. \emph{Distribution shift} means that the test windows
come from a different population than the training windows (another area, a
later period or another vessel type) without such shared information. It
measures transfer and usually raises error. Under this scenario, a manifest that
shares vessels does not estimate unseen-vessel performance, whether the vessels
share reports with test windows (random windows) or only their identity
(the purged chronological manifest); we call such manifests leakage tests in this
sense, not because sharing a vessel is improper in every deployment
(Section~\ref{sec:disc-leakage}). A single manifest can both leak and shift, so
Table~\ref{tbl:splits} lists, for each manifest, what it withholds from
training and what it still shares with it, and Table~\ref{tbl:splitcounts}
gives the sizes.

\begin{table*}
\caption{Split manifests and what each one measures. \emph{Withheld} is what the training windows never contain; \emph{shared} is what they still have in common with the test windows. Leakage manifests relax the vessel-disjoint reference and, for the paper's target of prediction for vessels absent from training, lower error when shared information helps; shift manifests withhold part of the area or of the fleet and measure transfer. Sizes are in Table~\ref{tbl:splitcounts}.}\label{tbl:splits}
\centering\small
\begin{tabular}{@{}P{0.17\textwidth}P{0.27\textwidth}P{0.16\textwidth}P{0.2\textwidth}P{0.1\textwidth}@{}}
\toprule
Manifest & Assignment rule & Withheld & Shared with training & Section \\
\midrule
\multicolumn{5}{@{}l}{\emph{Reference}} \\
Vessel-disjoint & Whole vessels (MMSI) to one subset & Test vessels & Area, period & all \\
\midrule
\multicolumn{5}{@{}l}{\emph{Leakage tests}} \\
Random window & Windows at random, 70/15/15 & --- & Vessels (97--98\%), overlapping reports & \ref{sec:vesselleak} \\
Chronological & Anchor time, earliest 70\% / next 15\% / last 15\% & Last 15\% of the period & Vessels (55--69\%) & \ref{sec:timeleak} \\
Chronological, purged & As chronological; windows overlapping the next subset dropped & Last 15\% of the period & Vessels; no report & \ref{sec:timeleak} \\
Causal history & Fixed test set; training windows of the test vessels swapped in at equal size & --- & Test vessels' earlier windows or other voyages, by design & \ref{sec:causal} \\
\midrule
\multicolumn{5}{@{}l}{\emph{Distribution-shift tests}} \\
Region, longitude cut & Median anchor longitude; smaller side is test; straddling windows dropped & One side of the area & Vessels (39--51\%) & \ref{sec:shift} \\
Region, latitude cut & Median anchor latitude; otherwise as above & One side of the area & Vessels (37--65\%) & \ref{sec:shift} \\
Region, spatial blocks & $k$-means blocks of anchors, held out to $\ge$15\% of anchors & Blocks of the area & Vessels (41--49\%), most position bins & \ref{sec:shift} \\
Composition-matched & Test windows of a spatial hold-out reweighted to the training side's vessel-type (or type $\times$ speed) mix & As the hold-out & As the hold-out & \ref{sec:shift} \\
Vessel-type hold-out (Danish) & Vessel-disjoint test windows of one type; training without it; size-matched control & One vessel type & Area, period & \ref{sec:shift} \\
\bottomrule
\end{tabular}
\end{table*}

The \emph{vessel-disjoint} manifest is the reference: whole vessels are
assigned to one subset, so no test vessel contributes a training window, while
the area and the period are shared. Unless stated otherwise, every number in
this paper is measured on it. Three manifests relax it and are read as leakage
tests (Section~\ref{sec:leakage}). The \emph{random-window} manifest assigns
windows to the subsets at random (70/15/15): 97--98\% of its test vessels also
appear in training, and its test windows overlap training windows of the same voyage. The
\emph{chronological} manifest gives the earliest 70\% of anchor times to
training, the next 15\% to validation and the last 15\% to test, the convention
TrAISformer's published results use; out-of-sample chronological hold-out is
the recommended practice for non-stationary series~\citep{cerqueira2020evaluating,
roberts2017blockcv}. Here 55--69\% of the test vessels also appear in training. No training
window comes within 4.3~days of a test window on either corpus, because the
whole validation block lies between them; only windows at a boundary between
adjacent subsets can share reports, by at most 5.7~h. The
\emph{purged chronological} manifest removes that overlap: any window whose span
overlaps a window of the next subset is dropped (657 Danish and 601 US windows),
which leaves at least 10~minutes between subsets and no shared report. Finally,
the \emph{causal-history} design holds one test set fixed and varies only
whether the training set contains other windows of the test vessels, at equal
training size (Section~\ref{sec:causal}).

Four manifests withhold a part of the area or of the fleet and are read as
distribution-shift tests (Section~\ref{sec:shift}). The \emph{longitude} and
\emph{latitude} region manifests cut the pool at the median anchor longitude
(11.64$^{\circ}$E Danish, 94.48$^{\circ}$W US) or latitude (56.81$^{\circ}$N,
29.45$^{\circ}$N). The larger side is split 85/15 into training and validation,
the smaller side is the test set, and every window with steps on both sides of
the cut is dropped, so no window straddles training and test waters. The
\emph{spatial-block} manifest clusters anchor positions into eight blocks with
$k$-means, so that blocks follow traffic density, and holds out randomly drawn
blocks until they contain at least 15\% of the anchors (23\% Danish, 31\% US).
Unlike a straight cut, a held-out block shares its latitude range with training
blocks at other longitudes and its longitude range with training blocks at other
latitudes (Fig.~\ref{fig:study-area}b), so most of its latitude and longitude
bins also occur in training (all of them on the Danish corpus); this separates an
unfamiliar location from an unseen bin. The \emph{composition-matched} evaluation reweights the
test windows of each spatial hold-out by the ratio of training to test share of
their vessel type, or of their vessel type and anchor-speed quartile, so that
the test set matches the training side in traffic mix and speed; model error and
constant-velocity error are weighted alike.
On the Danish corpus, the \emph{vessel-type hold-out} tests on the
vessel-disjoint test windows of one type (Fishing, Passenger or Tanker) and
trains without that type; its control trains on a random subset of the same size
that includes the type.

The region manifests are not vessel-disjoint: 37--65\% of their test vessels
also appear in training (Table~\ref{tbl:splitcounts}). A region-split error
therefore mixes a location shift with some vessel sharing, which is why
Section~\ref{sec:shift} reads it against both the vessel-disjoint and the
random-window reference.

\subsection{Decoding protocols and the oracle control}\label{sec:decoding}

Decoding deserves explicit treatment because it carries much of the published
result. The published protocol draws $N=16$ rollouts per test window and, at
each horizon, keeps the one closest to the ground truth,
\begin{equation}
d_h^{\mathrm{bo}N}=\frac{1}{|\mathcal{I}_{\mathrm{te}}|}\sum_{a\in\mathcal{I}_{\mathrm{te}}}
\min_{n\le N}\mathrm{hav}\big(\hat p^{(n)}_{a+h},p_{a+h}\big),
\label{eq:bestofn}
\end{equation}
where $\mathrm{hav}$ is the great-circle distance. We call this selection an
\emph{oracle} because it consults the answer: it rewards a predictor for being
diverse as well as for being right, and it is not comparable to any
\emph{single-shot} error, for which one prediction per window is fixed before
the truth is seen. We therefore report the same trained model under
best-of-16, under a single random draw and under greedy (arg-max) decoding. We add a deliberately trivial control, constant
velocity perturbed by Gaussian noise whose scale is fitted on validation,
sampled 16 times and given the same oracle selection, to measure what the
protocol grants a predictor without a learned trajectory model. Checkpoint selection is done
separately for each decoding protocol on validation, so no reported number is
depressed by a checkpoint chosen to favor a different decoder. Greedy decoding
picks the most probable next report under the model, within the vicinity mask,
not necessarily the most probable trajectory.
Appendix~\ref{app:r2-decoding} shows that beam search over the joint token
finds likelier trajectories but not closer ones, and that on the
vessel-disjoint manifest greedy trajectories are as kinematically consistent
as the quantized ground truth.

\subsection{Metrics, normalization and replication}\label{sec:metrics}

The prediction error at horizon $h$ is the single-shot counterpart of
Eq.~\eqref{eq:bestofn}, the mean great-circle distance $d_h$ between predicted
and true positions over the test windows, in kilometers; the one-hour error is
the headline figure. Gap imputation is scored over the masked steps and at the
middle of the gap (Section~\ref{sec:tfmethod}).

Manifests do not produce equally hard test sets, and neither do corpora: the
one-hour error of constant-velocity extrapolation, which carries the anchor's
SOG along its COG, ranges from 2.5 to 4.1~km across the Danish manifests and
from 3.3 to 7.5~km across the US ones.
Every error is therefore also reported as a \emph{constant-velocity ratio},
$\rho_h=d_h/d_h^{\mathrm{CV}}$, where $d_h$ and $d_h^{\mathrm{CV}}$ are the mean
errors of the model and of constant velocity over the same test windows (a ratio
of means, not a mean of per-window ratios). The ratio gives the manifests a common
reference; it does not remove differences in traffic composition or other task
characteristics between test sets, which Section~\ref{sec:shift-composition}
examines separately, and every ratio table lists the constant-velocity error in
kilometers so that its ratios can be converted back. The ratio is
available for every manifest of both corpora, and the leakage gaps of
Section~\ref{sec:leakage} are differences of ratios. For the region manifests
we also report the share of test-context position bins that never occur in a
training context.

Every arm is trained with three seeds (42, 43 and 44), and the GATransformer
ablations on the Danish corpus with six. We report the mean and the sample standard
deviation over seeds, and we compare two arms by the mean seed-paired
difference and its standard error rather than by the spread of either arm,
because seed-to-seed variance can exceed the differences a benchmark
reports~\citep{bouthillier2021variance}. The capacity trend of Section~\ref{sec:capacity} is the
least-squares slope of the per-seed greedy gap on the base-10 logarithm of the
parameter count, with a 95\% interval from 5,000 seed-bootstrap resamples and a
Spearman rank correlation. The greedy gap of one trained size and seed is
$\rho_1^{\mathrm{vessel}}-\rho_1^{\mathrm{random}}$, the difference of its
vessel-disjoint and random-window ratios at one hour, both trained with that seed;
percentages quote the same gap as a share of the vessel-disjoint ratio. The
resampling unit is one (size, seed) point, so a curve of five sizes with three
seeds each has 15 points, and the Spearman correlation uses these 15 points. The
absolute- and relative-token curves are resampled independently for their slope
difference. These intervals, like every standard deviation and standard error in
this paper, describe variation across training seeds on fixed test windows and
split manifests; they do not describe uncertainty across other months, regions
or spatial partitions.

\subsection{Fitting discipline}\label{sec:fitting}

No model in this paper is pretrained: every network, including the compact
encoder, is trained from scratch on the training windows of the manifest it is
evaluated on. The only structures fitted from data before training are the
waypoint graph of the imputation comparator ($k$-means centroids and transition
counts) and the waterway-intersection nodes of GATransformer. Both are refit for
every manifest, from that manifest's own training windows alone. Everything else
is fixed by constants rather than estimated: the absolute coordinate bins are
defined by the published region, the relative bins by a fixed cell size and
range, and the position and speed scales of the imputation models are fixed
lengths (20~km, 20~kn), so there is no normalization statistic to leak. What no
manifest can govern is the traffic context of a model that reads surrounding
vessels, which Section~\ref{sec:gat} measures instead.

\section{Models}\label{sec:method}

Table~\ref{tbl:models} lists every model, its parameter count and the role it
plays. Four are published approaches, rebuilt and evaluated under the protocol
of Section~\ref{sec:framework} (Section~\ref{sec:tfmethod}). Three are not
proposed methods but controls: two change only how position is represented
(Section~\ref{sec:relcontrols}), and a compact encoder gives a small-capacity
reference.

\begin{table*}
\caption{Models in this study. Parameter counts are those of the configurations reported throughout (selected on validation for the comparators). \emph{Position input} is how a model sees where the vessel is: absolute cells of the fixed region, cells of the offset from the window anchor, or continuous offsets.}\label{tbl:models}
\centering\small
\begin{tabular}{@{}P{0.17\textwidth}P{0.11\textwidth}P{0.14\textwidth}P{0.22\textwidth}rp{0.17\textwidth}@{}}
\toprule
Model & Task & Output & Position input & Params & Role \\
\midrule
TrAISformer & Prediction & Autoregressive tokens & Absolute cells & 7.4~M & Published comparator \\
AISFormer-inspired & Prediction & Autoregressive tokens & Absolute cells & 8.3~M & Published comparator (reconstruction) \\
GATransformer & Prediction & Direct regression & Kinematics, local-frame neighbors, waterway distances & 1.8~M & Published comparator \\
MGFormer-inspired & Gap imputation & Direct regression & Offsets + waypoint graph & 0.71~M & Published comparator (reconstruction) \\
\midrule
TrAISformer, relative tokens & Prediction & Autoregressive tokens & Anchor-offset cells & 7.6~M & Coordinate-frame control \\
LSTM on displacements & Prediction & Direct regression & Step displacements, SOG, COG & 1.1~M & Coordinate-frame control \\
Compact encoder & Prediction, imputation & Direct regression & Anchor offsets, SOG, COG & 0.43~M\textsuperscript{a} & Small-capacity reference \\
\bottomrule
\end{tabular}
\par\smallskip\footnotesize\textsuperscript{a}\,0.60~M with the masked-reconstruction head used for gap imputation.
\end{table*}

\subsection{Published comparators}\label{sec:tfmethod}

We rebuild four published approaches and evaluate them on the released split
manifests: TrAISformer~\citep{nguyen2021traisformer}, the standard transformer
baseline for AIS trajectory prediction; AISFormer~\citep{yu2025aisformer},
which keeps TrAISformer's classification framing but replaces its attention with
a frequency-aware design; GATransformer~\citep{yuan2025gatransformer}, which adds
graph attention over surrounding traffic and an engineered waterway feature; and
MGFormer~\citep{mgformer2026}, a long-range gap imputer, evaluated on the task it
was published for.

TrAISformer (from its released code) and GATransformer (from its full article)
are reimplementations. The AISFormer and MGFormer articles have no public code
and leave key details unspecified, so we build controlled reconstructions of the
mechanism each names, called \emph{AISFormer-inspired} and
\emph{MGFormer-inspired}, and restrict every conclusion about them to that
mechanism. Where the reconstruction is uncertain, components the source does not
specify are held identical to the comparator it is read against, so the named
mechanism is the only difference. Table~\ref{tbl:deviations} summarizes, per
system, what the source specifies, what is ours, and the evidence that the
rebuild is faithful; Appendix~\ref{app:comparators} gives the details. Every
reported number is a property of our implementation under our protocol, not the
original authors' result, and all four implementations are released with this
paper~\citep{raisi2026code}. Below, ``AISFormer'' and ``MGFormer'' denote the
reconstructions unless the source article is named.

\begin{table*}
\caption{Fidelity of the rebuilt comparators: what the source provides, what we follow, what is ours, and the evidence that the rebuild is faithful. All four implementations are released~\citep{raisi2026code}; Appendix~\ref{app:comparators} gives the details.}\label{tbl:deviations}
\centering\small
\begin{tabular}{@{}P{0.12\textwidth}P{0.09\textwidth}P{0.23\textwidth}P{0.23\textwidth}P{0.23\textwidth}@{}}
\toprule
System & Source & Followed & Ours / deviations & Fidelity evidence \\
\midrule
TrAISformer & Released code & Four-hot representation, causal transformer, multi-resolution loss, top-10 and 40-bin vicinity sampler & Configuration selected by a sweep (7.4~M; released default 57.4~M) & Published errors reproduced within 7\% under the source's split and decoder (0.86/1.75/2.84 vs 0.89/1.74/3.04~km); best-of-16 varies $\le$0.05~km across 7.4--57.4~M \\
AISFormer-inspired & Article only, no code & Four-hot representation, 8-layer 8-head backbone, multi-resolution loss, frequency-split attention & Causal prefix-spectrum attention (App.~\ref{app:spectral}); TrAISformer's smoothing term for the coarse-bin loss; deterministic mode set & Identical to TrAISformer except the attention; causality verified numerically; conclusions limited to the attention mechanism \\
GATransformer & Article only, no code & Architecture, positional encoding, attention equations, loss, decoder, 4-Fs and 102-Fs feature sets & Neighbors (8 nearest within 10~km); waterway extraction adapted to a 10-min grid; dropout 0.2 (source 0.5 unstable here) & Source's dominant feature reproduced (waterway distances); neighbor definition swept over a 4.7-fold occupancy range; six seeds \\
MGFormer-inspired & Article only, no code & Waypoint graph, gated graph network, bidirectional encoder, position and speed/course heads, motion consistency & $k$-means waypoints; vessel length and type for draught; nearest-waypoint embedding with a transfer context for masked steps & 38\% better than linear interpolation (source: 21\% over its baselines); the article's own specification keeps the graph's sign (App.~\ref{app:comparators}) \\
\bottomrule
\end{tabular}
\end{table*}

\textbf{TrAISformer and AISFormer.} TrAISformer reads the four-hot tokens of
Eq.~\eqref{eq:fourhot} with a causal 8-layer transformer and predicts by
autoregressive rollout. We follow the released code, including two details a
summary omits: a multi-resolution loss (two smoothing passes over each predicted
distribution) and a sampler restricted to the ten most likely bins and, for
position, to a 40-bin vicinity of the previous position. AISFormer holds the
representation, loss, decoder, trainer, schedule and evaluation code identical
and changes only the attention, to a causal frequency-decomposing analogue
(Appendix~\ref{app:spectral}), so the pair differs in one mechanism. Both are
swept over dropout, depth, width and learning rate and reported at their best
validation setting (7.4~M and 8.3~M parameters, respectively).

\textbf{GATransformer.} This comparator is a direct multi-step regressor. A
spatial encoder applies graph attention over the eight nearest vessels within
10~km; a temporal encoder reads the target's own sequence; and an engineered
input gives distances to waterway intersections. It emits the whole horizon at
once under mean squared error and is reported single-shot, because decoding it
autoregressively, or scoring it under the best-of-16 oracle, would misread it.
The neighbor set, the local-frame features, the waterway network and the
hyper-parameters are ours. The source's dropout of 0.5 and learning rate of
$10^{-3}$ were unstable on this dataset, so the selected configuration uses
dropout 0.2 and otherwise the source's settings (Appendix~\ref{app:comparators}).

\textbf{A neighborhood cannot be held out.} One consequence deserves separate
treatment, because it limits what any split manifest can promise rather than
being a property of this reconstruction. Vessel-disjointness is well defined
for a model that reads one trajectory: hold the vessel out, and nothing it did
is in training. It is not well defined for a model that reads the traffic around
that vessel. A held-out vessel still sails among the training fleet (73\% of the
neighbors of a Danish test window and 67\% of a US one belong to training
vessels), so its context is largely material the model has already seen,
however strictly the target is partitioned.

Measuring this requires a control, because the obvious experiment is
confounded. Restricting each window to neighbors from its own split is the
strict reading of disjointness, but it also thins the neighborhood, and
unequally: on the Danish corpus a training window retains 70\% of its
neighbors and a test window 13\%, since the training split holds most of the fleet. An arm restricted that
way would lose accuracy partly because its neighbors are unfamiliar and partly
because it has far fewer of them. We therefore run three arms that differ in the
neighbor set and in nothing else: \emph{all}, every vessel actually present,
which is what a deployed system observes; \emph{strict}, only vessels from the
window's own split; and \emph{density}, all traffic thinned at random to the
retention rate that \emph{strict} produces on the same split, matched to within
0.03 percentage points. The step from \emph{all} to \emph{density} is the cost
of a sparser neighborhood; the step from \emph{density} to \emph{strict} is what
remains once density is fixed, and only that step is evidence about leakage.

\textbf{MGFormer and the gap-imputation protocol.} MGFormer reconstructs a long
missing stretch of a track from the observations on \emph{both} sides of it.
Asking it to forecast would remove the context its architecture is built
around, so we evaluate it on the task it was published for, on the same windows
and split manifests as the prediction study. Each 36-step window loses one
contiguous block of 11 steps (30.6\% of the window, about 110~minutes), placed
in the interior with at least three observed steps on each side. The position of
the block is drawn once from a generator keyed by the window index, so every
system reconstructs identical gaps. Error is the mean great-circle distance over
the masked steps, overall and at the middle of the gap.

Our MGFormer-inspired comparator follows the four components the article names:
a waypoint graph (256 $k$-means clusters, 2,926 directed edges on the Danish
corpus), a gated graph network, a bidirectional transformer encoder with
position and speed/course heads, and a motion-consistency term. Three departures
are forced by the feeds or by the article's silence
(Table~\ref{tbl:deviations}). Three references share the protocol: the model
without its waypoint graph, the compact encoder with a masked-reconstruction
head, and linear interpolation across the gap, which uses no training data and
so cannot leak. A waypoint graph extracted from the whole dataset has seen the
routes of the held-out vessels, a leakage channel the source does not discuss.
We fit the graph on the training split by default and run the whole-dataset
alternative as an explicit arm, with model, training and split fixed.

\subsection{Coordinate-frame controls and the compact encoder}\label{sec:ownencoder}\label{sec:relcontrols}

The four comparators differ in architecture, but the two tokenized ones also
share a choice of problem formulation: their position tokens are absolute
cells of a fixed region. Many AIS models instead read positions relative to the
vessel, or only its speed and course; EnvShip-Bench~\citep{ma2026envship}, for example,
expresses trajectories in vessel-centric local coordinates. To separate what belongs to an architecture from what belongs to
this formulation, we add two controls that change the position representation
and nothing else that could be held fixed. They are not proposed methods.

\textbf{TrAISformer with relative position tokens.} The model, loss, decoder
(including the vicinity sampler), training schedule and hyper-parameters are
those of the selected TrAISformer configuration. Only the position codes of
Eq.~\eqref{eq:fourhot} change, from absolute $0.01^{\circ}$ cells to the east
and north offset from the window anchor in 0.8~km cells over $\pm180$~km,
measured in a local equirectangular projection at the anchor's latitude. An
offset beyond this range would be clamped to the edge bin, but none occurs in
the context or target of any manifest of either corpus (the largest is 175~km). The
cell size is chosen so that the two representations have the same resolution:
encoding the true future position and decoding it back gives a mean error of
0.32~km with relative tokens against 0.32--0.33~km (Danish) and 0.40--0.41~km (US) with
absolute ones,
and the 40-bin vicinity spans $\pm16$~km, between the absolute grid's
$\pm22$~km in latitude and about $\pm12$~km (Danish) to $\pm19$~km (US) in
longitude. The larger position
vocabulary adds 0.2~M parameters (7.6~M in total). A region the model never saw
uses the same tokens as one it did, so this control isolates the absolute-cell
assumption within one architecture.

\textbf{LSTM on displacements.} A two-layer LSTM (hidden size 256) reads, per
context step, the east and north displacement from the previous step, the SOG
and the sine and cosine of the COG, and a linear head emits the 18 future
positions as offsets from the anchor in one shot, trained with a smooth-$L_1$
loss (1.1~M parameters). Its input contains no absolute coordinate, so it is
translation invariant by construction; it is the recurrent displacement model
in its most common form.

\textbf{Compact encoder.} A transformer encoder of about 0.43~M parameters
(hidden dimension 128, three layers, rotary positional encoding over the report
index) reads, per context step, the east/north offset from the anchor, the time
to the anchor, the SOG and the sine and cosine of the COG, and regresses the
future offsets. It is trained from scratch on the same
windows and manifests as every other model. Its purpose is a known-capacity
reference rather than architectural novelty: a model much smaller than the
prediction comparators, against which a leakage or protocol effect shows what
the protocol alone buys, uninflated by capacity. Its positional encoding was
chosen in a separate ablation on the same split manifests, outside the scope of
this paper.

\subsection{Implementation and computational cost}\label{sec:impl}

All models are implemented in PyTorch. The three prediction comparators each
receive a tuning sweep over depth, width, dropout and learning rate: four
configurations from 7.4~M to 57.4~M parameters for TrAISformer, five for
AISFormer (adding a mode-budget arm) and six for GATransformer (adding a
graph-head arm). The selected configuration is then retrained under every
manifest. GATransformer also carries the cost of its co-present traffic tensor,
which is built once over the whole dataset, so its neighbors may belong to any
split (the channel measured in Section~\ref{sec:gat}), whereas its
waterway-intersection nodes are extracted from each manifest's training windows
alone. The full program, including the capacity sweeps for both position
representations, comprises 377 training runs in this revision on top of the
original study, about 43 GPU-hours of training on one NVIDIA GeForce RTX 4090
and one RTX 3090 (24~GB each); the longest single run trains in 85~minutes.

\section{Decoding protocol: best-of-$N$ against single-shot error}\label{sec:tfresult}\label{sec:results}\label{sec:noaaresult}

\begin{figure*}
\centering
\includegraphics[width=\textwidth]{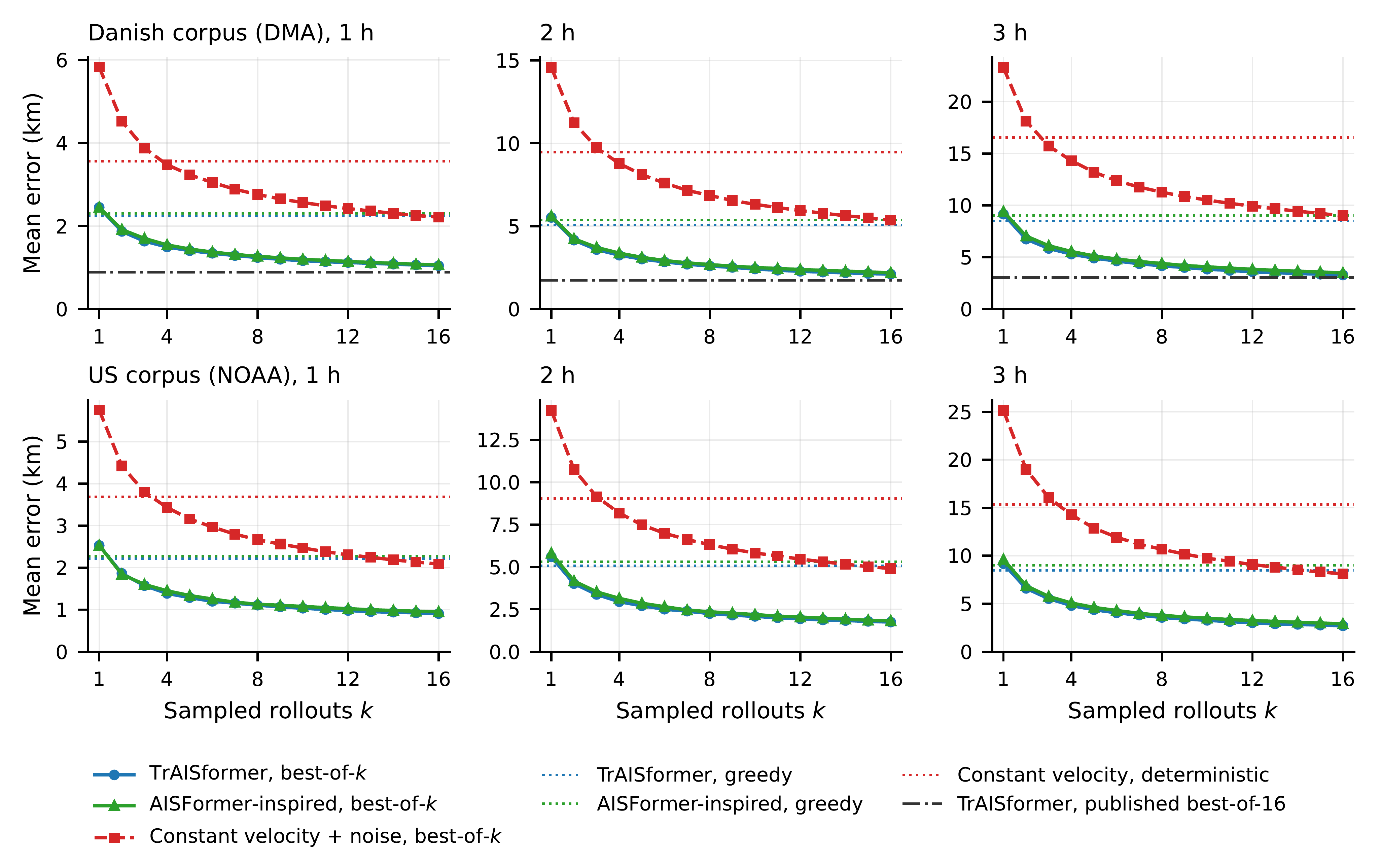}
\caption{Mean displacement error against the number of sampled rollouts $k$ under best-of-$k$ selection, on the vessel-disjoint test windows of the Danish corpus (top) and the US corpus (bottom), one training seed (42) per model. Both token models and the noise-perturbed constant-velocity control improve steeply with $k$, because the selection consults the ground truth; the control learns nothing from the trajectory, so its descent is due to the protocol alone. Dotted lines: greedy (token models) or deterministic (constant velocity) error; dash-dotted line (Danish only): the published best-of-16 value.}
\label{fig:bestofk}
\end{figure*}

This section measures what the decoding protocol contributes to a reported
error, before any question of splits. Every number is measured on the
vessel-disjoint manifest of each corpus, so that the oracle effect is not
mixed with leakage. Table~\ref{tbl:tfprotocol} reports all models at one, two
and three hours, and Fig.~\ref{fig:bestofk} shows how the error of the token
models falls as more samples are given to the oracle.

\begin{table*}
\caption{Mean displacement error (km) at one, two and three hours on the vessel-disjoint test windows of both corpora, under the best-of-16 oracle and under single-shot decoding of the same trained models. Best-of-16 keeps, per window and horizon, the sampled rollout closest to the ground truth. Means $\pm$ standard deviations over three seeds (six for Danish GATransformer). Lower is better.}\label{tbl:tfprotocol}
\centering\footnotesize\setlength{\tabcolsep}{4pt}
\begin{tabular}{@{}lrrr@{\hspace{1.5em}}rrr@{}}
\toprule
 & \multicolumn{3}{c}{Danish (DMA)} & \multicolumn{3}{c}{US (NOAA)} \\
\cmidrule(lr){2-4}\cmidrule(lr){5-7}
Model and decoding & 1\,h & 2\,h & 3\,h & 1\,h & 2\,h & 3\,h \\
\midrule
\multicolumn{7}{@{}l}{\emph{TrAISformer, absolute tokens}} \\
\quad as published$^{a}$ & 0.89 & 1.74 & 3.04 & --- & --- & --- \\
\quad best-of-16 (oracle) & 1.05 $\pm$ 0.01 & 2.14 $\pm$ 0.04 & 3.36 $\pm$ 0.11 & 0.89 $\pm$ 0.01 & 1.71 $\pm$ 0.04 & 2.66 $\pm$ 0.10 \\
\quad single random draw & 2.49 $\pm$ 0.05 & 5.62 $\pm$ 0.12 & 9.33 $\pm$ 0.21 & 2.50 $\pm$ 0.03 & 5.59 $\pm$ 0.02 & 9.17 $\pm$ 0.01 \\
\quad greedy & 2.23 $\pm$ 0.04 & 5.08 $\pm$ 0.05 & 8.46 $\pm$ 0.05 & 2.20 $\pm$ 0.02 & 5.07 $\pm$ 0.03 & 8.44 $\pm$ 0.03 \\
\multicolumn{7}{@{}l}{\emph{AISFormer-inspired, absolute tokens}} \\
\quad best-of-16 (oracle) & 1.06 $\pm$ 0.03 & 2.16 $\pm$ 0.07 & 3.39 $\pm$ 0.10 & 0.94 $\pm$ 0.01 & 1.80 $\pm$ 0.02 & 2.87 $\pm$ 0.04 \\
\quad single random draw & 2.44 $\pm$ 0.05 & 5.53 $\pm$ 0.17 & 9.22 $\pm$ 0.31 & 2.51 $\pm$ 0.06 & 5.72 $\pm$ 0.09 & 9.51 $\pm$ 0.11 \\
\quad greedy & 2.27 $\pm$ 0.03 & 5.37 $\pm$ 0.10 & 8.99 $\pm$ 0.10 & 2.25 $\pm$ 0.04 & 5.20 $\pm$ 0.10 & 8.80 $\pm$ 0.19 \\
\multicolumn{7}{@{}l}{\emph{TrAISformer, relative tokens}} \\
\quad best-of-16 (oracle) & 1.15 $\pm$ 0.02 & 2.71 $\pm$ 0.06 & 4.40 $\pm$ 0.06 & 0.95 $\pm$ 0.01 & 2.16 $\pm$ 0.03 & 3.61 $\pm$ 0.05 \\
\quad single random draw & 2.74 $\pm$ 0.02 & 6.79 $\pm$ 0.09 & 11.27 $\pm$ 0.23 & 2.34 $\pm$ 0.00 & 5.62 $\pm$ 0.03 & 9.53 $\pm$ 0.02 \\
\quad greedy & 2.64 $\pm$ 0.04 & 7.10 $\pm$ 0.11 & 11.98 $\pm$ 0.17 & 2.20 $\pm$ 0.02 & 5.53 $\pm$ 0.08 & 9.61 $\pm$ 0.13 \\
\multicolumn{7}{@{}l}{\emph{Single-shot regressors}} \\
\quad GATransformer & 2.33 $\pm$ 0.09 & 4.72 $\pm$ 0.12 & 7.82 $\pm$ 0.15 & 2.43 $\pm$ 0.03 & 4.99 $\pm$ 0.16 & 8.04 $\pm$ 0.20 \\
\quad LSTM on displacements & 2.43 $\pm$ 0.02 & 5.12 $\pm$ 0.05 & 8.51 $\pm$ 0.03 & 2.16 $\pm$ 0.02 & 4.60 $\pm$ 0.03 & 7.76 $\pm$ 0.04 \\
\quad Compact encoder & 2.82 $\pm$ 0.04 & 5.73 $\pm$ 0.09 & 9.41 $\pm$ 0.10 & 2.84 $\pm$ 0.04 & 5.95 $\pm$ 0.04 & 9.76 $\pm$ 0.07 \\
\multicolumn{7}{@{}l}{\emph{References}} \\
\quad constant velocity, best-of-16$^{b}$ & 2.21 & 5.36 & 9.00 & 2.09 & 4.91 & 8.11 \\
\quad constant velocity, deterministic & 3.56 & 9.47 & 16.53 & 3.69 & 9.03 & 15.33 \\
\quad quantization floor, absolute tokens$^{c}$ & 0.32 & 0.33 & 0.33 & 0.41 & 0.40 & 0.40 \\
\quad quantization floor, relative tokens$^{c}$ & 0.32 & 0.32 & 0.32 & 0.32 & 0.32 & 0.32 \\
\bottomrule
\end{tabular}
\par\smallskip\begin{minipage}{0.95\textwidth}\footnotesize\raggedright
$^{a}$~Table~I of \citet{nguyen2021traisformer}, converted from nautical miles, on a chronological split of Danish AIS in which the same vessels appear in training and test.\par
$^{b}$~The anchor's speed and course propagated forward, perturbed by isotropic Gaussian noise whose scale is fitted on validation, sampled 16 times and given the same oracle selection; it reads nothing from the trajectory beyond its last report.\par
$^{c}$~Error incurred by encoding the true future position into the tokens and decoding it back; no model using that representation can go below it.\par
\end{minipage}
\end{table*}

\textbf{Reproduction.} Under TrAISformer's own split convention (a
chronological cut with vessels shared) and its best-of-16 decoder, our
reimplementation reaches 0.86, 1.75 and 2.84~km at one, two and three hours on
the Danish corpus, against the published 0.89, 1.74 and
3.04~km~\citep{nguyen2021traisformer}: within 7\% at all horizons, and never
more than 1\% above the published value. The agreement holds across a sweep
spanning 7.4--57.4~M parameters, over which the one-hour best-of-16 error on
the vessel-disjoint split varies only between 1.04 and 1.09~km, so it does not
rest on a fortunate hyper-parameter choice. We take this as evidence that the
comparator is a faithful rendering of the published system. No published
number exists for the US corpus.

\textbf{The oracle decoder.} Decoding the same trained TrAISformer greedily
instead of with the best-of-16 oracle moves its one-hour error from 1.05 to
2.23~km on the Danish corpus and from 0.89 to 2.20~km on the US corpus, and
its three-hour error from 3.36 to 8.46~km and from 2.66 to 8.44~km. The oracle
is therefore worth a factor of 2.1 to 2.5 on the Danish corpus and 2.5 to 3.2
on the US corpus, larger than the vessel-sharing gains of the selected configurations in Section~\ref{sec:leakage}. AISFormer behaves the same way. With relative
tokens the factor at one hour is 2.3 on both corpora, so the oracle effect
does not depend on how position is represented.

Some of that factor is genuine multimodality and some is the metric rewarding
spread. Giving the same oracle to the noise-perturbed constant-velocity control
separates them. Best-of-16 improves the noise-perturbed constant-velocity control from
3.56 to 2.21~km at one hour on the Danish corpus and from 3.69 to 2.09~km on
the US corpus, a factor of 1.6 to 1.8 obtained without reading the trajectory.
In log terms, about three fifths of the improvement the oracle confers on
TrAISformer would accrue to any sufficiently diverse predictor, on both
corpora. The remainder is real: TrAISformer's best-of-16 error is 2.1 (Danish)
and 2.3 (US) times better than the control's, and its greedy error beats
deterministic constant velocity by a factor of 1.6 to 1.7. The system does
learn, but a best-of-$N$ figure and a single-shot figure are different
quantities. Tabulating one beside the other, as happens when sampled
trajectory models are compared with regression
baselines~\citep{li2024diffutraj, fang2026sde} or when a variety-style oracle
is adopted from other domains~\citep{gupta2018socialgan, thiede2019variety},
overstates the accuracy of the former by a factor of 2 to 3.

The released TrAISformer code also differs from plain cross-entropy training
in two details, a multi-resolution loss and a decoder restricted to the top-10
logits and a 40-bin vicinity of the previous position. Adding them one at a
time on the Danish corpus (Table~\ref{tbl:protocolabl}, Appendix~\ref{app:supp})
leaves the best-of-16 figure almost unchanged (0.98 to 1.05~km) but moves the
oracle factor from 2.6 for plain cross-entropy to 2.1 for the full released
protocol, mainly through the decoder restriction. We report the full protocol
throughout.

\textbf{Single-shot comparison.} Under matched, single-shot decoding the
ranking of the systems changes. TrAISformer's greedy error is 21\%, 11\% and
10\% lower than the compact encoder's at one, two and three hours on the
Danish corpus (2.23 against 2.82~km at one hour) and 22\%, 15\% and 14\% lower
on the US corpus, for 17 times the parameters. GATransformer is less accurate
than TrAISformer at one hour on both corpora but the most accurate single-shot
model at three hours on the Danish corpus (7.82~km). The 1.1~M-parameter LSTM
on displacements is within 10\% of greedy TrAISformer at every horizon on the
Danish corpus and better than it at every horizon on the US corpus (2.16,
4.60 and 7.76~km against 2.20, 5.07 and 8.44~km). TrAISformer with relative
tokens equals the absolute model at one hour on the US corpus but falls behind
at longer horizons on both (11.98 against 8.46~km at three hours on the
Danish corpus, 9.61 against 8.44~km on the US one). It inherits the absolute
model's hyper-parameters and vicinity sampler, which were selected for
absolute tokens and not re-tuned, so it serves here as a one-hour control for
Sections~\ref{sec:leakage} and~\ref{sec:shift}, not as a better predictor.
Finally, encoding the true future position into the tokens and decoding it
back gives a mean error of 0.32~km on the Danish corpus and 0.41~km on the US
one, so the
published one-hour best-of-16 figure sits about three times above the floor
its own discretization imposes. Broken down by maneuver
(Appendix~\ref{app:r2-maneuver}), every model gains most over constant
velocity on turning windows. Under RMSE or the 90th percentile instead of
the mean error, GATransformer ranks first at three hours on both corpora
(Appendix~\ref{app:r2-metrics}).

\section{Component audits}\label{sec:components}
\subsection{Frequency-aware attention (AISFormer-inspired)}\label{sec:aisformer}

AISFormer differs from TrAISformer in its attention
alone; representation, loss, decoder, trainer, schedule and evaluation code
are identical, and each has its own sweep. Both select their smallest arm
(7.4~M and 8.3~M parameters, respectively); the largest variants (57.4~M and
63.3~M) overfit within a few epochs. Under best-of-16 at one hour the two are indistinguishable:
1.06~$\pm$~0.03 against 1.05~$\pm$~0.01~km over three seeds each, and they
stay within 0.03~km of each other at two and three hours. Under greedy
decoding of the same trained models AISFormer is 2\% worse at one hour
(2.27~$\pm$~0.03 against 2.23~$\pm$~0.04~km) and 6\% worse at two and three
hours (5.37~$\pm$~0.10 against 5.08~$\pm$~0.05~km, and 8.99~$\pm$~0.10
against 8.46~$\pm$~0.05~km), a gap of several seed standard deviations at the
longer horizons. Under a single random draw the two are again level. AISFormer
is not better than TrAISformer beyond seed noise at any horizon under any
decoder we tried, and the
spectral path costs about twice the training step time at matched batch size.
On the US corpus AISFormer is again no better: 0.94 against 0.89~km under
best-of-16 and 2.25 against 2.20~km under greedy decoding at one hour, and 4\%
worse under greedy decoding at three hours (8.80 against 8.44~km).

The oracle hides part of this. Best-of-16 rewards a diverse predictor and
compresses the distance between systems that differ in how sharply they
concentrate probability, so the frequency-aware attention appears neutral
through that protocol alone and mildly worse under greedy decoding at longer
horizons. The claim is narrow. Our
spectral operator may differ from the original, and the source reports
horizons up to ten hours, where a mechanism aimed at long-range structure has
more room to act than at three. What the comparison establishes is where this
family's performance comes from: both systems sit far above the motion
models and the quantization floor, and at the same place as each other, so
the shared sparse classification framing appears to account for most of the
performance, and the added attention provides no measurable benefit under
the tested conditions.

\subsection{Graph attention over surrounding traffic (GATransformer)}\label{sec:gat}

GATransformer regresses the whole horizon at once under mean squared error, so
best-of-$N$ has no meaning for it and Table~\ref{tbl:tfprotocol} reports it
single-shot. On the Danish vessel-disjoint split it reaches
3.86~$\pm$~0.10~km ADE and 7.82~$\pm$~0.15~km FDE over three hours, and
2.33~$\pm$~0.09~km at one hour. Table~\ref{tbl:gatablation} reproduces its own
ablations on that split; the neighborhood sweep and the neighbor-provenance arms
below are also Danish, and the US corpus repeats the two ablations that separate
the engineered feature from the learned mechanism.

We replicate this system at six seeds rather than three. At three seeds its
baseline spread was 0.035~km, which made two ablations look resolved at three
times their own noise; six seeds put it at 0.10~km with the mean essentially
unmoved. The iterative-decoding row behaved the same (0.015~km over three
seeds, 0.10~km over six). In both cases the point estimate survived and only
the stated precision moved, the harmless version of this failure, which is
detectable only by replication. That seed-to-seed variance can exceed the
differences a benchmark reports is a known
pitfall~\citep{bouthillier2021variance}. We therefore compare each ablation
against the standard error of its difference from the full model rather than
against a raw spread.

\begin{table*}
\caption{GATransformer on the vessel-disjoint split: the source's two input feature sets, three variant models and two neighbor-set arms. ADE and FDE are the mean error over the three-hour horizon and at its final step. $\Delta$ ADE is relative to the full model; absolute values are not comparable with the source, because the dataset, region and sampling interval differ. Entries are means over 6 seeds with sample standard deviations, except ``Neighbors thinned to restricted'' at 5, one replicate having been disqualified by an optimiser stall.}\label{tbl:gatablation}
\begin{tabular*}{\tblwidth}{@{} L R R R@{}}
\toprule
Configuration & ADE (km) & FDE (km) & $\Delta$ ADE \\
\midrule
GATransformer (full model) & 3.86\,$\pm$\,0.10 & 7.82\,$\pm$\,0.15 & --- \\
Kinematics only, no waterway & 4.71\,$\pm$\,0.07 & 9.52\,$\pm$\,0.09 & $+$21.9\% \\
Without temporal encoder & 3.96\,$\pm$\,0.07 & 8.03\,$\pm$\,0.09 & $+$2.5\% \\
Without spatial encoder & 3.78\,$\pm$\,0.06 & 7.71\,$\pm$\,0.12 & $-$2.1\% \\
Single-step iterative decoding & 4.03\,$\pm$\,0.10 & 8.34\,$\pm$\,0.20 & $+$4.4\% \\
Neighbors thinned to restricted & 3.84\,$\pm$\,0.07 & 7.78\,$\pm$\,0.16 & $-$0.6\% \\
Neighbors from own split only & 3.79\,$\pm$\,0.06 & 7.71\,$\pm$\,0.13 & $-$1.8\% \\
\bottomrule
\end{tabular*}
\vspace{2pt}
\begin{minipage}{\tblwidth}\footnotesize\raggedright
About three quarters of a test window's neighbors are training vessels. Restricting a window to its own split thins its neighborhood as well as changing whose it is, so the ``thinned'' row matches that thinning while leaving the traffic unrestricted.
\end{minipage}
\end{table*}

The effects are very unequal. Removing the waterway distances and leaving the
kinematic channels alone costs 21.9\%, seventeen standard errors clear of
the full model and the largest single effect in this paper's comparator
studies; this supports the source's emphasis on that feature. Removing the
temporal encoder costs 2.5\%, marginal at under two standard errors. Removing the
\emph{spatial} encoder, the graph attention over surrounding vessels that
the system is named for, costs nothing: the apparent 2.1\% improvement (2.2 paired standard errors) is
small and does not hold across neighborhoods (Table~\ref{tbl:gatnbr}), so the
six seeds support that the mechanism gives no measurable benefit here, not that
it is reliably harmful. It gives
none while accounting for half of the parameter count.

This is not evidence that vessel interactions are uninformative in general.
The source works in the Ningbo-Zhoushan approaches at one-minute sampling,
where a three-hour horizon covers a small fraction of the ground it covers
here and traffic is dense enough that a neighbor's maneuver constrains the
target's. On a 46,000~km$^2$ region sampled every ten minutes, a vessel has
about three neighbors within 10~km and none in open water, and the next
three hours are dominated by where it was already going. Here an engineered
feature accounts for the system's advantage, while the learned interaction
mechanism does not measurably improve the error. AISFormer showed a related pattern,
in which the mechanism was neutral under one decoder and mildly worse under
another; in both systems, the named mechanism is not what drives accuracy.

\textbf{The null is not an artifact of our neighborhood.} The source does
not fix a neighborhood definition, so ours (the eight nearest vessels
within 10~km) is a choice, and the obvious objection is that it gave the
graph too few neighbors to use. Table~\ref{tbl:gatnbr} varies the definition over a 4.7-fold range in
how full the neighborhood is, from 16\% of neighbor slots occupied on the
test split to 76\%, retraining at six seeds each. The
without-spatial-encoder control never reads the neighbor tensor, so one
six-seed control serves every row.

\begin{table*}
\caption{GATransformer's spatial encoder against the neighborhood definition. Occupancy is the fraction of neighbor slots filled on the test split. ADE and FDE are six-seed means with sample standard deviations. The last column is each row's ADE minus the without-spatial-encoder control (3.78\,km), paired per seed, with the standard error of the difference; positive means the graph attention hurts.}\label{tbl:gatnbr}
\begin{tabular*}{\tblwidth}{@{} L R R R R@{}}
\toprule
Neighborhood & Occupancy & ADE (km) & FDE (km) & $\Delta$ ADE vs.\ no spatial encoder \\
\midrule
$K{=}8$ within 5\,km & 16\% & 3.82\,$\pm$\,0.06 & 7.75\,$\pm$\,0.12 & +0.04\,$\pm$\,0.04 \\
$K{=}16$ within 10\,km & 20\% & 3.83\,$\pm$\,0.05 & 7.75\,$\pm$\,0.09 & +0.05\,$\pm$\,0.01 \\
$K{=}8$ within 10\,km (reported) & 38\% & 3.86\,$\pm$\,0.10 & 7.82\,$\pm$\,0.15 & +0.08\,$\pm$\,0.04 \\
$K{=}8$ within 20\,km & 76\% & 3.79\,$\pm$\,0.07 & 7.70\,$\pm$\,0.13 & +0.01\,$\pm$\,0.02 \\
\bottomrule
\end{tabular*}
\end{table*}

The objection does not hold. At no neighborhood does the spatial encoder
help: all four differences against the control are positive. The arm closest
to the control is the \emph{densest}: with 76\% of slots filled against 38\%
at the reported setting, the full model lands 0.01~$\pm$~0.02~km from the
model with no graph attention. Quadrupling the traffic the graph can see does
not make it useful; it brings the full model to the ablation that deletes the
graph. Two
caveats apply. The differences are small (0.3\% to 2.1\% of ADE) and only two
of the four are resolved at more than two standard errors, so graph attention
here is inert to mildly harmful, not reliably harmful at any setting. And all
four arms share distance-ranked selection and ten-minute sampling, so a
definition that selected neighbors by predicted conflict is untested.
Split by interaction class at the forecast origin
(Appendix~\ref{app:r2-encounter}), the encoder gives no detectable improvement
in head-on, crossing or overtaking encounters either, the situations it is
designed for.

\textbf{Decoding step by step costs accuracy.} The source's SSI-Prediction
emits one step at a time and feeds each prediction back, against the
single-shot regression used in every other row. It is 4.4\% worse, at
4.03~$\pm$~0.10~km ADE (about four standard errors). The evidence is stronger at
the final step: FDE rises 6.7\%, to 8.34~$\pm$~0.20 against
7.82~$\pm$~0.15~km, five standard errors clear. That asymmetry is what
compounding error would produce, since it falls hardest on the last point.
We treat the FDE margin as the more reliable half of the result; the ADE margin
is narrower than the means suggest, since the single-shot baseline's weakest
seed (4.01~km) still exceeds the iterative decoder's best (3.87~km).

An iterative decoder can be trained on the true history or on its own
predictions, and the article does not say which. Conditioning on the true
history fails outright: in a 20-epoch pilot it reached 9.41~km test ADE
against 4.16~km for a decoder trained on its own rollout, and its validation
error rose from the fifth epoch while training loss kept falling, which is
consistent with exposure bias. We select the training mode on validation and
report the free-running decoder, so the penalty above is the strategy at its
best.

\textbf{The leakage channel is real and, here, empty.} A vessel-disjoint
manifest cannot make a held-out window's \emph{surroundings} unseen: on the
Danish corpus 73.4\% of occupied neighbor slots in a test window carry training
vessels, against
70.2\% in a training window, and no manifest closes this. Whether it inflates
anything is answered by the last two rows of Table~\ref{tbl:gatablation}.
Restricting a window to neighbors from its own split lowers ADE by 1.2\%
against the density-matched control, the wrong sign for leakage; the point
estimate is the same as at three seeds.
Thinning the neighborhood itself does not matter either: the control keeps
about an eighth of a test window's neighbors (4.9\% of slots occupied
against 37.6\%) and lands on the unrestricted model within 0.6\%, about one
standard error.

That control is the least stable arm to train. Matching the restricted arm's
density means matching its \emph{sparsity}: slots are occupied 27.4\% of the
time in a training window against 5.8\% in validation and 4.9\% in test, a
factor of five. One of its six replicates stops learning under that
asymmetry, reproducibly at the same seed, and is excluded under the policy of
Appendix~\ref{app:comparators}, so the row reports five seeds where the rows
it controls for report six. We prefer the sparse control to a denser one
that would confound provenance with density, but a reconstruction should
expect it to be the hardest arm to optimize.

The two findings explain each other. A channel can only leak what the model
reads through it, and this model reads almost nothing through its neighbors.
We record the null rather than the channel: a model whose graph attention
carried weight could well show inflation, and the density-matched control is
the design that would separate it from a change in density. The channel should be
declared in any paper whose model reads co-present traffic, and measured
rather than assumed.

\textbf{US corpus.} Table~\ref{tbl:gatablation-noaa} repeats the two
ablations that separate this system's engineered feature from its learned
interaction mechanism. The waterway feature is worth 11.8\% there against
21.9\% on the Danish corpus: present on both corpora, and smaller where much of
the largest class (tug and towing traffic, Section~\ref{sec:data}) already
moves along narrow, well-defined channels that the kinematic features partly
encode. Removing the spatial encoder changes ADE by $-2.7\%$, the same sign as
the Danish $-2.1\%$, so graph attention over neighboring vessels again gives no
measurable benefit. Both architectural findings replicate.

\begin{table}
\caption{GATransformer on the NOAA vessel-disjoint split: the two ablations replicated from Table~\ref{tbl:gatablation}. Entries are means over 3 seeds (42/43/44) with sample standard deviations; $\Delta$ ADE is relative to the full model. The neighbor-density and leakage-channel arms of Table~\ref{tbl:gatablation} were not re-run on this corpus.}\label{tbl:gatablation-noaa}
\begin{tabular*}{\tblwidth}{@{} L R R R@{}}
\toprule
Configuration & ADE (km) & FDE (km) & $\Delta$ ADE \\
\midrule
GATransformer (full model) & 4.06\,$\pm$\,0.10 & 8.04\,$\pm$\,0.20 & --- \\
Kinematics only, no waterway & 4.54\,$\pm$\,0.08 & 9.06\,$\pm$\,0.20 & $+$11.8\% \\
Without spatial encoder & 3.95\,$\pm$\,0.06 & 7.77\,$\pm$\,0.15 & $-$2.7\% \\
\bottomrule
\end{tabular*}
\end{table}

\subsection{Waypoint graph for gap imputation (MGFormer-inspired)}\label{sec:imputeresult}

The fourth comparator is measured on a different task. MGFormer fills a long
gap in a track from the observations on both sides of it;
Section~\ref{sec:tfmethod} defines the protocol, in which every 36-step
window loses the same block of 11 steps and all systems reconstruct
identical holes. Table~\ref{tbl:impute} reports the vessel-disjoint result.

\begin{table*}
\caption{Long-range gap imputation (11 of 36 steps masked, about 110~minutes). Each cell is the mean error over the masked steps in km, with the ratio to linear interpolation on the same test windows in parentheses; for every other manifest the ratio's change against the random-window manifest (\emph{Random} column) is given in brackets. \emph{Mid} is the error at the middle step of the gap on the vessel-disjoint manifest; \emph{graph on all} fits the waypoint graph on every window instead of the training windows. Mean over three seeds.}\label{tbl:impute}
\centering\footnotesize\setlength{\tabcolsep}{3.5pt}
\begin{tabular}{@{}llrrrrrr@{}}
\toprule
Corpus & Model & Vessel & Chronological & Lon.\ cut & Random & Mid & Graph on all \\
\midrule
Danish & MGFormer-inspired & 1.004 (0.620) [+4\%] & 0.897 (0.611) [+3\%] & 2.604 (1.772) [+198\%] & 0.898 (0.595) & 1.244 & 0.998 \\
 & MGFormer, no graph & 1.020 (0.629) & --- & --- & --- & 1.329 & --- \\
 & Compact encoder & 1.033 (0.638) [+1\%] & 0.934 (0.636) [+1\%] & 1.710 (1.164) [+85\%] & 0.951 (0.630) & 1.323 & --- \\
 & Linear interpolation & 1.620 & 1.468 & 1.470 & 1.509 & & \\
\midrule
US & MGFormer-inspired & 1.017 (0.701) [$-$1\%] & 1.054 (0.722) [+1\%] & 1.876 (1.386) [+95\%] & 1.068 (0.712) & 1.210 & 1.020 \\
 & MGFormer, no graph & 1.123 (0.774) & --- & --- & --- & 1.449 & --- \\
 & Compact encoder & 1.120 (0.772) [$-$1\%] & 1.128 (0.772) [$-$1\%] & 1.466 (1.082) [+39\%] & 1.171 (0.780) & 1.439 & --- \\
 & Linear interpolation & 1.450 & 1.461 & 1.354 & 1.501 & & \\
\bottomrule
\end{tabular}
\end{table*}

Over the missing steps, our MGFormer-inspired reconstruction reaches 1.004~$\pm$~0.015~km
against 1.620~km for linear interpolation, a 38\% improvement, of the right
order for the 21\% reduction over baselines the source reports at a
comparable missing ratio. That headline is the part the reconstruction can
distort. The part it cannot distort is what happens when one component is
removed inside one implementation. Deleting the waypoint graph and its graph
network costs only 1.6\% overall (1.020~$\pm$~0.016 against
1.004~$\pm$~0.015~km), about one seed standard deviation, but 6.8\% at the
mid-gap step (1.329 against 1.244~km), roughly three to four. The graph earns its
place where the mechanism predicts it should, at the point of the gap where
the nearest observation is furthest away and route structure is the only
information left, and essentially nowhere else; averaged over the whole gap
the contribution is diluted below the noise floor. The compact encoder with its
masked-reconstruction head (0.60~M parameters) lands at
1.033~$\pm$~0.006~km, within noise of the no-graph ablation at equal
capacity.

On the US corpus the reconstruction reaches 1.017~km against 1.450~km for
linear interpolation (30\% better), and the waypoint graph matters more:
removing it costs 10.4\% overall (1.123 against 1.017~km) and 19.8\% at the
middle of the gap (1.449 against 1.210~km). The compact encoder lands at
1.120~km, again level with the no-graph ablation. The graph's effect thus has
the same direction on both corpora, largest at mid-gap on both, and a size that
depends on the traffic.

\begin{figure*}
\centering
\includegraphics[width=\textwidth]{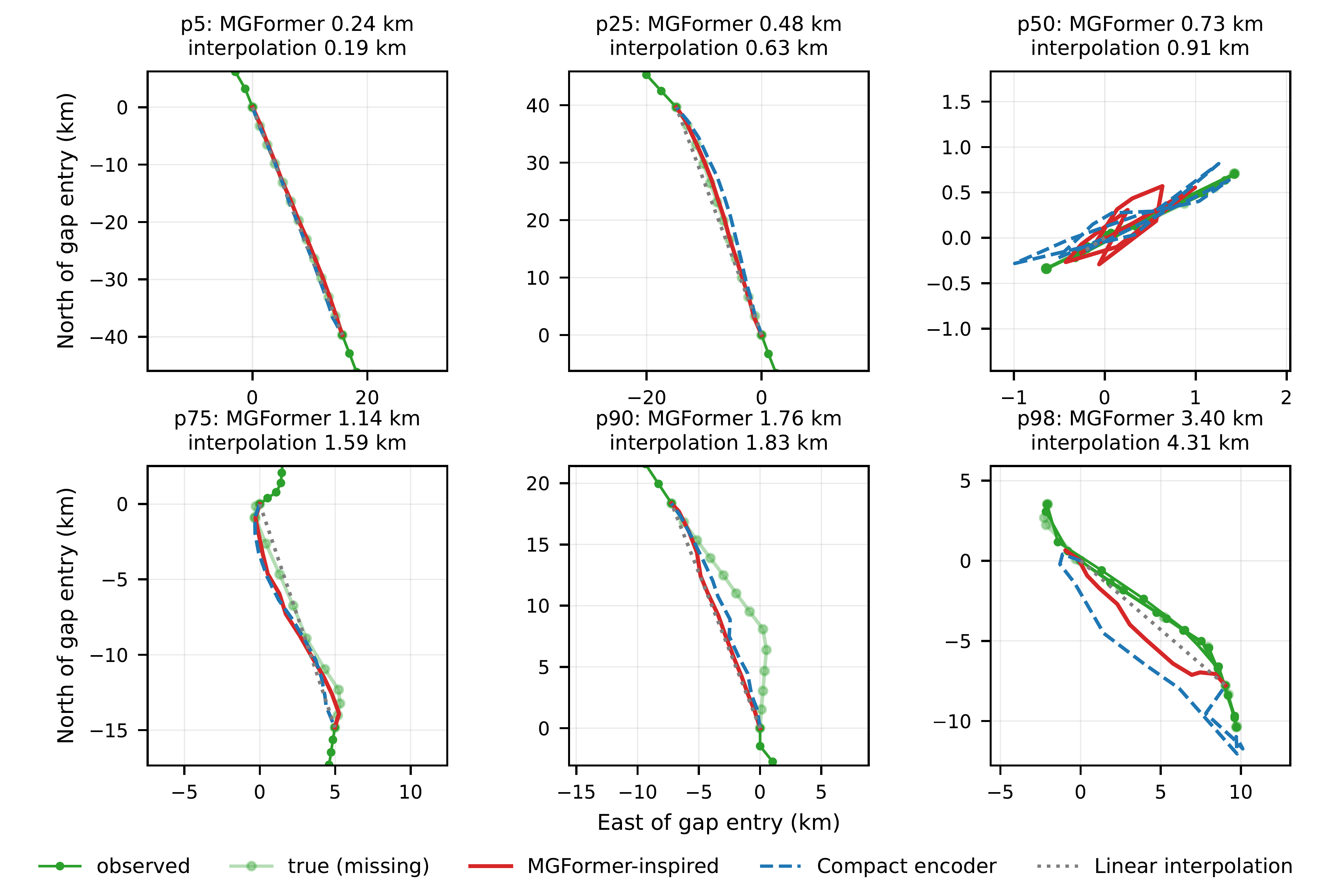}
\caption{Six imputed gaps from the Danish vessel-disjoint test split (seed-42 models), drawn at fixed percentiles of MGFormer's per-window error (p5--p98 in the panel titles); within each percentile band the window with the median interpolation error is shown (rule-based, not hand-picked). Each panel is framed in kilometers on the last observed report, and the scale differs between panels (about 45~km across at p5, about 3~km at p50). Green: observed steps; pale green: the eleven missing steps. In the median-error example neither learned system is reliably better than the straight line; the average advantage is carried by the minority of gaps where interpolation fails badly (see text).}
\label{fig:qualimpute}
\end{figure*}

Fig.~\ref{fig:qualimpute} shows where the 38\% comes from, using six test
gaps drawn at fixed percentiles of MGFormer's own error. In the median-error example
the advantage is modest. At the fifth percentile a straight line beats the
model (0.19 against 0.24~km), and from the twenty-fifth to the ninetieth
percentile the model leads by 0.06 to 0.45~km (0.48 against 0.63~km at the
twenty-fifth, 1.14 against 1.59~km at the seventy-fifth, 1.76 against 1.83~km at
the ninetieth). At the median the vessel is nearly stationary, the whole track
spans under three kilometers, and both learned systems draw zigzags around it.
The average advantage comes from elsewhere. The straight line's error has a
heavy tail that is unrelated to MGFormer's own rank: the 28\% of test gaps where
interpolation misses by more than 2~km carry 96\% of the total difference
between the two, and MGFormer is the better of the two on only 59\% of gaps
(0.99 against 1.62~km on average for the figure's checkpoint). The learned
systems therefore gain their advantage on the minority of gaps where the vessel
does not travel in a straight line, the same result the mid-gap column gives,
seen one track at a time.

\begin{figure*}
\centering
\includegraphics[width=0.85\textwidth]{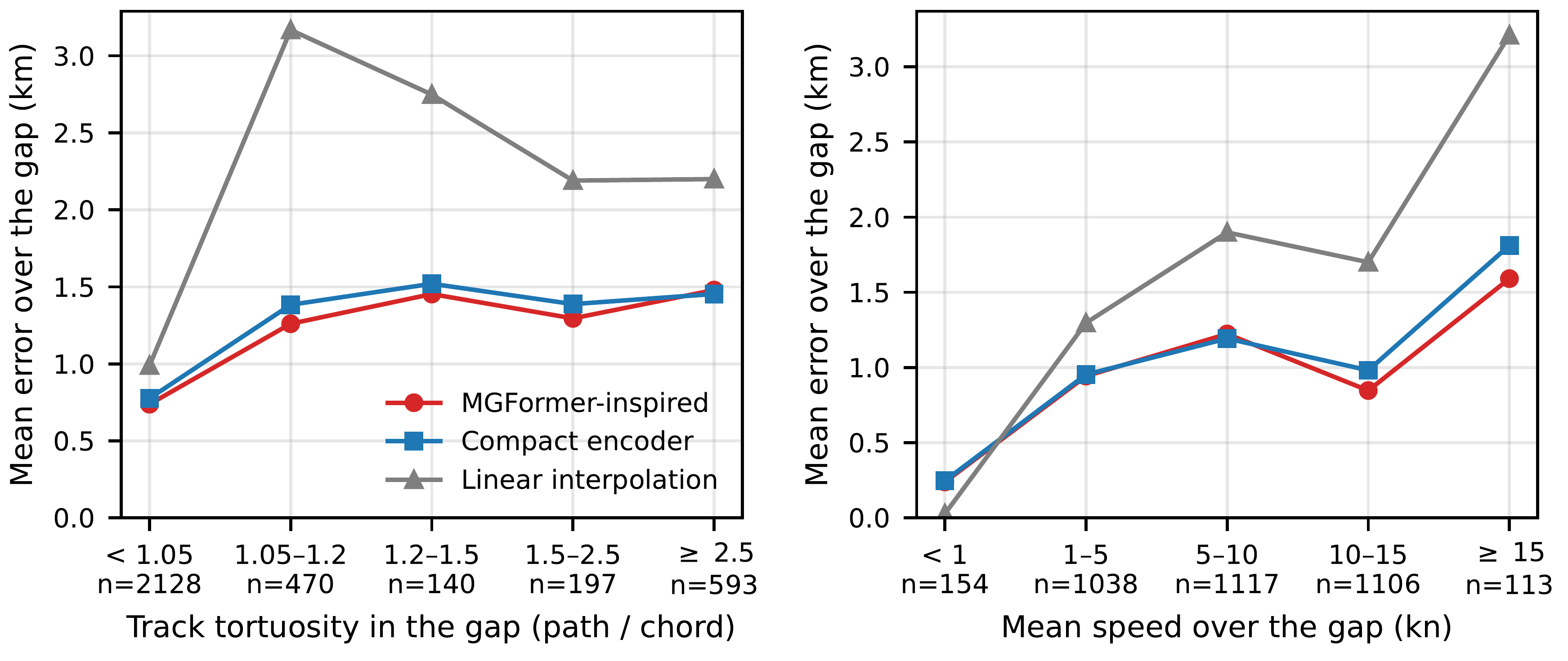}
\caption{Gap-imputation error on the Danish vessel-disjoint test windows (3,528 gaps, seed-42 models) against the tortuosity of the missing track (path length over straight-line distance, left) and the mean speed over the gap (right); $n$ is the number of gaps per bin. Interpolation fails on bending tracks, where the learned models keep most of their accuracy; on nearly stationary vessels (below 1~kn) interpolation is almost exact and the learned models are not.}
\label{fig:impute-factors}
\end{figure*}

The panels of Fig.~\ref{fig:qualimpute} differ in scale because the vessels
moved different distances. The median gap spans 26.6~km between the last
report before it and the first after it, so a mean error of about 1~km is about
4\% of the distance covered, and the 45-km-wide panel belongs to a fast vessel.
Fig.~\ref{fig:impute-factors} relates the error to the shape of the missing
track and to speed. Interpolation's error depends above all on how much the
track bends: 0.99~km on near-straight gaps (path length within 5\% of the
straight-line distance) against 2.2--3.2~km on bending ones. The learned models
rise much less, from 0.74 to 1.3--1.5~km, and MGFormer beats interpolation on
60--83\% of bending gaps against 52\% of near-straight ones. Across gaps,
MGFormer's error correlates with tortuosity (Spearman 0.41) far more than with
speed (0.07) or with the distance covered ($-$0.06); the mean error of all three
methods still rises with speed, because faster vessels cover more ground. Two
cases account for the visibly large errors. Below 1~kn interpolation is almost
exact (0.03~km) while the learned models scatter around the vessel (0.24~km),
which is the zigzag in the median panel; and the worst 10\% of gaps carry 33\%
of MGFormer's total error and are the turning ones (median tortuosity 1.24
against 1.02 over all gaps). The large panels therefore reflect long, curved
gaps rather than a failure of the reconstruction, and the advantage of the
learned models is concentrated where the track turns.

The last column of Table~\ref{tbl:impute} tests something the split
manifests cannot reach. The waypoint graph is fitted from trajectories, so it
has its own training data, and a manifest partitions the windows a
model trains on, not the windows from which an auxiliary structure is
extracted. A graph built over the whole dataset has in principle seen the
routes of the vessels held out for testing. This is the maritime instance of
preprocessing performed on training and test data together, among the
commonest leakage forms in a cross-disciplinary
survey~\citep{kapoor2023leakage}. Refitting the graph over every window,
with model, training, split and holes unchanged and three seeds each,
changes the error by 0.6\% at 256 waypoints (0.998~$\pm$~0.007 against
1.004~$\pm$~0.015~km) and by 0.2\% in the opposite direction at 1024
waypoints (0.991~$\pm$~0.010 against 0.989~$\pm$~0.013~km). Both lie inside
the seed spread and the sign does not hold, so on this dataset the channel
does not open. On the US corpus the whole-dataset graph gives 1.020 against
1.017~km, the same null. Quadrupling the Danish graph's resolution (2,926 to
14,253 edges) is itself worth 1.5\%, again about one seed spread.

The traffic is the likely reason on the Danish corpus. Its feed inside this region is
lane-dominated: vessels follow a few heavily used routes through the
Kattegat and the Belts, so a held-out vessel's track is already described by
waypoints that training vessels laid down. Adding the held-out tracks yields
6\% more edges at 256 waypoints and 9\% more at 1024, but almost no new route
structure. The US corpus, dominated by tug and tanker traffic in narrow
channels, behaves the same way. That protection belongs to the data, not the
method: traffic with idiosyncratic per-vessel routing, such as fishing grounds or offshore
service, or a graph fine enough to resolve a berth, would not enjoy it. The
useful conclusion is procedural. Fitting a graph, codebook, clustering or
normalization statistic over the full dataset before splitting is routine;
whether it leaks is a property of the dataset and cannot be inferred from the
manifest; and the check costs one retraining run. Any system with a fitted
auxiliary component should report where it was fitted from alongside its
split protocol.

\section{Leakage}\label{sec:leakage}\label{sec:noaaleakage}

\begin{figure*}
\centering
\includegraphics[width=0.9\textwidth]{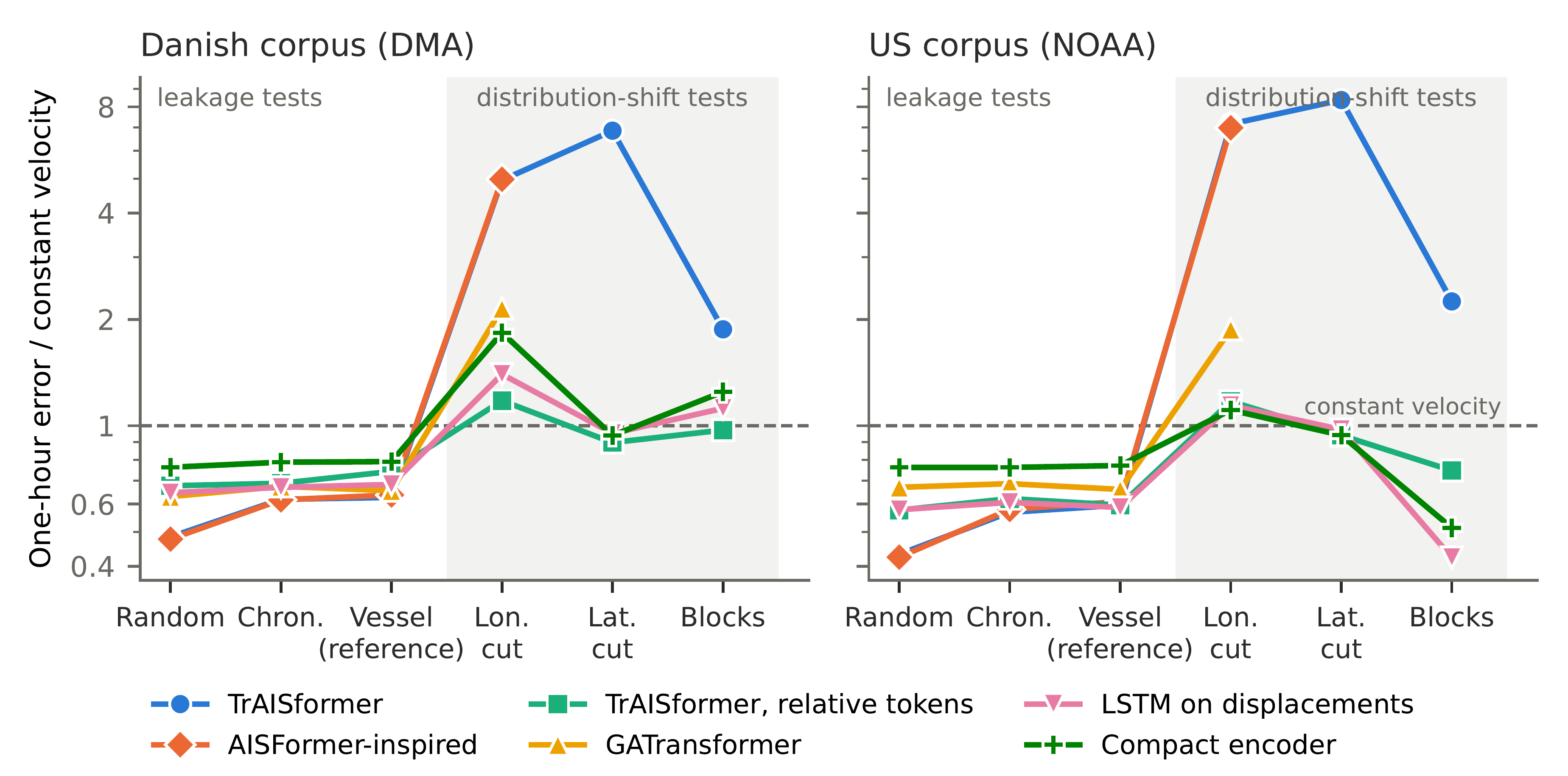}
\caption{One-hour single-shot error of every prediction model on every manifest, as a ratio to constant velocity on that manifest's test windows (log scale; dashed line: constant velocity). Manifests run from the most permissive leakage test to the hardest distribution-shift test; \emph{Chron.} is the chronological manifest without purging, and \emph{Vessel} the vessel-disjoint reference. Values in Tables~\ref{tbl:leak} and~\ref{tbl:shift}.}\label{fig:grid}
\end{figure*}

Every result so far is measured on the vessel-disjoint manifest. This section
asks how much a split that relaxes it would have overstated, on both corpora.
The prediction windows are drawn from each voyage every hour, a dense,
overlapping regime in which leakage should matter most, because overlapping
windows of one voyage can fall on both sides of a permissive split. Each
system is retrained on every manifest with the window pool, the architecture and
the training budget fixed. Table~\ref{tbl:leak} and Fig.~\ref{fig:grid} report the one-hour error as a
ratio to constant velocity, which gives the manifests' test sets, whose
difficulty differs, a common reference (Section~\ref{sec:metrics}).

\begin{table*}
\caption{One-hour single-shot error as a ratio to constant velocity on each manifest's own test windows, from the vessel-disjoint reference to the most permissive manifest. \emph{Gap} is the vessel-disjoint ratio minus the random-window ratio, as a share of the vessel-disjoint ratio: the error reduction a random split would report. For the chronological manifests, the change of the ratio against the random-window ratio is given in brackets. Mean over three seeds (six for Danish GATransformer), with the sample standard deviation over seeds in the line below; raw errors in km are in the released archive~\citep{raisi2026code}. \emph{rel.}: relative position tokens. Lower is better.}\label{tbl:leak}
\centering\footnotesize\setlength{\tabcolsep}{3pt}
\begin{tabular}{@{}lrrrrr@{\hspace{1.2em}}rrrrr@{}}
\toprule
 & \multicolumn{5}{c}{Danish (DMA)} & \multicolumn{5}{c}{US (NOAA)} \\
\cmidrule(lr){2-6}\cmidrule(lr){7-11}
Model & Vessel & Purged & Chron. & Random & Gap & Vessel & Purged & Chron. & Random & Gap \\
\midrule
TrAISformer & 0.627 & 0.622\,{\scriptsize[+28\%]} & 0.619\,{\scriptsize[+28\%]} & 0.484 & 23\% & 0.596 & 0.579\,{\scriptsize[+34\%]} & 0.569\,{\scriptsize[+31\%]} & 0.433 & 27\% \\
 & {\scriptsize$\pm$0.010} & {\scriptsize$\pm$0.005} & {\scriptsize$\pm$0.004} & {\scriptsize$\pm$0.003} &  & {\scriptsize$\pm$0.006} & {\scriptsize$\pm$0.005} & {\scriptsize$\pm$0.011} & {\scriptsize$\pm$0.005} &  \\
AISFormer-inspired & 0.637 & 0.612\,{\scriptsize[+28\%]} & 0.617\,{\scriptsize[+29\%]} & 0.477 & 25\% & 0.610 & 0.575\,{\scriptsize[+35\%]} & 0.580\,{\scriptsize[+37\%]} & 0.424 & 30\% \\
 & {\scriptsize$\pm$0.010} & {\scriptsize$\pm$0.010} & {\scriptsize$\pm$0.004} & {\scriptsize$\pm$0.001} &  & {\scriptsize$\pm$0.010} & {\scriptsize$\pm$0.006} & {\scriptsize$\pm$0.008} & {\scriptsize$\pm$0.004} &  \\
\addlinespace
TrAISformer, rel. & 0.742 & --- & 0.687\,{\scriptsize[+2\%]} & 0.676 & 9\% & 0.596 & --- & 0.622\,{\scriptsize[+8\%]} & 0.577 & 3\% \\
 & {\scriptsize$\pm$0.011} &  & {\scriptsize$\pm$0.001} & {\scriptsize$\pm$0.001} &  & {\scriptsize$\pm$0.005} &  & {\scriptsize$\pm$0.006} & {\scriptsize$\pm$0.003} &  \\
GATransformer & 0.655 & --- & 0.673\,{\scriptsize[+7\%]} & 0.630 & 4\% & 0.660 & --- & 0.685\,{\scriptsize[+2\%]} & 0.669 & $-$1\% \\
 & {\scriptsize$\pm$0.026} &  & {\scriptsize$\pm$0.016} & {\scriptsize$\pm$0.011} &  & {\scriptsize$\pm$0.008} &  & {\scriptsize$\pm$0.008} & {\scriptsize$\pm$0.020} &  \\
LSTM on displacements & 0.682 & --- & 0.669\,{\scriptsize[+4\%]} & 0.646 & 5\% & 0.587 & --- & 0.607\,{\scriptsize[+5\%]} & 0.578 & 1\% \\
 & {\scriptsize$\pm$0.004} &  & {\scriptsize$\pm$0.004} & {\scriptsize$\pm$0.000} &  & {\scriptsize$\pm$0.006} &  & {\scriptsize$\pm$0.003} & {\scriptsize$\pm$0.003} &  \\
Compact encoder & 0.791 & 0.786\,{\scriptsize[+3\%]} & 0.788\,{\scriptsize[+3\%]} & 0.763 & 4\% & 0.772 & 0.771\,{\scriptsize[+1\%]} & 0.762\,{\scriptsize[+0\%]} & 0.762 & 1\% \\
 & {\scriptsize$\pm$0.012} & {\scriptsize$\pm$0.014} & {\scriptsize$\pm$0.012} & {\scriptsize$\pm$0.010} &  & {\scriptsize$\pm$0.012} & {\scriptsize$\pm$0.010} & {\scriptsize$\pm$0.017} & {\scriptsize$\pm$0.019} &  \\
\midrule
Constant velocity (km) & 3.56 & 3.13 & 3.13 & 3.39 &  & 3.69 & 3.80 & 3.80 & 3.84 &  \\
\bottomrule
\end{tabular}
\end{table*}

\subsection{Vessel sharing}\label{sec:vesselleak}

For the two absolute-token systems, sharing vessels between training and test
is worth a large error reduction on both corpora. TrAISformer's ratio falls
from 0.627 on the vessel-disjoint manifest to 0.484 on the random-window one on
the Danish corpus (a 23\% gap) and from 0.596 to 0.433 on the US corpus (27\%);
AISFormer shows 25\% and 30\%. Under best-of-16 the Danish gap is 33\% (0.295 to
0.196) and the US gap 29\% (0.242 to 0.172). The four other models show much
smaller gaps: 9\% and 3\% for TrAISformer with relative tokens, 4\% and $-$1\%
for GATransformer, 5\% and 1\% for the LSTM, and 4\% and 1\% for the
compact encoder.

The contrast between the two TrAISformer variants is the informative one,
because they share the architecture, the capacity, the loss, the decoder and
the training schedule and differ only in the position tokens. Changing the
tokens from absolute cells to offsets from the anchor cuts the gap from 23\% to
9\% on the Danish corpus and from 27\% to 3\% on the US corpus. Sensitivity to
shared vessels is therefore not a property of the model class as such: it
follows whether the input states where the vessel is. A model that reads
absolute positions can learn vessel-specific routes at specific cells, and a
random split rewards it for doing so, because the test windows of a shared
vessel lie on routes it has already seen. Where the vessel-disjoint split is
enforced, absolute TrAISformer's validation error bottoms out at epoch 14 and
rises afterwards, whereas on the random split it keeps improving to epoch 29.
The models that read only motion relative to the vessel gain little from the
same shared windows.

The relative representation is not uniformly better. In distribution, on the
vessel-disjoint manifest, absolute tokens are 16\% more accurate on the Danish
corpus (2.23 against 2.64~km) and equal on the US corpus (2.20~km each). On
lane-dominated Danish waters a location prior is worth something for new
vessels in known waters. Section~\ref{sec:capacity} shows that this advantage
shrinks with capacity, and Section~\ref{sec:shift} shows what it costs in new
waters.

\subsection{Chronological splits}\label{sec:timeleak}

The chronological manifest shares 55--69\% of its test vessels with training but
separates the periods by at least 4.3~days. Measured as a ratio to constant
velocity, it lowers TrAISformer's error by about 1\% against the
vessel-disjoint manifest on the Danish corpus (0.619 against 0.627) and by
4.5\% on the US corpus (0.569 against 0.596). Across all six models and both
corpora, the chronological ratio lies within 8\% of the vessel-disjoint one, in
either direction; for the two absolute-token systems it recovers at most a
fifth of what the random-window manifest gains. In raw kilometers the two corpora appear to disagree:
TrAISformer's error falls by 13\% on the Danish corpus (2.23 to 1.94~km) but by
2\% on the US corpus (2.20 to 2.16~km). The difference is in the test sets, not
in the model: the Danish chronological test windows are easier (constant
velocity 3.13 against 3.56~km on the vessel-disjoint test windows), while the
US ones are not (3.80 against 3.69~km). Normalized, the effect is small and of
the same sign on both corpora.

The windows at a boundary between adjacent subsets can share up to 5.7~h of
reports (Section~\ref{sec:splits}). Purging them produces small changes:
TrAISformer moves from 1.94 to 1.95~km on the Danish corpus and from 2.16 to
2.20~km on the US one. Over the three systems and both corpora, the seed-paired
differences lie between $-0.019$ and $+0.041$~km (at most 1.9\%), and only
the Danish TrAISformer change ($+0.010\pm0.002$~km, 0.5\%) exceeds two standard
errors. Every ratio stays within 0.01 of the unpurged value, and no conclusion
changes. The large
random-window gain therefore needs the shared windows to be close in time to
the test windows; sharing the vessel across days is worth little.

\subsection{Causal test: the test vessels' own history}\label{sec:causal}

The manifests above differ in more than whether vessels are shared: a random
split also changes how much of each vessel's track the model sees.
Table~\ref{tbl:causal} isolates the vessel effect. The test set is held fixed
and only the training set changes: the leaky arms replace training-vessel
windows with windows of the test vessels themselves that share no report with
any test window, either earlier in time (\emph{early}) or from other voyages
(\emph{voyage}), so the number of training windows is unchanged.

\begin{table}
\caption{Causal test of vessel leakage (TrAISformer, one-hour error, km). The test set is fixed within each block and only the training set changes; $\Delta$ is the seed-paired difference to \emph{clean} (mean $\pm$ SE, three seeds). \emph{reduced}: clean minus $|H|$ random windows (data-volume control); \emph{h25}--\emph{h100}: equal-size training sets in which 25--100\% of the test vessels' available history $H$ replaces training-vessel windows.}\label{tbl:causal}
\centering\footnotesize\setlength{\tabcolsep}{2pt}
\begin{tabular}{@{}llrrrr@{}}
\toprule
 & Arm & Greedy & $\Delta$ greedy & Bo16 & $\Delta$ Bo16 \\
\midrule
Danish & early, clean & 2.11 & --- & 1.029 & --- \\
 & early, reduced & 2.19 & $+0.079\pm0.023$ & 1.016 & $-0.013\pm0.010$ \\
 & early, h25 & 1.88 & $-0.230\pm0.038$ & 0.839 & $-0.191\pm0.004$ \\
 & early, h50 & 1.85 & $-0.260\pm0.026$ & 0.833 & $-0.197\pm0.016$ \\
 & early, h100 & 1.82 & $-0.292\pm0.045$ & 0.815 & $-0.214\pm0.010$ \\
 & voyage, clean & 2.11 & --- & 0.955 & --- \\
 & voyage, h100 & 1.88 & $-0.223\pm0.035$ & 0.806 & $-0.149\pm0.011$ \\
\midrule
US & early, clean & 2.21 & --- & 0.892 & --- \\
 & early, reduced & 2.10 & $-0.110\pm0.029$ & 0.867 & $-0.026\pm0.012$ \\
 & early, h25 & 2.13 & $-0.072\pm0.053$ & 0.862 & $-0.030\pm0.019$ \\
 & early, h50 & 2.06 & $-0.148\pm0.043$ & 0.826 & $-0.066\pm0.019$ \\
 & early, h100 & 2.04 & $-0.168\pm0.051$ & 0.853 & $-0.039\pm0.008$ \\
 & voyage, clean & 2.21 & --- & 0.892 & --- \\
 & voyage, h100 & 1.89 & $-0.322\pm0.019$ & 0.780 & $-0.112\pm0.017$ \\
\bottomrule
\end{tabular}
\end{table}

On the Danish corpus, adding a test vessel's earlier windows lowers the greedy
error from 2.11 to 1.82~km (a paired difference of $-0.292\pm0.045$~km,
$-$13.8\%) and the best-of-16 error by $-0.214\pm0.010$~km; other voyages of the
same vessels give $-0.223\pm0.035$ and $-0.149\pm0.011$~km. A clean arm reduced
by the same number of windows does not improve ($+0.079\pm0.023$~km), so the
gain is not a data-volume effect. It also saturates quickly: a quarter of the
available history already gives $-0.230$~km.

The US corpus replicates the voyage variant, more strongly: other voyages of
the test vessels lower the greedy error by $-0.322\pm0.019$~km ($-$14.6\%) and
the best-of-16 error by $-0.112\pm0.017$~km. The early variant is weaker. Its
dose-response is monotone ($-0.072$, $-0.148$ and $-0.168$~km for 25, 50 and
100\% of the history, $-$7.6\% at full dose), but the reduced control also
improves on the clean arm ($-0.110\pm0.029$~km), unlike on the Danish corpus.
The step from \emph{reduced} to \emph{h100}, which adds the same history on top
of an identical base, is then only $-0.058$~km. We therefore treat the voyage
variant as the evidence on the US corpus and the early variant as consistent
with it but not conclusive. On both corpora, what a random split makes
available to the model is the test vessel's own history, and the model uses it.

\subsection{Capacity}\label{sec:capacity}

\begin{figure*}
\centering
\includegraphics[width=0.9\textwidth]{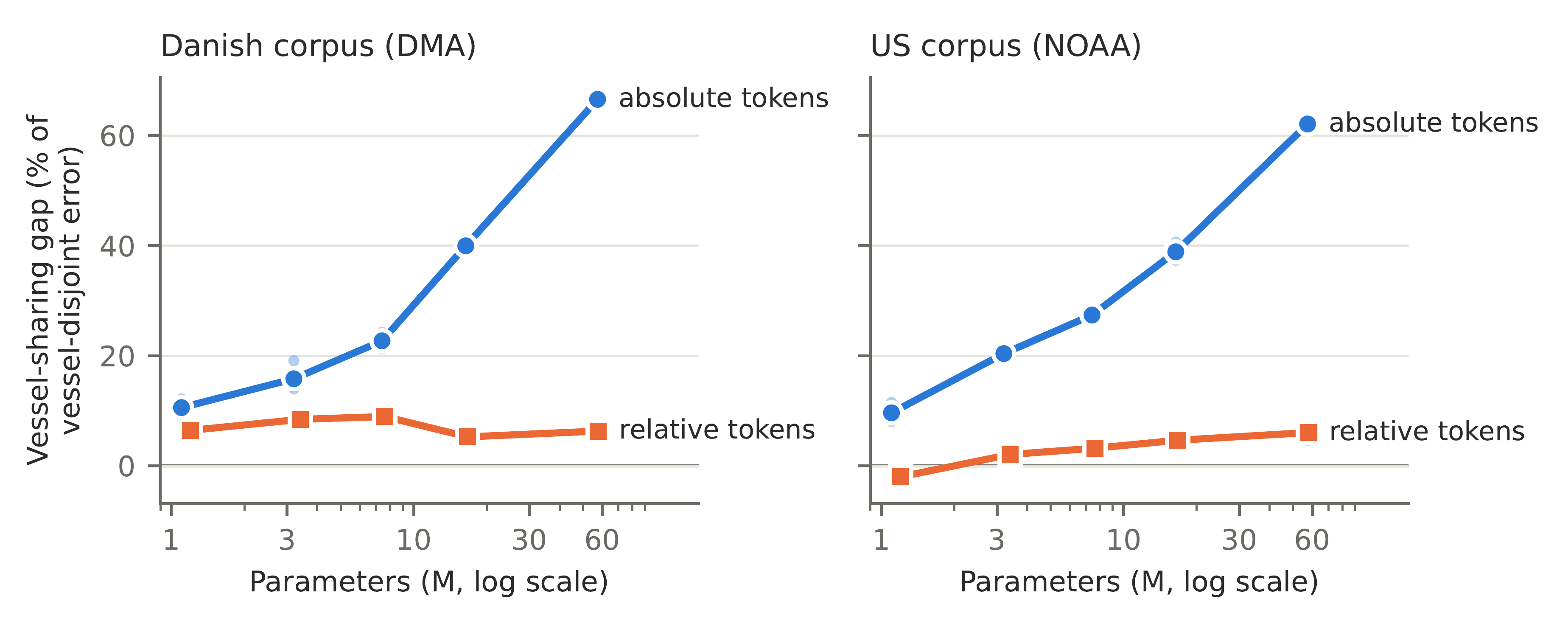}
\caption{Vessel-sharing gap against model size for TrAISformer with absolute and relative position tokens: the one-hour greedy ratio on the vessel-disjoint manifest minus that on the random-window manifest, as a share of the former. Lines join the means over three seeds; small marks are single seeds. Values in Table~\ref{tbl:capacity}.}\label{fig:capacity}
\end{figure*}

If vessel-sharing gains come from memorizing absolute locations, they should
grow with the capacity to memorize, and only for absolute tokens.
Fig.~\ref{fig:capacity} and Table~\ref{tbl:capacity} test both predictions
with TrAISformer at five sizes, from 1.1 to 57.4~M parameters, on the
vessel-disjoint and random-window manifests of both corpora, three seeds per
point.

With absolute tokens, the gap grows monotonically with size on both corpora,
from 11\% at 1.1~M to 67\% at 57.4~M parameters on the Danish corpus and from
10\% to 62\% on the US corpus. The slope of the per-seed gap on
$\log_{10}$ parameters is 0.212 per decade (95\% seed-bootstrap interval
[0.160, 0.263]) on the Danish corpus and 0.180 [0.150, 0.203] on the US corpus,
with Spearman $\rho=0.98$ on both. Nearly all of the change is on the
permissive side: the vessel-disjoint ratio stays between 0.63 and 0.68 (Danish)
and 0.59 and 0.60 (US) at every size, while the random-window ratio falls from
0.610 to 0.216 and from 0.535 to 0.226. Extra capacity does not help a new
vessel; it helps a vessel the model has already seen.

With relative tokens, the gap stays between 5 and 9\% at every size on the
Danish corpus and between $-$2 and 6\% on the US corpus. The slopes are
$-0.011$ [$-0.023$, $-0.003$] and 0.027 [0.018, 0.035] per decade, and the
difference between the absolute and relative slopes is 0.223 [0.168, 0.286] on
the Danish corpus and 0.153 [0.116, 0.183] on the US corpus; both intervals
exclude zero. Capacity therefore amplifies leakage substantially only in
absolute-location models: the relative-token slopes, $-0.01$ and 0.03 per
decade, are small beside the absolute ones (0.21 and 0.18). The relative models also improve with size on
the vessel-disjoint manifest (from 0.809 to 0.670 and from 0.679 to 0.560),
whereas the absolute ones do not, so the in-distribution advantage of absolute
tokens noted in Section~\ref{sec:vesselleak} shrinks with capacity: at 57~M
parameters, relative tokens are better on the US corpus (0.560 against 0.598)
and within 4\% on the Danish corpus (0.670 against 0.647).
The training and validation curves (Appendix~\ref{app:r2-capacity}) give no
indication that under-training explains these trends: the larger absolute-token
models reach their best vessel-disjoint validation error by epoch 11 of 30 and
then degrade while their training loss keeps falling.

\subsection{Gap imputation under relaxed splits}\label{sec:imputeleak}

Gap imputation shows no comparable leakage on either corpus
(Table~\ref{tbl:impute}). Against linear interpolation on the same test
windows, the MGFormer-inspired model's ratio is 4\% worse on the
vessel-disjoint manifest than on the random-window one on the Danish corpus
(0.620 against 0.595) and 1\% better on the US corpus (0.701 against 0.712);
the chronological manifest lies within 3\% of the random-window ratio on both,
and the compact encoder within 1\%. Both imputation models are small (0.71~M
and 0.60~M parameters), which is consistent with Section~\ref{sec:capacity}, where
vessel-sharing gains grow with capacity. The region manifest is a
different matter (Section~\ref{sec:shift}).

\section{Distribution shift: region and vessel-type hold-outs}\label{sec:shift}

The spatial manifests withhold part of the area rather than part of the fleet,
so they measure transfer to new waters. Table~\ref{tbl:shift} reads them
against the vessel-disjoint reference on both corpora. Because 37--65\% of
their test vessels also appear in training, they evaluate transfer to new waters
for a mixture of familiar and unfamiliar vessels, not a fully vessel- and
region-disjoint case. Each hold-out's change against the random-window ratio is
given in brackets.

\begin{table*}
\caption{One-hour single-shot error as a ratio to constant velocity under the spatial hold-outs, against the vessel-disjoint reference. A ratio above 1 is worse than extrapolating the anchor's speed and course. Constant velocity itself (km) is in the last row; the unseen-bin rows give the share of test-context absolute latitude/longitude bins that never occur in training. In brackets: the change of each hold-out's ratio against the same model's random-window ratio (Table~\ref{tbl:leak}). \emph{rel.}: relative position tokens. Mean over three seeds (six for Danish GATransformer), with the sample standard deviation over seeds in the line below.}\label{tbl:shift}
\centering\footnotesize\setlength{\tabcolsep}{3pt}
\begin{tabular}{@{}lrrrr@{\hspace{1.2em}}rrrr@{}}
\toprule
 & \multicolumn{4}{c}{Danish (DMA)} & \multicolumn{4}{c}{US (NOAA)} \\
\cmidrule(lr){2-5}\cmidrule(lr){6-9}
Model & Vessel & Lon.\ cut & Lat.\ cut & Blocks & Vessel & Lon.\ cut & Lat.\ cut & Blocks \\
\midrule
TrAISformer & 0.627 & 4.958\,{\scriptsize[+924\%]} & 6.847\,{\scriptsize[+1314\%]} & 1.875\,{\scriptsize[+287\%]} & 0.596 & 7.151\,{\scriptsize[+1553\%]} & 8.362\,{\scriptsize[+1833\%]} & 2.249\,{\scriptsize[+420\%]} \\
 & {\scriptsize$\pm$0.010} & {\scriptsize$\pm$0.109} & {\scriptsize$\pm$0.231} & {\scriptsize$\pm$0.145} & {\scriptsize$\pm$0.006} & {\scriptsize$\pm$0.069} & {\scriptsize$\pm$0.501} & {\scriptsize$\pm$0.060} \\
AISFormer-inspired & 0.637 & 4.988\,{\scriptsize[+945\%]} & --- & --- & 0.610 & 6.977\,{\scriptsize[+1544\%]} & --- & --- \\
 & {\scriptsize$\pm$0.010} & {\scriptsize$\pm$0.080} &  &  & {\scriptsize$\pm$0.010} & {\scriptsize$\pm$0.379} &  &  \\
\addlinespace
TrAISformer, rel. & 0.742 & 1.178\,{\scriptsize[+74\%]} & 0.896\,{\scriptsize[+33\%]} & 0.972\,{\scriptsize[+44\%]} & 0.596 & 1.172\,{\scriptsize[+103\%]} & 0.939\,{\scriptsize[+63\%]} & 0.746\,{\scriptsize[+29\%]} \\
 & {\scriptsize$\pm$0.011} & {\scriptsize$\pm$0.016} & {\scriptsize$\pm$0.025} & {\scriptsize$\pm$0.004} & {\scriptsize$\pm$0.005} & {\scriptsize$\pm$0.039} & {\scriptsize$\pm$0.012} & {\scriptsize$\pm$0.033} \\
GATransformer & 0.655 & 2.148\,{\scriptsize[+241\%]} & --- & --- & 0.660 & 1.875\,{\scriptsize[+180\%]} & --- & --- \\
 & {\scriptsize$\pm$0.026} & {\scriptsize$\pm$0.195} &  &  & {\scriptsize$\pm$0.008} & {\scriptsize$\pm$0.213} &  &  \\
LSTM on displacements & 0.682 & 1.399\,{\scriptsize[+117\%]} & 0.951\,{\scriptsize[+47\%]} & 1.120\,{\scriptsize[+73\%]} & 0.587 & 1.141\,{\scriptsize[+97\%]} & 0.975\,{\scriptsize[+69\%]} & 0.423\,{\scriptsize[$-$27\%]} \\
 & {\scriptsize$\pm$0.004} & {\scriptsize$\pm$0.014} & {\scriptsize$\pm$0.010} & {\scriptsize$\pm$0.007} & {\scriptsize$\pm$0.006} & {\scriptsize$\pm$0.018} & {\scriptsize$\pm$0.010} & {\scriptsize$\pm$0.004} \\
Compact encoder & 0.791 & 1.834\,{\scriptsize[+141\%]} & 0.938\,{\scriptsize[+23\%]} & 1.247\,{\scriptsize[+64\%]} & 0.772 & 1.108\,{\scriptsize[+45\%]} & 0.940\,{\scriptsize[+23\%]} & 0.514\,{\scriptsize[$-$33\%]} \\
 & {\scriptsize$\pm$0.012} & {\scriptsize$\pm$0.091} & {\scriptsize$\pm$0.026} & {\scriptsize$\pm$0.024} & {\scriptsize$\pm$0.012} & {\scriptsize$\pm$0.006} & {\scriptsize$\pm$0.013} & {\scriptsize$\pm$0.010} \\
\midrule
Constant velocity (km) & 3.56 & 3.02 & 4.14 & 2.50 & 3.69 & 3.44 & 3.25 & 7.52 \\
Unseen abs.\ lat./lon.\ bins & 0/0\% & 0/100\% & 100/31\% & 0/0\% & 0/0\% & 0/100\% & 100/0\% & 32/0\% \\
\bottomrule
\end{tabular}
\end{table*}

\subsection{Straight cuts break absolute-position models}\label{sec:shift-cuts}

Under a longitude cut, the two absolute-token systems fail on both corpora: on
the Danish corpus TrAISformer's one-hour error rises from 2.23 to 14.98~km, five
times constant velocity on the held-out side (ratio 4.96), and on the US corpus
from 2.20 to 24.59~km (ratio 7.15); AISFormer behaves the same way. A latitude
cut does the same (28.3 and 27.2~km; ratios 6.85 and 8.36), so the degradation does
not depend on the direction of the cut. Under both cuts, essentially all test
contexts fall in absolute position bins that never occur in training: 100\% of
the longitude bins under the longitude cut on both corpora, and 100\% of the
latitude bins under the latitude cut.

The same architecture with relative tokens does not degrade in this way: its ratios are
1.18 and 0.90 (Danish longitude and latitude cuts) and 1.17 and 0.94 (US).
Measured in kilometers, relative tokens remove 90 and 94\% of the absolute
model's degradation on the Danish corpus and 92 and 97\% on the US corpus, where
the removed share is the absolute model's region error minus the relative one,
over the absolute model's region error minus its vessel-disjoint error. The
other translation-invariant models also stay bounded: 0.95--1.40 for the LSTM
and 0.94--1.83 for the compact encoder. GATransformer, which reads
distances to waterway intersections fitted on the training side, lies between
the two groups (2.15 and 1.88 under the longitude cut).

\subsection{Dependence on absolute location}\label{sec:shift-mechanism}

The straight cuts cannot separate two explanations: the absolute model fails
because its held-out bins are untrained, or because it has memorized
\emph{locations} and the held-out waters are new ones. The spatial-block
manifest separates them. Each held-out block shares its latitude and longitude
ranges with training blocks elsewhere, and on the Danish corpus none of their
test-context latitude or longitude bins is unseen.
Absolute TrAISformer still degrades, to 4.70~km, 2.1 times its vessel-disjoint
error and 1.9 times constant velocity (ratio 1.88), while relative TrAISformer
reaches 2.43~km (0.97). Unseen bins therefore do not explain the
degradation on their own. Together with the relative-token control, the result
indicates a strong dependence on the absolute-location representation,
consistent with reliance on location-specific patterns (combinations of latitude
and longitude, or the motion associated with them) that transfer poorly to
held-out areas, even when each bin has been trained. On the US corpus, where 32\% of the held-out
blocks' latitude bins are unseen, the absolute model reaches 16.9~km (2.25) and
the relative one 5.62~km (0.75). The vessel-sharing gain of
Section~\ref{sec:vesselleak} is consistent with the same reading, seen from the other side:
an absolute-position model is rewarded for knowing the places of its training
data and penalized in places it does not know. The failure is already present
at the first step: given the true history, absolute TrAISformer's one-step
error at one hour under the longitude cut is 6.9~km (Danish) and 10.6~km (US),
against 0.53--0.66~km for relative tokens (Appendix~\ref{app:r2-decoding}).

\subsection{The shift that remains}\label{sec:shift-composition}

Relative models still degrade in new waters, but by less: in ratio terms,
TrAISformer with relative tokens is 1.2--2.0 times its own vessel-disjoint ratio
across the six spatial hold-outs, the LSTM 0.7--2.1 times and the compact encoder
0.7--2.3 times, against 3.0--14.0 times for absolute TrAISformer.
This remainder reflects the distribution shift the region manifests were meant
to measure, together with any residual imbalance and model-specific limits, and
the held-out sides do differ in traffic. On the Danish longitude
cut the held-out side carries one-sixth the training side's share of Passenger
windows and more than twice its share of Fishing windows, and the vessel-type
total-variation distance between the sides is 0.354, against 0.197 on the US
cut; their speed and six-hour travel distributions also differ more (two-sample
Kolmogorov--Smirnov statistics 0.16 and 0.22, against 0.12 and 0.14). This
stronger traffic shift is consistent with the compact encoder's larger
region-to-random degradation on the Danish corpus (2.14 against 1.30
times in kilometers).

Table~\ref{tbl:compmatch} measures how much of the degradation this traffic
shift explains, by reweighting the test windows to the training side's vessel
types, or to its vessel types and speeds. Model and constant-velocity errors are
weighted alike. Each test window receives the weight
$\pi^{\mathrm{tr}}_k/\pi^{\mathrm{te}}_k$ of its stratum $k$, the stratum's
share of the training side over its share of the test side; strata absent from
the training side receive weight zero, training strata without test windows
cannot be represented, and weights are not capped. The matched ratio is the
weighted model error divided by the weighted constant-velocity error, and its
spread is again over seeds. Matching balances only the measured variables; route
geometry, turning behavior and other unmeasured characteristics may still
differ, and where the effective sample size is small or the weights extreme, the
matched estimate does not isolate the contribution of composition. Matching moves a harder class such as ferries (constant velocity
9.1~km on the Danish corpus) into the test set, and an error in kilometers would
register that as worse transfer. On the Danish longitude cut, matching the
vessel types explains a substantial part of the relative models' degradation:
their excess over their own vessel-disjoint ratio shrinks by a third to a
half (TrAISformer with relative tokens 1.18 to 1.03, the LSTM 1.40 to 1.06, the
compact encoder 1.83 to 1.27), and by about a third when speed is matched as
well. With the
composition matched, the compact encoder degrades by 1.6 times its
vessel-disjoint ratio on the Danish corpus and 1.4 times on the US one, so the
traffic mix accounts for most of the difference between the corpora. On the
other spatial hold-outs, matching lowers the relative models' ratios by less
than 10\%, or raises them, as on the US spatial blocks; the exception is
the speed-matched US latitude cut discussed below.

The absolute models' degradation survives every matching: absolute TrAISformer stays at
3.3--7.8 times constant velocity under both cuts on both corpora, and at 2.4--2.8
on the US spatial blocks, where the relative variant stays at 0.83--0.86.
The largest change is on the US latitude cut, where matching speed as well as
type lowers it from 8.4 to 4.8; the weights there are extreme (effective
sample size 6\% of the test windows), but the held-out side does differ strongly
in speed. The Danish spatial blocks cannot be matched at all: the held-out blocks
are a fishing ground (82\% Fishing windows, mean speed 3.1~kn), and too few of
their windows resemble the training traffic. On the US corpus the held-out
blocks cover the Galveston--Bolivar ferry shuttle (Passenger and Other about 50\%
of windows, constant velocity 7.5~km). Composition shift is therefore real and
explains part of the relative models' degradation, but not the absolute-position
models' degradation.

\begin{table*}
\caption{Spatial hold-outs with the test windows reweighted to the training side's vessel-type mix (\emph{type}) or to its joint distribution of vessel type and anchor speed in quartiles (\emph{type $\times$ speed}). Values are one-hour single-shot error as a ratio to constant velocity, both weighted alike, so a harder reweighted test set does not register as worse transfer. \emph{ESS}: Kish effective sample size of the weights, as a share of the test windows; cells below 5\% are omitted and those below 10\% marked $^{\dagger}$. All rows of a block come from the same three runs per model, those that saved per-window errors; some are re-runs of the runs in Table~\ref{tbl:shift}, so the unmatched ratios can differ slightly from it. \emph{rel.}: relative position tokens.}\label{tbl:compmatch}
\centering\footnotesize\setlength{\tabcolsep}{4pt}
\begin{tabular}{@{}llrrrrrr@{}}
\toprule
Corpus & Test windows & TrAISformer & AISFormer-insp. & TrAISformer, rel. & LSTM & Compact enc. & ESS \\
\midrule
Danish & Vessel-disjoint & 0.63 & --- & 0.74 & 0.68 & 0.79 & \\
 & Longitude cut & 5.03 & --- & 1.18 & 1.40 & 1.83 &  \\
 & \quad type-matched & 3.28 & --- & 1.03 & 1.06 & 1.27 & 35\% \\
 & \quad type $\times$ speed & 3.81 & --- & 1.02 & 1.16 & 1.44 & 23\% \\
 & Latitude cut & 6.85 & --- & 0.90 & 0.95 & 0.94 &  \\
 & \quad type-matched & 6.62 & --- & 0.88 & 0.88 & 0.90 & 53\% \\
 & \quad type $\times$ speed & 6.70 & --- & 0.87 & 0.86 & 0.88 & 47\% \\
 & Spatial blocks & 1.88 & --- & 0.97 & 1.12 & 1.25 &  \\
 & \quad type-matched & --- & --- & --- & --- & --- & 4\% \\
 & \quad type $\times$ speed & --- & --- & --- & --- & --- & 2\% \\
\midrule
US & Vessel-disjoint & 0.59 & --- & 0.60 & 0.59 & 0.76 & \\
 & Longitude cut & 7.10 & 6.98 & 1.17 & 1.14 & 1.10 &  \\
 & \quad type-matched & 7.80 & 7.65 & 1.16 & 1.12 & 1.04 & 61\% \\
 & \quad type $\times$ speed & 7.62 & 7.46 & 1.15 & 1.12 & 1.04 & 77\% \\
 & Latitude cut & 8.36 & --- & 0.94 & 0.98 & 0.94 &  \\
 & \quad type-matched & 6.96 & --- & 0.91 & 0.95 & 1.04 & 24\% \\
 & \quad type $\times$ speed & 4.82$^{\dagger}$ & --- & 0.81$^{\dagger}$ & 0.72$^{\dagger}$ & 0.81$^{\dagger}$ & 6\% \\
 & Spatial blocks & 2.25 & --- & 0.75 & 0.42 & 0.51 &  \\
 & \quad type-matched & 2.77 & --- & 0.83 & 0.52 & 0.63 & 43\% \\
 & \quad type $\times$ speed & 2.40 & --- & 0.86 & 0.57 & 0.69 & 23\% \\
\bottomrule
\end{tabular}
\end{table*}

\subsection{Vessel-type hold-out}\label{sec:shift-type}

Holding out a vessel type rather than an area gives the same picture on the
Danish corpus (Table~\ref{tbl:type}, Appendix~\ref{app:supp}). Each arm is
tested on the same held-out vessels of one type, once trained without that type
and once on an equal-size training set that includes it. An unseen type costs
little for tankers, which follow lanes much like other cargo traffic (1.04--1.07
times the control for all three models), and much more for passenger vessels,
which run fixed, type-specific routes. There the absolute-token TrAISformer pays
3.07 times the control error, against 1.79 times with relative tokens and 1.42
times for the compact encoder: relative tokens remove 62\% of the absolute
model's excess penalty. A residual penalty remains: every model pays
more than 1.4 times on unseen ferries, whose shuttling motion has to be seen in
training. The trade-off of Section~\ref{sec:vesselleak} also reappears. When
ferries are in training, absolute tokens are better on them (2.62 against
3.59~km), so a location prior helps on known routes and hurts on unknown ones.

\section{Discussion and limitations}\label{sec:discussion}

\begin{figure*}
\centering
\includegraphics[width=0.9\textwidth]{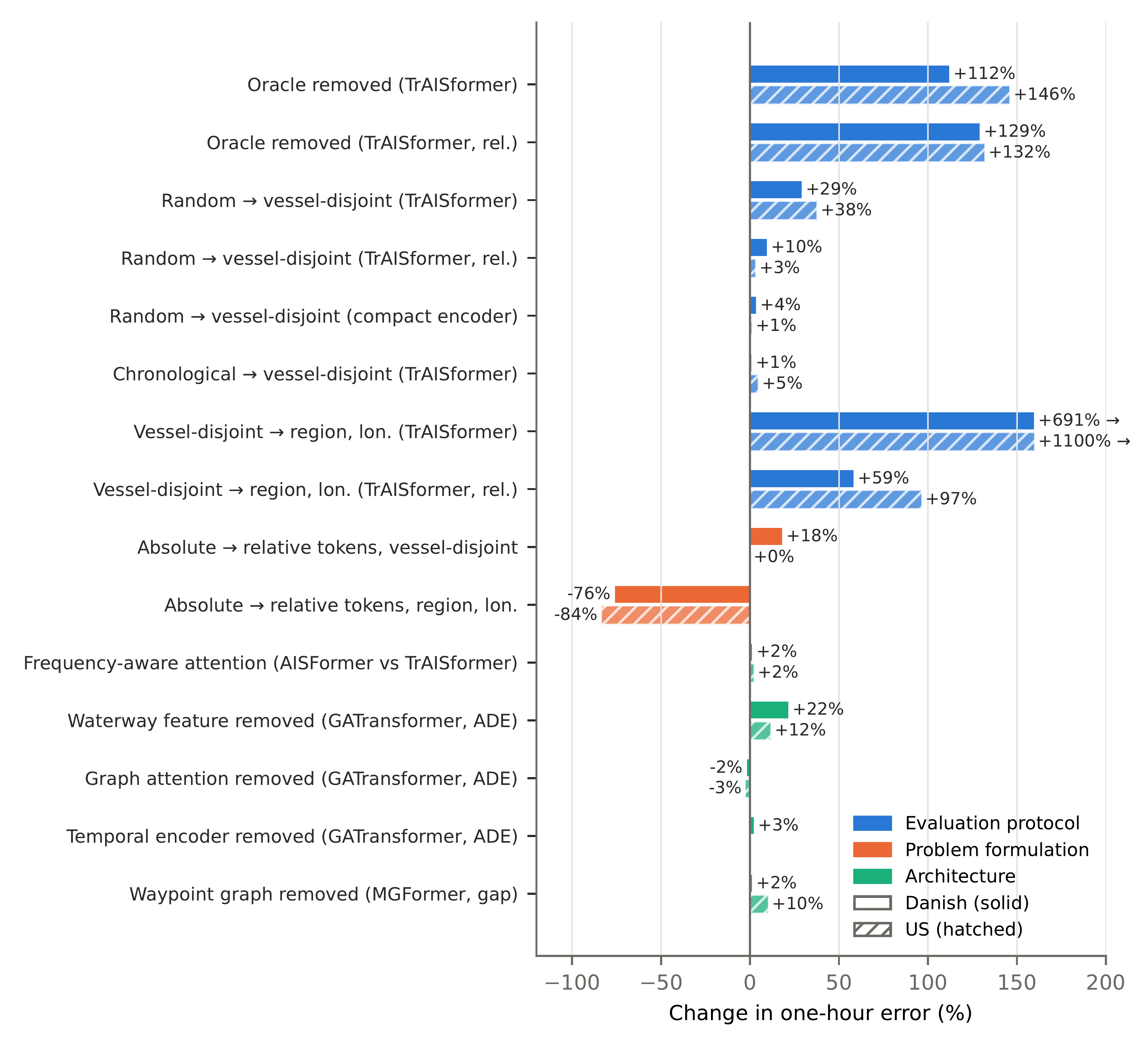}
\caption{Change in one-hour error when the protocol is tightened, the position representation changed or a component removed, on both corpora (solid: Danish; hatched: US). Split effects compare constant-velocity ratios on each manifest's own test windows; oracle effects compare decoders of the same trained models; GATransformer and MGFormer rows compare ablations (ADE over three hours, and mean error over the gap). Bars beyond 160\% are cut, with the value printed.}\label{fig:effects}
\end{figure*}

A reported error in this setting has three sources, and Fig.~\ref{fig:effects}
puts them on one scale. The evaluation protocol moves TrAISformer's one-hour error
by a factor of 2.1--2.5 through the decoder, by 29--38\%
through vessel sharing (the rise from a random to a vessel-disjoint split) and by up to an order of magnitude through a region
hold-out. The problem formulation, here the coordinate frame, decides most of
the last two. The architectural mechanisms the source papers emphasize move it
by a few percent, with the exception of GATransformer's engineered waterway
feature (12--22\%). The rest of this section discusses what these effects mean,
which of them hold on both corpora, and where the evidence stops.

\subsection{Leakage versus generalization}\label{sec:disc-leakage}

Whether a vessel-sharing gain is leakage depends on the claim it supports, and
three kinds of sharing should be kept apart: windows that share reports with a
test window (the random-window manifest), a shared vessel identity without
shared reports (the purged chronological manifest), and a vessel's earlier
history, which a deployed system may legitimately hold (the causal test). Our
primary target is prediction for vessels absent from training. For that claim,
vessel-sharing manifests do not estimate unseen-vessel performance and can be
optimistic: the causal test of Section~\ref{sec:causal} shows that the gain
comes from the test vessels' own history. For a deployment in which the same fleet recurs, such as a ferry
operator or a vessel traffic service watching familiar traffic, part of that gain
is legitimate generalization to a known vessel's future, and a split that shares
vessels can be the right one, provided that all training information precedes
the prediction time and that preprocessing and fitted structures respect the
same boundary. The error is then still not comparable with a
vessel-disjoint error, and the scenario should be stated with it. The
distribution-shift manifests answer a third question, transfer to new waters or
a new vessel type. Because the region manifests still share 37--65\% of their
test vessels with training, they measure transfer to new waters for a mixture of
familiar and unfamiliar vessels and are not a fully vessel- and region-disjoint
evaluation. A
chronological split, the convention of TrAISformer's published results, turns
out to be close to vessel-disjoint here: with 4.3~days between training and test
periods it recovers, for the absolute-token models, at most a fifth of the
random-split gain. It is a weak test
of temporal generalization over one month and should not be read as a test of
either other question.

\subsection{The coordinate frame is part of the protocol}\label{sec:disc-frame}

Absolute position tokens give a model a prior over places. In known waters the
prior is worth something: on the Danish corpus absolute TrAISformer is 16\% more
accurate than its relative variant on new vessels, and on held-out ferries,
when ferries are in training, it is 27\% more accurate (2.62 against 3.59~km). The same prior
is what a random split rewards, what grows with capacity, and what fails in new
waters, including held-out blocks whose bins were all trained. A single number
from an absolute-position model on a random or within-region split therefore
mixes skill with memory of the training waters, and the mix changes with model
size. We do not conclude that relative representations are better. The relative
variant was not re-tuned, it is worse at longer horizons, and its in-distribution
deficit shrinks with capacity but does not vanish on the Danish corpus. The
conclusion is narrower: the coordinate frame is a protocol-level choice that
decides how much of a reported number reflects leakage and how much transfers,
so it should be stated and controlled like the split.

\subsection{What holds on both corpora}\label{sec:disc-evidence}

Table~\ref{tbl:evidence} lists the main findings with their values on each
corpus. The decoding, leakage, capacity, coordinate-frame and component findings
hold on both corpora with the same sign and comparable size, despite different
geography, traffic mix and reporting interval. Four qualifications apply. The
early variant of the causal test is weaker on the US corpus, where its
data-volume control also improves, so the voyage variant carries that evidence.
The spatial-block result is cleanest on the Danish corpus, where none of the
held-out bins is unseen. The traffic-composition shift differs between the
corpora: it explains part of the relative models' degradation on the Danish
longitude cut and little on the US one. And five analyses, the vessel-type hold-out, the
TrAISformer decoder ablation and three GATransformer analyses, were run on the
Danish corpus only.

\begin{table*}
\caption{Evidence status of the main findings. \emph{Both}: measured on both corpora with the same sign and comparable size; values are one-hour, single-shot unless stated.}\label{tbl:evidence}
\centering\footnotesize\setlength{\tabcolsep}{4pt}
\begin{tabular}{@{}P{0.38\textwidth}P{0.2\textwidth}P{0.2\textwidth}P{0.12\textwidth}@{}}
\toprule
Finding & Danish (DMA) & US (NOAA) & Status \\
\midrule
\multicolumn{4}{@{}l}{\emph{Decoding (Section~\ref{sec:tfresult})}} \\
Best-of-16 oracle factor, TrAISformer (1\,h / 3\,h) & 2.1 / 2.5 & 2.5 / 3.2 & both \\
Share of the oracle factor a null predictor also obtains (log terms) & about 3/5 & about 3/5 & both \\
\multicolumn{4}{@{}l}{\emph{Leakage (Section~\ref{sec:leakage})}} \\
Vessel-sharing gap, absolute-token models (7--8\,M) & 23--25\% & 27--30\% & both \\
Vessel-sharing gap, relative tokens and regressors & 4--9\% & $-$1 to 3\% & both \\
Gap slope per decade of parameters, absolute / relative & 0.21 / $-$0.01 & 0.18 / 0.03 & both \\
Chronological effect, TrAISformer (normalized) & 1\% & 5\% & both \\
Purging the boundary windows & no change & no change & both \\
Causal history, voyage variant (h100) & $-$10.6\% & $-$14.6\% & both \\
Causal history, early variant (h100) & $-$13.8\% & $-$7.6\%, control caveat & both \\
Imputation, vessel-sharing gap (MGFormer) & 4\% & $-$1\% & both \\
\multicolumn{4}{@{}l}{\emph{Distribution shift (Section~\ref{sec:shift})}} \\
Absolute-token TrAISformer under longitude / latitude cut (ratio to constant velocity) & 4.96 / 6.85 & 7.15 / 8.36 & both \\
Share of that degradation removed by relative tokens & 90--94\% & 92--97\% & both \\
Absolute model degrades on held-out blocks & 1.88, no unseen bins & 2.25, 32\% unseen lat.\ bins & both \\
Type matching, longitude cut: absolute / compact encoder (ratio to constant velocity) & 5.03$\to$3.28 / 1.83$\to$1.27 & 7.10$\to$7.80 / 1.10$\to$1.04 & partly Danish \\
Absolute-model degradation survives composition matching & yes & yes & both \\
Vessel-type hold-out, Passenger (absolute / relative / encoder) & 3.07 / 1.79 / 1.42 & --- & Danish only \\
\multicolumn{4}{@{}l}{\emph{Components (Section~\ref{sec:components})}} \\
Frequency-aware attention improves on TrAISformer & no & no & both \\
Graph attention removed (GATransformer ADE) & $-$2.1\% & $-$2.7\% & both \\
Waterway feature removed (GATransformer ADE) & $+$21.9\% & $+$11.8\% & both \\
Waypoint graph removed, mid-gap (MGFormer) & $+$6.8\% & $+$19.8\% & both \\
Waypoint graph fitted on all windows & no effect & no effect & both \\
GATransformer neighbor sweep, provenance arms, temporal and iterative ablations & see Section~\ref{sec:gat} & --- & Danish only \\
TrAISformer decoder and loss ablation & Table~\ref{tbl:protocolabl} & --- & Danish only \\
\bottomrule
\end{tabular}
\end{table*}

\subsection{Reconstruction fidelity}\label{sec:disc-fidelity}

The comparators reproduce or reconstruct the published approaches under one
protocol; Table~\ref{tbl:deviations} records what each follows and what is ours.
TrAISformer has public code, and our comparator reproduces its published errors
within 7\%. GATransformer's article specifies the architecture, attention
equations, loss and decoder, so our comparator differs mainly in
hyper-parameters, the neighborhood definition the source leaves open, and the
data; we swept the neighborhood because it could most plausibly have produced
the null graph-attention result, and the null held across a 4.7-fold range of
occupancy. AISFormer and MGFormer are the weaker cases: each article omits
details a faithful reimplementation needs, so their absolute numbers are
properties of our reconstructions, and the comparisons they support are internal
(AISFormer differs from TrAISformer only in its attention, and MGFormer's
ablations change one component at a time). The diffusion, stochastic and
language-model approaches of Section~\ref{sec:rel-baselines} were not evaluated;
the decoding and coordinate-frame findings suggest where their numbers should be
checked first.

\subsection{Limitations}\label{sec:disc-limits}

Each corpus covers one month (May 2026 for the Danish corpus, June 2023 for the
US one). Seasonal traffic, such as the Danish leisure fleet in summer or Gulf
weather later in the hurricane season, and longer-term changes in routes are not
represented, so the chronological manifest spans days rather than seasons and
cannot test temporal generalization at the scale an operational system faces.
Both corpora share one window geometry (a 10-minute grid, three-hour context and
horizon) and one region size; other horizons and sampling rates may change the
size of the effects. The 10-minute grid also limits what an interaction model
can see: in 56\% (Danish) and 40\% (US) of the test encounters, the
close-quarters phase projected at constant velocity lasts less than one
sampling interval
(Appendix~\ref{app:r2-encounter}). The null result for graph attention holds at
this sampling interval and horizon; it does not show that vessel interactions
are uninformative. Hyper-parameters were selected on the Danish
vessel-disjoint split and reused on the US corpus and for the relative-token
control, which if anything understates their accuracy. The spatial hold-outs
are cuts within each corpus; no third geography was tested. Finally, the leakage measured
here is that of the models tested; a model that reads information the
comparators do not, such as weather, port schedules or voyage plans, may leak
through channels this protocol does not cover.

\section{Conclusion}\label{sec:conclusion}

In this paper, we measured how much of a reported AIS trajectory-prediction
error belongs to the model, using a leakage-aware protocol that separates
leakage from distribution shift on two corpora with different geography and
traffic. On both, the evaluation protocol and the problem formulation account
for more of the reported number than the architectural mechanisms do. A
best-of-16 oracle lowers TrAISformer's error by a factor of 2.1 to 3.2, about
three fifths of which a noise-perturbed constant-velocity predictor also obtains. Sharing
vessels between training and test lowers the error of absolute-position models
by 23--30\%, a gain that comes from the test vessels' own history, grows with
model size to 62--67\%, and largely disappears when only the position
representation is made relative. The same absolute-position models fail in new
waters, including held-out blocks whose position bins were all trained, while
relative and regression models degrade by a bounded amount. Graph attention over
surrounding traffic and frequency-aware attention give no measurable benefit,
an engineered waterway feature gives 12--22\%, and a waypoint graph helps
imputation mainly in the middle of a gap. Table~\ref{tbl:recommendations}
condenses these results into reporting practice. In future work, we plan to
extend the protocol across seasons and to further corpora, and to apply it to
generative and language-model predictors, for which the decoding and
coordinate-frame effects are likely to matter most. We release the manifests,
pipelines and all model implementations~\citep{raisi2026code}.

\begin{table*}
\caption{Recommended reporting practice for AIS trajectory prediction and imputation studies, with the evidence in this paper behind each item.}\label{tbl:recommendations}
\centering\footnotesize\setlength{\tabcolsep}{4pt}
\begin{tabular}{@{}P{0.16\textwidth}P{0.6\textwidth}P{0.16\textwidth}@{}}
\toprule
Aspect & Recommendation & Evidence \\
\midrule
Split & Use a vessel-disjoint manifest for any claim about new vessels. If a random or chronological split is used, state that vessels are shared and report the vessel-disjoint error beside it. Release the manifest. & Sections~\ref{sec:vesselleak}, \ref{sec:causal} \\
Chronological split & Separate subsets by at least one window span (purge) and state the gap between training and test periods. & Section~\ref{sec:timeleak} \\
Spatial transfer & Claims about new waters need a spatial hold-out: a cut and, for absolute-position models, held-out blocks whose position bins occur in training; report the share of unseen position bins. & Section~\ref{sec:shift} \\
Coordinate frame & State how position is represented. For absolute-position models, report the vessel-sharing gap and a spatial hold-out; do not read a random-split gain as generalization. & Sections~\ref{sec:capacity}, \ref{sec:shift-mechanism} \\
Decoding & Report single-shot error. A best-of-$N$ figure goes beside it, with $N$, and with a null-predictor control under the same oracle. & Section~\ref{sec:tfresult} \\
Fitted structures & Fit graphs, codebooks, clusterings and normalization statistics on training windows only, and state where they were fitted; the whole-data alternative costs one rerun to check. & Section~\ref{sec:imputeresult} \\
Context models & For models that read surrounding traffic, report the share of neighbors that are training vessels and use a density-matched control before attributing a change to leakage. & Section~\ref{sec:gat} \\
Metric & When comparing splits or corpora, normalize by a simple reference (constant velocity, linear interpolation) on the same test windows. & Section~\ref{sec:metrics} \\
Replication & Use at least three seeds and compare arms by seed-paired differences; mark which findings were measured on one corpus only. & Section~\ref{sec:metrics}, Table~\ref{tbl:evidence} \\
\bottomrule
\end{tabular}
\end{table*}

\appendix
\setcounter{equation}{0}\setcounter{table}{0}\setcounter{figure}{0}
\renewcommand{\theequation}{\Alph{section}.\arabic{equation}}
\renewcommand{\thetable}{\Alph{section}.\arabic{table}}
\renewcommand{\thefigure}{\Alph{section}.\arabic{figure}}
\makeatletter\@addtoreset{equation}{section}\@addtoreset{table}{section}\@addtoreset{figure}{section}\makeatother
\section{Comparator implementation details}\label{app:comparators}

This appendix states, per comparator, what the source specifies and what is
our reconstruction, in support of Section~\ref{sec:tfmethod}.

\textbf{TrAISformer.} Each observation is discretized at 0.01$^{\circ}$ in
latitude and longitude, 1~knot in speed and 5$^{\circ}$ in course, giving a
four-hot vector over 250, 270, 30 and 72 bins. The four one-hot components
are embedded into 256, 256, 128 and 128 dimensions and concatenated into a
768-wide embedding, consumed by a causal 8-layer, 8-head transformer with a
learned index positional encoding. Training minimizes the four next-step
cross-entropies plus the released multi-resolution term: each predicted
distribution is smoothed twice with a three-bin uniform kernel and the
negative smoothed probability of the true bin is added with weight~1. The
sampler restricts each step to the ten most likely bins of each attribute
and, for latitude and longitude, to a 40-bin vicinity of the previous
position. The default model has 57.4~M parameters. We sweep dropout, depth,
width and learning rate and report the variant best on validation, so the
baseline is measured at its best vessel-disjoint setting rather than its
default.

\textbf{AISFormer.} The article specifies the four-hot representation, the
eight-layer, eight-head backbone, a multi-resolution loss over coarse bins,
and an FFT-domain attention that splits low- and high-frequency components
at a radius. It does not say how that attention stays causal under
next-step training, on which patch layout the transform runs, the radius, or
the loss weight. We therefore keep everything else identical to the
TrAISformer comparator, use TrAISformer's smoothing term in place of the
coarse-bin term, and replace the attention with a causal analogue built on
the running spectrum of the prefix (Appendix~\ref{app:spectral}). The
default configuration has 16 modes, a 96-wide spectral projection and
63.3~M parameters. The sweep mirrors TrAISformer's and adds one over the
mode budget.

\textbf{GATransformer.} The article gives the architecture, positional
encoding, attention equations, loss and decoder, which we follow. The
spatial encoder applies a feed-forward augmentation, graph attention over
surrounding vessels, positional encoding and multi-head attention; the
temporal encoder omits the graph attention; the two outputs are
concatenated and decoded by linear layers. The source's two feature sets are
kept under its names, 4-Fs (kinematics) and 102-Fs (kinematics plus
distances to waterway intersections), although the kinematic input has five
channels here because course enters as sine and cosine.

Three things are ours. (i)~Neighbors are the eight nearest vessels within
10~km at each observed step, drawn from the same resampled dataset, and are
expressed in the target's local frame. About 38\% of slots are occupied on the test split, so
a vessel attends to about three neighbors on average and, in open water, to
none, in which case the attention reduces to the vessel's own features
through the self coefficient. (ii)~The waterway network follows the source's
procedure (rasterize at 100~m, threshold at 20 traffic units per cell,
skeletonize, keep skeleton cells with three or more neighbors, merge those
within 150~m), with one adaptation: on a ten-minute grid a vessel moves
about 3~km between reports, so each segment is sampled along its length to
keep the raster connected. It yields 1,681 crossings over a region 27 times
larger than the source's; the busiest 98 are kept so the feature has the
source's width. It is extracted from the training split alone.
(iii)~Hyper-parameters are chosen over six configurations spanning depth,
width, dropout, learning rate and graph-head count. At the source's dropout
of 0.5 and learning rate of $10^{-3}$, four of the six arms met non-finite
updates and the widest stopped learning after discarding 91\% of its
updates. The two arms that softened one of those settings trained without
such updates, and we select dropout 0.2. We attribute this to the dataset, not
the method: the source works at one-minute sampling over about
1,700~km$^2$, where a three-hour horizon covers far less open water.
Training discards any update with a non-finite loss or gradient; the count is
recorded per run, and a run that stopped learning is marked in the released
selection record and excluded from comparison.

\textbf{MGFormer.} The article specifies the loss, hyper-parameters, feature
set, decoder heads, and a Douglas--Peucker and DBSCAN waypoint graph with a
gated graph network. It does not say how the graph embedding enters the
sequence, and its physical features include draught, which the Danish feed
does not carry. Our reconstruction differs in three ways: waypoints are found
by $k$-means rather than Douglas--Peucker and DBSCAN; the physical features
are vessel length and type; and the embedding of each observed step is that
of its nearest waypoint, while masked steps receive a transfer context
pooled from the last waypoint before the gap and the first after it. The
graph has 256 waypoints connected where consecutive reports transition at
least five times (2,926 directed edges), and a three-step gated graph
network gives each waypoint its embedding. A bidirectional transformer
encoder feeds two heads, one for position and one for speed and course, and
a motion-consistency term ties the displacement between imputed positions to
the predicted speed and course. Our default set-up is tuned on validation.
We also run the article's own specification (two layers, two heads,
feed-forward width 1024, dropout 0.3, learning rate $10^{-3}$, weight decay
$5\times10^{-4}$, loss of five times the positional term plus speed and
periodic course terms, no consistency term), with and without the graph, at
batch size 64 for 60 epochs to fit our compute budget. Over three seeds it reaches
1.297~$\pm$~0.020~km with the graph and 1.371~$\pm$~0.010~km without, against
1.004 and 1.020~km for our default, so the tuned set-up is the fairer comparator and
the graph's contribution keeps its sign but is larger under the article's specification (5.7\% against 1.6\%).

\section{Causal spectral attention in the AISFormer comparator}\label{app:spectral}

This appendix gives the construction summarized in
Section~\ref{sec:tfmethod}. It is our reconstruction of the
frequency-aware attention AISFormer describes, not a transcription of the
source, which has no public code; the fidelity caveats of
Section~\ref{sec:tfmethod} apply to everything below.

A frequency-enhanced attention block of the usual kind takes the discrete
Fourier transform of the whole input sequence. That is unusable here: the
comparator is trained with next-step cross-entropy at every position, so a
whole-sequence transform at position $t$ would read the targets at
positions after $t$ and the reported error would be meaningless. We use
the causal analogue.

Each block first splits its input with a left-padded moving average into a
low-pass trend and a residual, then processes the residual along two
paths. The first is ordinary causal self-attention. The second is a
spectral path that maintains the running spectrum of the prefix,
\begin{equation}
X_m(t) = \frac{1}{t}\sum_{s\le t} x_s \exp(-2\pi i m s / P),
\label{eq:prefixspec}
\end{equation}
over a fixed period $P$ and a selected mode set $M$. Because
Eq.~\eqref{eq:prefixspec} is a cumulative sum, all positions are computed in
one pass rather than one per step. Each retained mode receives a learned
complex channel-wise gain and is mapped back to the time domain.

Two properties are worth stating explicitly, since they are what make the
comparator a fair rendering rather than an approximation of convenience.
First, the value at step $t$ depends only on steps up to $t$; we verify
this numerically by perturbing future positions and confirming the output
at $t$ is unchanged. Second, the mode set still spans the spectrum, with
the low modes carrying the cruise component and the high modes the
maneuver component, which is the separation the source describes as the
mechanism's purpose. Modes are selected deterministically (the lowest
half of the budget, then evenly spaced across the remainder) rather
than at random, so the comparator is reproducible across seeds.

\section{Supplementary tables}\label{app:supp}

\begin{table*}
\caption{Size of every prediction split manifest: training, validation and test windows, the number of test vessels, and the share of those vessels that also appear in the training windows. All manifests are cut from one dense window pool per corpus; validation and test keep every third window of a voyage, so no two evaluated contexts share a report. Region manifests drop every window with steps on both sides of a boundary.}\label{tbl:splitcounts}
\centering\small
\begin{tabular}{@{}lrrrrrrrr@{}}
\toprule
 & \multicolumn{4}{c}{Danish (DMA)} & \multicolumn{4}{c}{US (NOAA)} \\
\cmidrule(lr){2-5}\cmidrule(lr){6-9}
Manifest & Train & Val. & Test & Test vessels (shared) & Train & Val. & Test & Test vessels (shared) \\
\midrule
Random window & 46,027 & 3,676 & 3,757 & 1,679 (98\%) & 30,833 & 2,631 & 2,689 & 1,134 (97\%) \\
Chronological & 46,016 & 3,757 & 3,781 & 970 (55\%) & 30,828 & 2,678 & 2,585 & 705 (69\%) \\
Chronological, purged & 45,548 & 3,568 & 3,781 & 970 (55\%) & 30,431 & 2,474 & 2,585 & 705 (68\%) \\
Vessel-disjoint & 46,168 & 3,887 & 3,528 & 465 (0\%) & 30,207 & 2,408 & 3,147 & 325 (0\%) \\
Region, longitude cut & 17,750 & 1,193 & 8,185 & 1,724 (39\%) & 14,732 & 1,048 & 6,541 & 1,301 (51\%) \\
Region, latitude cut & 19,083 & 1,327 & 8,728 & 2,493 (65\%) & 13,477 & 987 & 6,003 & 889 (37\%) \\
Region, spatial blocks & 27,256 & 1,887 & 2,363 & 255 (49\%) & 15,725 & 1,102 & 1,813 & 227 (41\%) \\
\bottomrule
\end{tabular}
\end{table*}

\begin{table*}
\caption{Capacity sweep. TrAISformer with absolute or relative position tokens at five sizes, one-hour error as a ratio to constant velocity on the vessel-disjoint (V) and random-window (R) manifests, three seeds per point. \emph{Gap} = (V $-$ R)/V. The slope is that of the per-seed greedy gap (V $-$ R) on $\log_{10}$ parameters, with a 95\% seed-bootstrap interval.}\label{tbl:capacity}
\centering\small
\begin{tabular}{@{}llrrrrrrr@{}}
\toprule
Corpus & Tokens & Params (M) & Greedy V & Greedy R & Gap & Bo16 V & Bo16 R & Gap \\
\midrule
Danish & absolute & 1.1 & 0.682 & 0.610 & 11\% & 0.291 & 0.245 & 16\% \\
Danish & absolute & 3.2 & 0.636 & 0.535 & 16\% & 0.285 & 0.219 & 23\% \\
Danish & absolute & 7.4 & 0.627 & 0.484 & 23\% & 0.295 & 0.196 & 33\% \\
Danish & absolute & 16.4 & 0.647 & 0.388 & 40\% & 0.299 & 0.168 & 44\% \\
Danish & absolute & 57.4 & 0.647 & 0.216 & 67\% & 0.304 & 0.132 & 57\% \\
\addlinespace
Danish & relative & 1.2 & 0.809 & 0.756 & 6\% & 0.370 & 0.349 & 6\% \\
Danish & relative & 3.4 & 0.755 & 0.691 & 8\% & 0.333 & 0.320 & 4\% \\
Danish & relative & 7.6 & 0.742 & 0.676 & 9\% & 0.324 & 0.314 & 3\% \\
Danish & relative & 16.7 & 0.691 & 0.654 & 5\% & 0.311 & 0.300 & 4\% \\
Danish & relative & 57.7 & 0.670 & 0.628 & 6\% & 0.315 & 0.294 & 7\% \\

\midrule
US & absolute & 1.1 & 0.592 & 0.535 & 10\% & 0.241 & 0.203 & 16\% \\
US & absolute & 3.2 & 0.586 & 0.467 & 20\% & 0.235 & 0.182 & 22\% \\
US & absolute & 7.4 & 0.596 & 0.433 & 27\% & 0.242 & 0.172 & 29\% \\
US & absolute & 16.4 & 0.595 & 0.363 & 39\% & 0.244 & 0.151 & 38\% \\
US & absolute & 57.4 & 0.598 & 0.226 & 62\% & 0.252 & 0.131 & 48\% \\
\addlinespace
US & relative & 1.2 & 0.679 & 0.692 & $-$2\% & 0.298 & 0.285 & 4\% \\
US & relative & 3.4 & 0.617 & 0.604 & 2\% & 0.267 & 0.257 & 4\% \\
US & relative & 7.6 & 0.596 & 0.577 & 3\% & 0.257 & 0.242 & 6\% \\
US & relative & 16.7 & 0.578 & 0.551 & 5\% & 0.254 & 0.235 & 8\% \\
US & relative & 57.7 & 0.560 & 0.526 & 6\% & 0.261 & 0.239 & 8\% \\

\bottomrule
\end{tabular}
\par\smallskip\footnotesize Danish, absolute: slope 0.212 per decade [0.160, 0.263], Spearman $\rho=0.98$; Danish, relative: slope $-$0.011 per decade [$-$0.023, $-$0.003], Spearman $\rho=-0.51$; Danish, absolute $-$ relative: 0.223 [0.168, 0.286]; US, absolute: slope 0.180 per decade [0.150, 0.203], Spearman $\rho=0.98$; US, relative: slope 0.027 per decade [0.018, 0.035], Spearman $\rho=0.95$; US, absolute $-$ relative: 0.153 [0.116, 0.183].
\end{table*}

\begin{table}
\caption{Vessel-type hold-out on the Danish corpus (one-hour single-shot error, km). Both arms are tested on the same held-out vessels of one type; \emph{held out} trains without that type, the \emph{control} on an equal-size random subset that includes it. Ratio = held out / control, per seed, mean over three seeds.}\label{tbl:type}
\centering\footnotesize\setlength{\tabcolsep}{3pt}
\begin{tabular}{@{}llrrrr@{}}
\toprule
Type & Model & $n$ & Held out & Control & Ratio \\
\midrule
Fishing & TrAISformer & 723 & 3.76 & 2.70 & 1.39 \\
 & TrAISformer, rel. &  & 2.84 & 2.46 & 1.15 \\
 & Compact encoder &  & 3.10 & 2.46 & 1.26 \\
\addlinespace
Passenger & TrAISformer & 493 & 8.03 & 2.62 & 3.07 \\
 & TrAISformer, rel. &  & 6.39 & 3.59 & 1.79 \\
 & Compact encoder &  & 5.86 & 4.13 & 1.42 \\
\addlinespace
Tanker & TrAISformer & 547 & 2.05 & 1.96 & 1.04 \\
 & TrAISformer, rel. &  & 2.62 & 2.51 & 1.05 \\
 & Compact encoder &  & 3.23 & 3.03 & 1.07 \\
\bottomrule
\end{tabular}
\end{table}

This appendix collects the detailed tables behind Sections~\ref{sec:framework}, \ref{sec:tfresult}, \ref{sec:capacity} and~\ref{sec:shift-type}: the sizes of all split manifests, the full capacity sweep, the vessel-type hold-out, and the TrAISformer decoder and loss ablation.

The released TrAISformer code also differs from plain cross-entropy training in
two details, a multi-resolution (blur) loss and a decoder that restricts each
step to the top-10 logits and a 40-bin vicinity of the previous position.
Table~\ref{tbl:protocolabl} adds them one at a time. The decoder restriction is
what moves the deterministic numbers: added to plain cross-entropy, it lowers
greedy error from 2.53 to 2.27~km and a single random draw from 3.35 to
2.50~km, while the blur loss alone changes
greedy error by less than its seed spread. The best-of-16 figure hardly moves
across the four rows (0.98 to 1.05~km), so the published one-hour number does
not depend on these details, but the oracle factor does, from 2.6 for plain
cross-entropy to 2.1 for the full released protocol. We report the full
protocol throughout.

\begin{table*}
\caption{Which released TrAISformer detail moves the number. The 7.4~M model on the vessel-disjoint split, one-hour mean displacement error (km), mean\,$\pm$\,SD over three seeds. The blur loss is the released multi-resolution cross-entropy (two smoothing passes, weight 1); the decoder restricts each step to the top-10 logits and, for latitude and longitude, to a 40-bin vicinity of the previous position. Oracle factor is greedy over best-of-16.}\label{tbl:protocolabl}
\begin{tabular*}{\tblwidth}{@{} L R R R R@{}}
\toprule
Configuration & Greedy & Single draw & Best-of-16 & Oracle factor \\
\midrule
Plain cross-entropy, unrestricted sampling & 2.53\,$\pm$\,0.23 & 3.35\,$\pm$\,0.08 & 0.977\,$\pm$\,0.013 & 2.59 \\
+ multi-resolution (blur) loss & 2.45\,$\pm$\,0.12 & 3.25\,$\pm$\,0.08 & 1.005\,$\pm$\,0.023 & 2.43 \\
+ vicinity / top-$k$ decoder & 2.27\,$\pm$\,0.01 & 2.50\,$\pm$\,0.02 & 1.012\,$\pm$\,0.025 & 2.24 \\
Both (published protocol) & 2.23\,$\pm$\,0.04 & 2.49\,$\pm$\,0.05 & 1.050\,$\pm$\,0.007 & 2.12 \\
\bottomrule
\end{tabular*}
\end{table*}

\section{Decoding, interaction, training and error analyses}\label{app:r2}

This appendix examines the decoding of the token models, GATransformer under
vessel encounters, the training curves of the capacity runs, the error by
maneuver class and further error metrics. The token models and GATransformer
were retrained with the configurations of the paper (three seeds; six for
Danish GATransformer) to save per-window outputs (and, for the token models,
checkpoints); at every
horizon their errors are within 2.3\% (0.14~km) of the runs reported in the
main text.
The other models use the per-window outputs of the original runs.

\subsection{Teacher-forced and free-running decoding}\label{app:r2-decoding}

\begin{figure*}
\centering
\includegraphics[width=\textwidth]{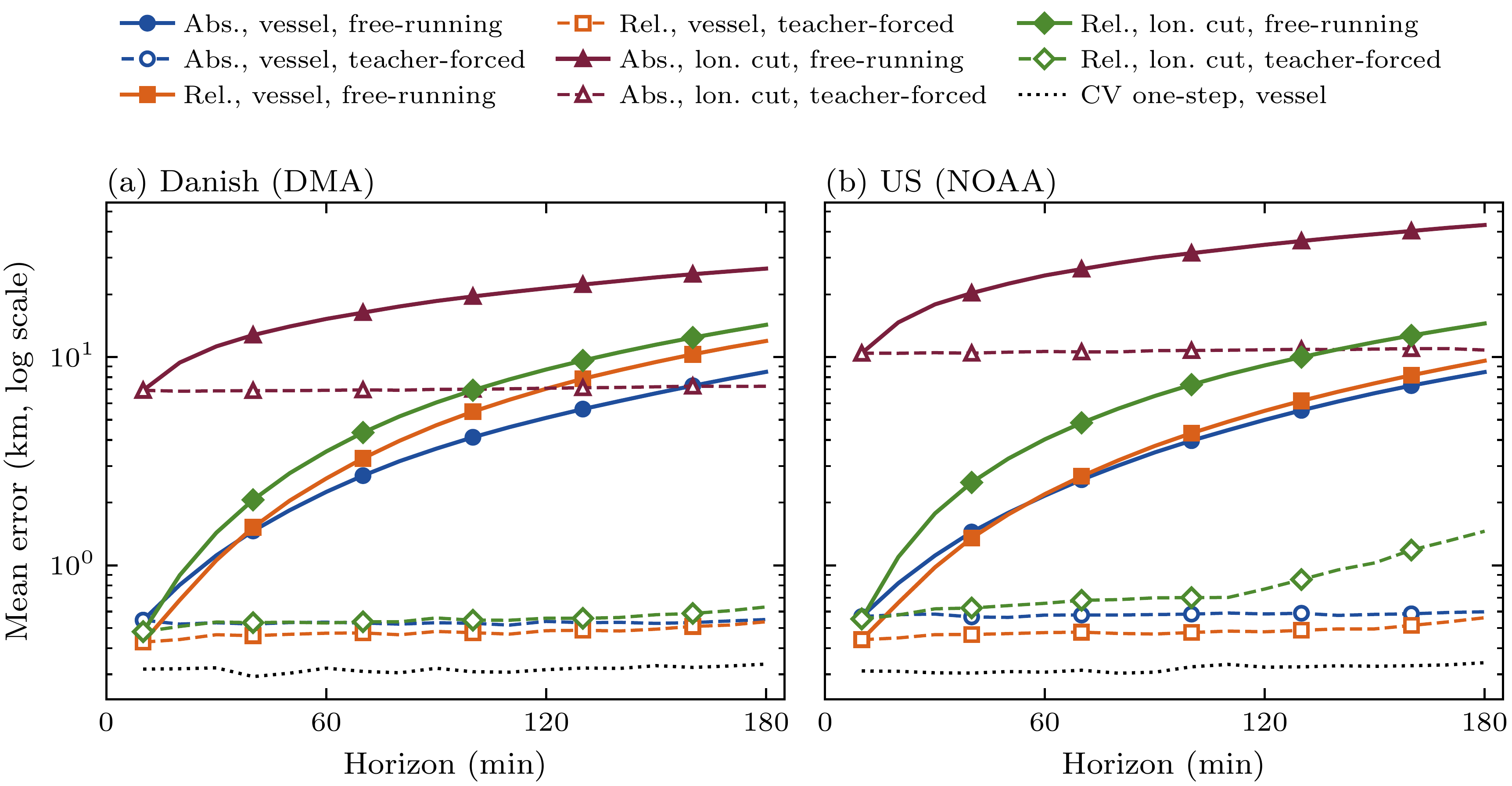}
\caption{Mean error at each 10-minute step of the horizon, free-running (filled
marks, solid) against teacher-forced (hollow marks, dashed), for TrAISformer
with absolute and relative tokens on the vessel-disjoint manifest and under the
longitude cut. Dotted: constant velocity from the true previous state. Mean over
three seeds; values at one and three hours in Table~\ref{tbl:r2decoding}.}
\label{fig:r2decoding}
\end{figure*}

Table~\ref{tbl:r2decoding} and Fig.~\ref{fig:r2decoding} compare the one-step
error of the token models with their free-running error. On the
vessel-disjoint manifest, one 10-minute step predicted from the true history
has a mean error of 0.47--0.60~km, 1.5--1.9 times constant velocity from the
true previous state and 0.15--0.24~km above the error of quantization alone
(0.32--0.41~km). The free-running error grows from there to 8.5--11.9~km at
three hours, 14--22 times the one-step error; constant velocity grows
45--49-fold over the same horizon. On this manifest, most of the three-hour
error therefore arises over the horizon rather than in a single step. Teacher
forcing removes both the compounding of the model's own errors and the
uncertainty of the vessel's future, so this comparison does not separate the
two. Under the longitude cut, relative tokens keep a one-step error of
0.53--0.66~km at one hour (at most 1.46~km at three hours). Given the true
history, absolute TrAISformer's one-step error at one hour rises to 6.9~km
(Danish) and 10.6~km (US), 30--38 times constant velocity: it fails at the
first step in unseen waters, before any rollout.

\begin{table*}
\caption{Teacher-forced against free-running error of the token models (km). \emph{Teacher-forced}: each step predicted greedily from the true history up to the previous step (one-step error, reported at the one- and three-hour steps); \emph{CV one-step}: constant velocity from the true previous state. \emph{Free-running}: the greedy rollout reported in the paper. \emph{Beam}: beam search over the joint four-attribute token (8 beams, three bins per attribute); \emph{likelier}: share of windows where it finds a sequence of higher model log-probability than greedy. Vessel: vessel-disjoint manifest; Lon.\ cut: region hold-out of Section~\ref{sec:shift}. Mean over three seeds (seed standard deviation in small type). Encoding and decoding the true position gives a mean error of 0.32--0.41~km (Section~\ref{sec:relcontrols}).}\label{tbl:r2decoding}
\centering\footnotesize\setlength{\tabcolsep}{3.2pt}
\begin{tabular}{@{}llrrrrrrrr@{}}
\toprule
 & & \multicolumn{3}{c}{One step, true history} & \multicolumn{2}{c}{Free-running} & \multicolumn{3}{c}{Beam search} \\
\cmidrule(lr){3-5}\cmidrule(lr){6-7}\cmidrule(lr){8-10}
Model & Manifest & TF 1~h & TF 3~h & CV 1~h & 1~h & 3~h & 1~h & 3~h & Likelier \\
\midrule
\multicolumn{10}{@{}l}{\emph{Danish (DMA)}} \\
TrAISformer & Vessel & 0.53{\scriptsize$\pm$0.00} & 0.55{\scriptsize$\pm$0.00} & 0.32 & 2.25{\scriptsize$\pm$0.02} & 8.49{\scriptsize$\pm$0.05} & 2.27{\scriptsize$\pm$0.01} & 8.57{\scriptsize$\pm$0.09} & 92\% \\
AISFormer-inspired & Vessel & 0.53{\scriptsize$\pm$0.00} & 0.55{\scriptsize$\pm$0.01} & 0.32 & 2.26{\scriptsize$\pm$0.04} & 8.85{\scriptsize$\pm$0.14} & 2.29{\scriptsize$\pm$0.03} & 9.00{\scriptsize$\pm$0.10} & 93\% \\
TrAISformer, rel. & Vessel & 0.47{\scriptsize$\pm$0.00} & 0.54{\scriptsize$\pm$0.00} & 0.32 & 2.61{\scriptsize$\pm$0.03} & 11.94{\scriptsize$\pm$0.02} & 2.66{\scriptsize$\pm$0.00} & 11.87{\scriptsize$\pm$0.06} & 90\% \\
TrAISformer & Lon.\ cut & 6.92{\scriptsize$\pm$0.25} & 7.24{\scriptsize$\pm$0.18} & 0.22 & 15.24{\scriptsize$\pm$0.25} & 26.57{\scriptsize$\pm$0.59} & 18.13{\scriptsize$\pm$0.51} & 31.29{\scriptsize$\pm$0.82} & 93\% \\
TrAISformer, rel. & Lon.\ cut & 0.53{\scriptsize$\pm$0.00} & 0.63{\scriptsize$\pm$0.00} & 0.22 & 3.52{\scriptsize$\pm$0.08} & 14.28{\scriptsize$\pm$0.26} & 3.74{\scriptsize$\pm$0.13} & 14.73{\scriptsize$\pm$0.30} & 91\% \\
\addlinespace
\multicolumn{10}{@{}l}{\emph{US (NOAA)}} \\
TrAISformer & Vessel & 0.58{\scriptsize$\pm$0.00} & 0.60{\scriptsize$\pm$0.01} & 0.31 & 2.16{\scriptsize$\pm$0.04} & 8.47{\scriptsize$\pm$0.08} & 2.20{\scriptsize$\pm$0.02} & 8.53{\scriptsize$\pm$0.09} & 90\% \\
AISFormer-inspired & Vessel & 0.57{\scriptsize$\pm$0.01} & 0.59{\scriptsize$\pm$0.01} & 0.31 & 2.20{\scriptsize$\pm$0.02} & 8.79{\scriptsize$\pm$0.08} & 2.21{\scriptsize$\pm$0.02} & 8.76{\scriptsize$\pm$0.04} & 89\% \\
TrAISformer, rel. & Vessel & 0.48{\scriptsize$\pm$0.00} & 0.56{\scriptsize$\pm$0.01} & 0.31 & 2.20{\scriptsize$\pm$0.02} & 9.60{\scriptsize$\pm$0.15} & 2.19{\scriptsize$\pm$0.03} & 9.56{\scriptsize$\pm$0.14} & 91\% \\
TrAISformer & Lon.\ cut & 10.63{\scriptsize$\pm$0.20} & 10.80{\scriptsize$\pm$0.17} & 0.28 & 24.64{\scriptsize$\pm$0.78} & 43.02{\scriptsize$\pm$1.46} & 29.66{\scriptsize$\pm$0.72} & 47.73{\scriptsize$\pm$1.09} & 89\% \\
TrAISformer, rel. & Lon.\ cut & 0.66{\scriptsize$\pm$0.01} & 1.46{\scriptsize$\pm$0.07} & 0.28 & 4.03{\scriptsize$\pm$0.13} & 14.49{\scriptsize$\pm$0.33} & 4.08{\scriptsize$\pm$0.08} & 14.47{\scriptsize$\pm$0.26} & 92\% \\
\bottomrule
\end{tabular}
\end{table*}

Greedy decoding chooses the most probable next report under the model, within
the vicinity mask, which need not be the most probable trajectory. Beam search
over the joint four-attribute token finds a sequence of higher model
probability in 89--93\% of windows, but these sequences are not more accurate.
On the vessel-disjoint manifest its error is within 0.15~km of greedy at every
horizon, and under the longitude cut it is worse for absolute tokens (18.1
against 15.2~km at one hour, Danish). This approximate search therefore gives
no evidence that greedy decoding understates the accuracy of these models.

Table~\ref{tbl:r2kinematics} checks the decoded trajectories for physical
consistency. On the vessel-disjoint manifest, for every model the median speed
and course mismatches of greedy and beam trajectories are no larger than those
of the quantized truth: the median speed mismatch is 0.52--0.97 against
0.79--1.00~kn, and the median
course mismatch 3.9--8.8 against 6.0--10.0$^\circ$. At most 0.03\% of greedy
steps imply more than 40 knots. A single
random draw is less consistent. Under the longitude cut, absolute-token
rollouts are not consistent: the median course mismatch is 52--55$^\circ$, and on
the US corpus 20\% of steps imply more than 40 knots.

\begin{table*}
\caption{Kinematic consistency of decoded trajectories. For every pair of consecutive predicted steps, the speed implied by the two positions is compared with the predicted SOG and the bearing of the displacement with the predicted COG (medians over all transitions; course only for displacements above 0.5~km). Because positions are bin centers, the true trajectory itself shows a mismatch once quantized, so each decoder is compared with the \emph{quantized truth} of the same representation and manifest (the unquantized truth gives 0.07--0.11~kn and 0.8--1.6$^\circ$). \emph{$>$40~kn}: share of transitions whose implied speed exceeds 40 knots (greedy). Mean over three seeds.}\label{tbl:r2kinematics}
\centering\footnotesize\setlength{\tabcolsep}{3.2pt}
\begin{tabular}{@{}llrrrrrrrrr@{}}
\toprule
 & & \multicolumn{4}{c}{Speed mismatch (kn)} & \multicolumn{4}{c}{Course mismatch (deg)} & \\
\cmidrule(lr){3-6}\cmidrule(lr){7-10}
Model & Manifest & Quant. & Greedy & Beam & Draw & Quant. & Greedy & Beam & Draw & $>$40~kn \\
\midrule
\multicolumn{11}{@{}l}{\emph{Danish (DMA)}} \\
TrAISformer & Vessel & 1.00 & 0.95 & 0.90 & 1.02 & 6.0 & 5.1 & 4.8 & 6.9 & 0.01\% \\
AISFormer-inspired & Vessel & 1.00 & 0.97 & 0.92 & 1.05 & 6.0 & 5.0 & 4.7 & 6.9 & 0.02\% \\
TrAISformer, rel. & Vessel & 0.79 & 0.68 & 0.52 & 0.83 & 6.2 & 4.4 & 3.9 & 6.6 & 0.00\% \\
TrAISformer & Lon.\ cut & 1.00 & 4.85 & 4.17 & 10.36 & 7.5 & 54.9 & 51.6 & 61.5 & 0.47\% \\
TrAISformer, rel. & Lon.\ cut & 0.76 & 0.50 & 0.50 & 0.84 & 7.5 & 6.2 & 3.9 & 9.4 & 0.00\% \\
\addlinespace
\multicolumn{11}{@{}l}{\emph{US (NOAA)}} \\
TrAISformer & Vessel & 1.00 & 0.75 & 0.74 & 1.21 & 10.0 & 8.6 & 8.5 & 10.8 & 0.02\% \\
AISFormer-inspired & Vessel & 1.00 & 0.76 & 0.74 & 1.18 & 10.0 & 8.8 & 8.5 & 10.2 & 0.03\% \\
TrAISformer, rel. & Vessel & 0.82 & 0.68 & 0.68 & 0.83 & 7.6 & 5.8 & 5.1 & 8.3 & 0.00\% \\
TrAISformer & Lon.\ cut & 0.94 & 4.64 & 4.23 & 17.79 & 8.7 & 51.7 & 44.2 & 60.1 & 19.91\% \\
TrAISformer, rel. & Lon.\ cut & 0.82 & 0.70 & 0.69 & 0.89 & 7.5 & 6.6 & 6.0 & 11.8 & 0.03\% \\
\bottomrule
\end{tabular}
\end{table*}

\subsection{Interaction classes and the sampling interval}\label{app:r2-encounter}

Table~\ref{tbl:r2encounter} splits the vessel-disjoint test windows by the
interaction at the forecast origin, using the neighbors the model reads. The
encounter classes are operational, based on the course difference alone; a full
COLREG classification would also use the relative bearing. Standard errors are
those of the per-seed paired difference (six seeds for the Danish corpus, three
for the US one). On the Danish corpus the spatial encoder makes no detectable
difference in any class: every difference is within 1.4 standard errors of
zero. On the US corpus, removing it lowers ADE by 0.18--0.21~km (3.8--5.9\%) in
co-present, head-on and crossing windows, at 3.5 or more standard errors; the
isolated and overtaking differences are not resolved. A wider encounter
definition (2~nmi, 60~minutes) gives the same pattern. At this sampling interval
the encoder thus gives no detectable improvement in any class, including the
encounters it is designed for. Over all windows the reruns reproduce the sign of the ablation of
Tables~\ref{tbl:gatablation} and~\ref{tbl:gatablation-noaa}, with removal
changing ADE by $-0.7\%$ (Danish) and $-4.4\%$ (US), against $-2.1\%$ and
$-2.7\%$ there.

Encounters are short at this sampling interval. The median time to the closest
point of approach is 7.9 (Danish) and 7.8~minutes (US), and the close-quarters
phase, the time the two vessels would spend within 1~nmi if both kept their
speed and course, lasts a median of 8.8 and 10.9~minutes. In 56\% and 40\% of encounters it is shorter than one 10-minute
step, so the model sees at most one report inside it.

\begin{table*}
\caption{GATransformer with and without its spatial encoder, by interaction class of the test window at the forecast origin (vessel-disjoint manifest). \emph{Isolated}: no vessel within 10~km; \emph{co-present}: neighbors but no encounter; an \emph{encounter} has both vessels at 2~knots or more and a constant-velocity closest point of approach within 1~nmi and 30~minutes, classed by course difference alone as head-on ($\geq157.5^\circ$), overtaking ($\leq67.5^\circ$) or crossing. ADE over the three-hour horizon (km), mean over six (Danish) and three (US) seeds; $\Delta$: full minus no-spatial-encoder, $\pm$ standard error of the per-seed paired difference (positive: the encoder hurts).}\label{tbl:r2encounter}
\centering\footnotesize\setlength{\tabcolsep}{4pt}
\begin{tabular}{@{}lrrrr@{\hspace{1.4em}}rrrr@{}}
\toprule
 & \multicolumn{4}{c}{Danish (DMA)} & \multicolumn{4}{c}{US (NOAA)} \\
\cmidrule(lr){2-5}\cmidrule(lr){6-9}
Class & $n$ & Full & No enc. & $\Delta$ & $n$ & Full & No enc. & $\Delta$ \\
\midrule
Isolated & 552 & 4.24 & 4.15 & $+0.08\pm0.06$ & 334 & 5.18 & 5.02 & $+0.16\pm0.12$ \\
Co-present & 2{,}193 & 3.66 & 3.65 & $+0.01\pm0.02$ & 1{,}397 & 4.20 & 3.99 & $+0.21\pm0.02$ \\
Head-on & 283 & 4.15 & 4.15 & $+0.00\pm0.05$ & 870 & 3.23 & 3.05 & $+0.18\pm0.04$ \\
Crossing & 213 & 4.32 & 4.29 & $+0.04\pm0.05$ & 283 & 5.11 & 4.92 & $+0.19\pm0.05$ \\
Overtaking & 287 & 3.81 & 3.76 & $+0.05\pm0.04$ & 263 & 4.13 & 4.11 & $+0.02\pm0.06$ \\
\bottomrule
\end{tabular}
\end{table*}

\subsection{Training curves across model sizes}\label{app:r2-capacity}

\begin{figure*}
\centering
\includegraphics[width=\textwidth]{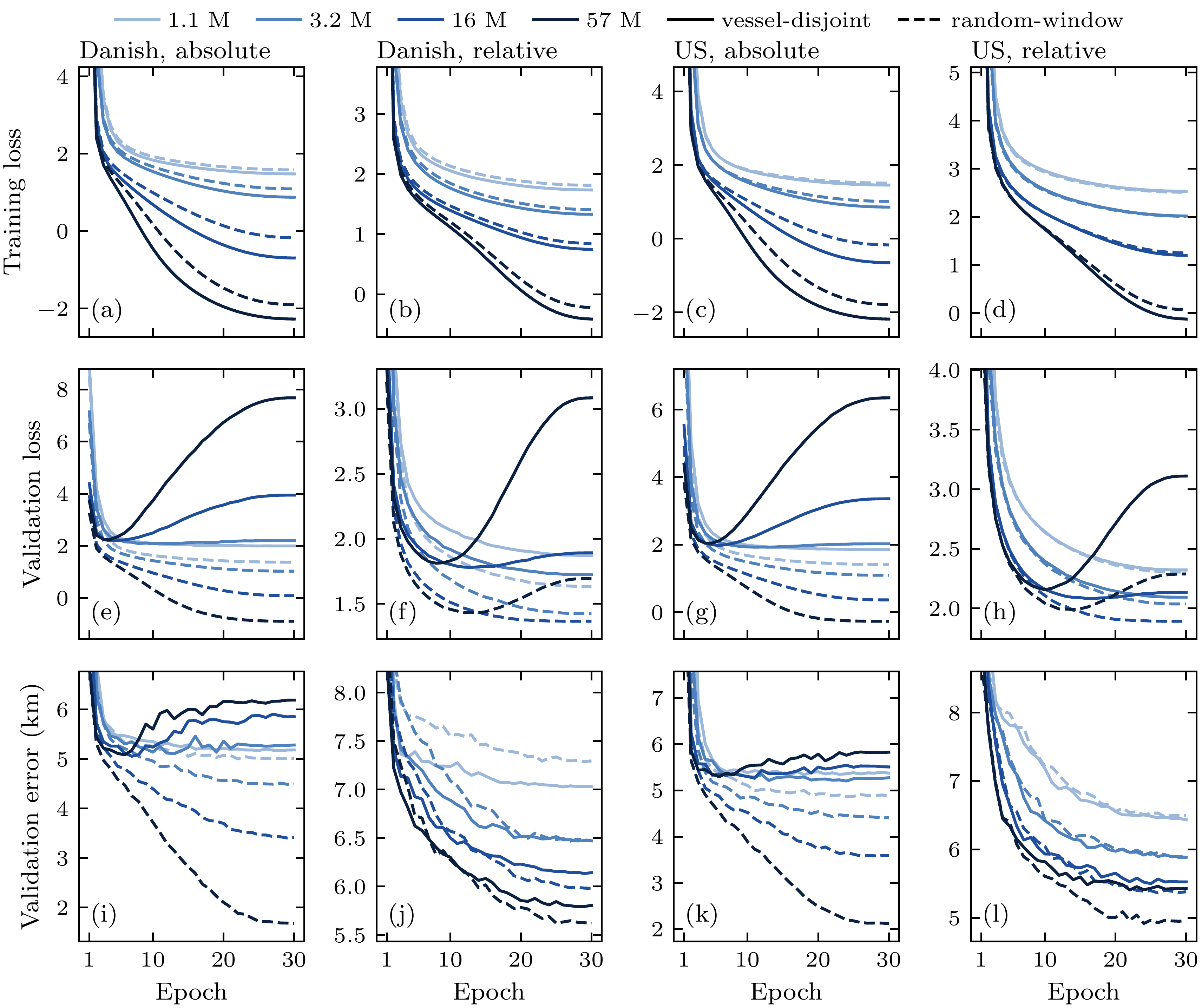}
\caption{Training and validation curves of the capacity runs of
Section~\ref{sec:capacity} for the vessel-disjoint (solid) and random-window
(dashed) manifests, mean over three seeds. Loss: the training objective, the
four cross-entropies plus the released multi-resolution (blur) terms, which are
negative by construction, so the loss can fall below zero. Validation error:
greedy error averaged over the one-, two- and three-hour horizons on the
validation windows, the checkpoint-selection score. Axes are scaled to epochs
3--30.}\label{fig:r2capacity}
\end{figure*}

Fig.~\ref{fig:r2capacity} shows the curves of the capacity runs of
Section~\ref{sec:capacity}. Epochs are numbered 1--30, and the cosine
learning-rate schedule decays to zero at epoch 30. The 16 and 57~M
absolute-token models on the vessel-disjoint manifest are selected at epochs
5--11. After that, their validation loss rises (to 7.7 on the Danish and 6.4 on
the US corpus at 57~M) while their training loss keeps falling, and their
last-epoch validation error is 0.3--1.2~km above the best. The 3.2~M
absolute-token model is selected at epochs 10--24. The 1.1~M models and all
random-window runs are selected at epochs 16--30 and end within 0.09~km of their
best validation error. Not all of them have converged: on the random-window
manifest the 16 and 57~M absolute-token models still improve by 0.14--0.43~km
between epochs 20 and 30. A longer schedule could lower their random-window
error further, which would widen, not close, the gap measured in
Section~\ref{sec:capacity}. With relative tokens, the selected epochs are 19--30
at every size, and the vessel-disjoint and random-window validation errors stay
within 0.5~km of each other. The validation loss of the 57~M relative model
reaches its minimum at epochs 8--14 and then rises, but its validation error
does not.

\subsection{Maneuver classes}\label{app:r2-maneuver}

Table~\ref{tbl:r2maneuver} labels each window by its true three-hour future.
Every model gains most over constant velocity on turning windows (ratios
0.51--0.76), where extrapolating the current course fails. The models that read
displacements or relative positions also gain on slow windows (0.41--0.60). On
course-keeping and speed-changing windows, all models are close to constant
velocity or worse (0.90--1.74). The ranking changes between classes: at one hour
the absolute-token models lead on turning windows (tied with GATransformer on
the Danish corpus), and the LSTM and the compact encoder on slow ones. With
looser ($20^\circ$, 2~kn) or stricter ($45^\circ$, 5~kn) thresholds, the turning
ratios stay within 0.50--0.78. The course-keeping ratios depend on the threshold
(0.81--2.67), because a looser turning threshold leaves only near-straight
tracks in that class, where constant velocity is hard to beat.

\begin{table*}
\caption{One-hour single-shot error by maneuver class of the true three-hour future, as a ratio to constant velocity on the same windows (vessel-disjoint manifest). Labels are assigned in this order. \emph{Slow}: median speed over the forecast origin and the future below 2~knots; \emph{Turn}: at some future step with a speed of 2~knots or more, the course differs from that at the origin by $30^\circ$ or more; \emph{Speed}: at some future step, the speed differs from that at the origin by 3~knots or more; \emph{Keep}: the rest. Window counts in the header; constant velocity itself (km) in the last row. Mean over seeds (seed standard deviation in small type).}\label{tbl:r2maneuver}
\centering\footnotesize\setlength{\tabcolsep}{3pt}
\begin{tabular}{@{}lrrrr@{\hspace{1.2em}}rrrr@{}}
\toprule
 & \multicolumn{4}{c}{Danish (DMA)} & \multicolumn{4}{c}{US (NOAA)} \\
\cmidrule(lr){2-5}\cmidrule(lr){6-9}
Model & Keep (1{,}008) & Turn (2{,}140) & Speed (114) & Slow (266) & Keep (405) & Turn (2{,}340) & Speed (121) & Slow (281) \\
\midrule
TrAISformer & 0.91{\scriptsize$\pm$0.00} & 0.58{\scriptsize$\pm$0.00} & 1.07{\scriptsize$\pm$0.03} & 0.67{\scriptsize$\pm$0.03} & 1.38{\scriptsize$\pm$0.04} & 0.51{\scriptsize$\pm$0.00} & 0.97{\scriptsize$\pm$0.06} & 0.91{\scriptsize$\pm$0.19} \\
AISFormer-inspired & 0.90{\scriptsize$\pm$0.05} & 0.58{\scriptsize$\pm$0.00} & 1.07{\scriptsize$\pm$0.03} & 0.65{\scriptsize$\pm$0.03} & 1.35{\scriptsize$\pm$0.09} & 0.51{\scriptsize$\pm$0.01} & 0.99{\scriptsize$\pm$0.03} & 1.07{\scriptsize$\pm$0.07} \\
TrAISformer, rel. & 0.91{\scriptsize$\pm$0.01} & 0.71{\scriptsize$\pm$0.01} & 1.02{\scriptsize$\pm$0.01} & 0.60{\scriptsize$\pm$0.03} & 1.13{\scriptsize$\pm$0.05} & 0.57{\scriptsize$\pm$0.01} & 0.93{\scriptsize$\pm$0.04} & 0.49{\scriptsize$\pm$0.03} \\
GATransformer & 1.10{\scriptsize$\pm$0.03} & 0.58{\scriptsize$\pm$0.01} & 1.08{\scriptsize$\pm$0.05} & 0.51{\scriptsize$\pm$0.04} & 1.73{\scriptsize$\pm$0.07} & 0.62{\scriptsize$\pm$0.01} & 1.10{\scriptsize$\pm$0.03} & 0.54{\scriptsize$\pm$0.03} \\
LSTM on displacements & 0.92{\scriptsize$\pm$0.02} & 0.66{\scriptsize$\pm$0.00} & 1.01{\scriptsize$\pm$0.01} & 0.41{\scriptsize$\pm$0.01} & 1.18{\scriptsize$\pm$0.01} & 0.56{\scriptsize$\pm$0.00} & 0.95{\scriptsize$\pm$0.00} & 0.45{\scriptsize$\pm$0.02} \\
Compact encoder & 1.10{\scriptsize$\pm$0.02} & 0.76{\scriptsize$\pm$0.02} & 1.05{\scriptsize$\pm$0.01} & 0.47{\scriptsize$\pm$0.02} & 1.74{\scriptsize$\pm$0.09} & 0.73{\scriptsize$\pm$0.01} & 1.15{\scriptsize$\pm$0.11} & 0.44{\scriptsize$\pm$0.02} \\
\midrule
Constant velocity (km) & 1.49 & 4.68 & 2.39 & 2.90 & 1.21 & 4.30 & 2.12 & 2.83 \\
\bottomrule
\end{tabular}
\end{table*}

\subsection{Further error metrics}\label{app:r2-metrics}

Table~\ref{tbl:r2metrics} adds the root mean square, median and 90th percentile
of the error, ADE and FDE. Rankings depend on the metric. The absolute-token
models have the lowest median three-hour error on both corpora. Their errors
are heavy-tailed, however (three-hour RMSE 1.7--1.8 times the MAE), and
GATransformer has the lowest three-hour RMSE and 90th percentile on both
corpora. These metrics qualify the ranking between models. The paper's leakage
and shift analyses compare each model with itself across manifests under the
mean error; we did not repeat them under the other metrics. We do not report MAPE:
a percentage error of a position depends on the origin of the coordinate
system, and the ratio to constant velocity is the scale-free measure used
instead.

\begin{table*}
\caption{Position-error metrics on the vessel-disjoint test windows (km). MAE: mean haversine distance between predicted and true position (the error reported throughout the paper); RMSE: root mean square of the same distance; median and P90 at three hours; ADE: mean over the 18 steps; FDE equals the three-hour MAE. The compact encoder predicts only the one-, two- and three-hour positions, so its ADE is not defined. Mean over seeds (seed standard deviation in small type). Token models and GATransformer are the reruns of this appendix; at every horizon their MAE is within 2.3\% (0.14~km) of the runs reported in the main text.}\label{tbl:r2metrics}
\centering\footnotesize\setlength{\tabcolsep}{4pt}
\begin{tabular}{@{}lrrrrrrr@{}}
\toprule
 & \multicolumn{2}{c}{One hour} & \multicolumn{4}{c}{Three hours} & \\
\cmidrule(lr){2-3}\cmidrule(lr){4-7}
Model & MAE & RMSE & MAE (FDE) & RMSE & Median & P90 & ADE \\
\midrule
\multicolumn{8}{@{}l}{\emph{Danish (DMA)}} \\
Constant velocity & 3.56 & 5.77 & 16.53 & 23.11 & 11.35 & 39.30 & 7.40 \\
TrAISformer & 2.25{\scriptsize$\pm$0.02} & 3.97{\scriptsize$\pm$0.06} & 8.49{\scriptsize$\pm$0.05} & 14.31{\scriptsize$\pm$0.14} & 4.15{\scriptsize$\pm$0.12} & 21.50{\scriptsize$\pm$0.20} & 4.08{\scriptsize$\pm$0.02} \\
AISFormer-inspired & 2.26{\scriptsize$\pm$0.04} & 4.12{\scriptsize$\pm$0.08} & 8.85{\scriptsize$\pm$0.14} & 14.75{\scriptsize$\pm$0.17} & 4.24{\scriptsize$\pm$0.22} & 22.21{\scriptsize$\pm$0.49} & 4.20{\scriptsize$\pm$0.07} \\
TrAISformer, rel. & 2.61{\scriptsize$\pm$0.03} & 4.49{\scriptsize$\pm$0.06} & 11.94{\scriptsize$\pm$0.02} & 18.66{\scriptsize$\pm$0.17} & 7.01{\scriptsize$\pm$0.08} & 28.50{\scriptsize$\pm$0.09} & 5.47{\scriptsize$\pm$0.02} \\
GATransformer & 2.32{\scriptsize$\pm$0.03} & 3.34{\scriptsize$\pm$0.04} & 7.72{\scriptsize$\pm$0.06} & 11.80{\scriptsize$\pm$0.15} & 5.25{\scriptsize$\pm$0.12} & 16.65{\scriptsize$\pm$0.28} & 3.84{\scriptsize$\pm$0.02} \\
LSTM on displacements & 2.43{\scriptsize$\pm$0.02} & 3.77{\scriptsize$\pm$0.03} & 8.51{\scriptsize$\pm$0.03} & 13.20{\scriptsize$\pm$0.06} & 5.43{\scriptsize$\pm$0.02} & 19.37{\scriptsize$\pm$0.17} & 4.12{\scriptsize$\pm$0.01} \\
Compact encoder & 2.82{\scriptsize$\pm$0.04} & 4.23{\scriptsize$\pm$0.07} & 9.41{\scriptsize$\pm$0.10} & 13.83{\scriptsize$\pm$0.14} & 6.50{\scriptsize$\pm$0.20} & 21.35{\scriptsize$\pm$0.15} & --- \\
\addlinespace
\multicolumn{8}{@{}l}{\emph{US (NOAA)}} \\
Constant velocity & 3.69 & 5.52 & 15.33 & 21.17 & 10.58 & 34.40 & 7.09 \\
TrAISformer & 2.16{\scriptsize$\pm$0.04} & 3.75{\scriptsize$\pm$0.06} & 8.47{\scriptsize$\pm$0.07} & 15.06{\scriptsize$\pm$0.15} & 3.42{\scriptsize$\pm$0.09} & 24.31{\scriptsize$\pm$0.45} & 4.02{\scriptsize$\pm$0.02} \\
AISFormer-inspired & 2.20{\scriptsize$\pm$0.02} & 3.90{\scriptsize$\pm$0.02} & 8.79{\scriptsize$\pm$0.08} & 15.55{\scriptsize$\pm$0.03} & 3.59{\scriptsize$\pm$0.05} & 24.87{\scriptsize$\pm$0.26} & 4.14{\scriptsize$\pm$0.04} \\
TrAISformer, rel. & 2.20{\scriptsize$\pm$0.02} & 3.84{\scriptsize$\pm$0.03} & 9.60{\scriptsize$\pm$0.15} & 16.31{\scriptsize$\pm$0.13} & 3.81{\scriptsize$\pm$0.14} & 27.57{\scriptsize$\pm$0.44} & 4.38{\scriptsize$\pm$0.07} \\
GATransformer & 2.49{\scriptsize$\pm$0.06} & 3.61{\scriptsize$\pm$0.04} & 8.08{\scriptsize$\pm$0.10} & 12.81{\scriptsize$\pm$0.21} & 4.52{\scriptsize$\pm$0.06} & 19.57{\scriptsize$\pm$0.24} & 4.11{\scriptsize$\pm$0.06} \\
LSTM on displacements & 2.16{\scriptsize$\pm$0.02} & 3.50{\scriptsize$\pm$0.03} & 7.76{\scriptsize$\pm$0.04} & 13.29{\scriptsize$\pm$0.11} & 3.65{\scriptsize$\pm$0.04} & 20.44{\scriptsize$\pm$0.29} & 3.76{\scriptsize$\pm$0.03} \\
Compact encoder & 2.81{\scriptsize$\pm$0.07} & 4.22{\scriptsize$\pm$0.09} & 9.75{\scriptsize$\pm$0.09} & 15.29{\scriptsize$\pm$0.10} & 5.32{\scriptsize$\pm$0.15} & 25.78{\scriptsize$\pm$0.29} & --- \\
\bottomrule
\end{tabular}
\end{table*}

\section*{Declaration of competing interest}

The authors declare that they have no known competing financial
interests or personal relationships that could have appeared to
influence the work reported in this paper.

\section*{Funding}

This research did not receive any specific grant from funding agencies in
the public, commercial, or not-for-profit sectors.

\section*{Data availability}

The raw AIS data underlying this study are third-party public feeds
published by the Danish Maritime Authority \citep{dma2026ais} and by NOAA's
MarineCadastre archive \citep{noaa_marinecadastre}, and were not collected
by the authors. The
preprocessing pipelines, all split manifests of both corpora
(Section~\ref{sec:splits}), and the implementations of every model in
Section~\ref{sec:method} and Appendix~\ref{app:comparators} are publicly available on Zenodo
\citep{raisi2026code}, together with the run configurations and result files.
Neither raw feed is redistributed; the archive holds only the split
manifests derived from them.

\section*{Declaration of generative AI and AI-assisted technologies in the manuscript preparation process}

During the preparation of this work, the authors used generative AI and
AI-assisted tools to assist with language editing, translation, grammar
correction, formatting, and readability improvement. The authors
reviewed and edited the content as needed and take full responsibility
for the content of the submitted manuscript. No generative AI tools were
used to generate scientific results, data, tables, figures, or
conclusions.

\printcredits

\bibliographystyle{cas-model2-names}
\bibliography{references}

\begin{thebibliography}{45}
\expandafter\ifx\csname natexlab\endcsname\relax\def\natexlab#1{#1}\fi
\providecommand{\url}[1]{\texttt{#1}}
\providecommand{\href}[2]{#2}
\providecommand{\path}[1]{#1}
\providecommand{\DOIprefix}{doi:}
\providecommand{\ArXivprefix}{arXiv:}
\providecommand{\URLprefix}{URL: }
\providecommand{\Pubmedprefix}{pmid:}
\providecommand{\doi}[1]{\href{http://dx.doi.org/#1}{\path{#1}}}
\providecommand{\Pubmed}[1]{\href{pmid:#1}{\path{#1}}}
\providecommand{\bibinfo}[2]{#2}
\ifx\xfnm\relax \def\xfnm[#1]{\unskip,\space#1}\fi
\bibitem[{Bouthillier et~al.(2021)Bouthillier, Delaunay, Bronzi, Trofimov,
  Nichyporuk, Szeto, {Mohammadi Sepahvand}, Raff, Madan, Voleti, {Ebrahimi
  Kahou}, Michalski, Arbel, Pal, Varoquaux and
  Vincent}]{bouthillier2021variance}
\bibinfo{author}{Bouthillier, X.}, \bibinfo{author}{Delaunay, P.},
  \bibinfo{author}{Bronzi, M.}, \bibinfo{author}{Trofimov, A.},
  \bibinfo{author}{Nichyporuk, B.}, \bibinfo{author}{Szeto, J.},
  \bibinfo{author}{{Mohammadi Sepahvand}, N.}, \bibinfo{author}{Raff, E.},
  \bibinfo{author}{Madan, K.}, \bibinfo{author}{Voleti, V.},
  \bibinfo{author}{{Ebrahimi Kahou}, S.}, \bibinfo{author}{Michalski, V.},
  \bibinfo{author}{Arbel, T.}, \bibinfo{author}{Pal, C.},
  \bibinfo{author}{Varoquaux, G.}, \bibinfo{author}{Vincent, P.},
  \bibinfo{year}{2021}.
\newblock \bibinfo{title}{Accounting for variance in machine learning
  benchmarks}, in: \bibinfo{booktitle}{Proceedings of Machine Learning and
  Systems (MLSys)}, pp. \bibinfo{pages}{747--769}.
\bibitem[{Cao et~al.(2018)Cao, Wang, Li, Zhou, Li and Li}]{cao2018brits}
\bibinfo{author}{Cao, W.}, \bibinfo{author}{Wang, D.}, \bibinfo{author}{Li,
  J.}, \bibinfo{author}{Zhou, H.}, \bibinfo{author}{Li, L.},
  \bibinfo{author}{Li, Y.}, \bibinfo{year}{2018}.
\newblock \bibinfo{title}{{BRITS}: bidirectional recurrent imputation for time
  series}, in: \bibinfo{booktitle}{Advances in Neural Information Processing
  Systems (NeurIPS) 31}.
\newblock \href{http://arxiv.org/abs/1805.10572}{\tt arXiv:1805.10572}.
\bibitem[{Capobianco et~al.(2021)Capobianco, Millefiori, Forti, Braca and
  Willett}]{capobianco2021rnn}
\bibinfo{author}{Capobianco, S.}, \bibinfo{author}{Millefiori, L.M.},
  \bibinfo{author}{Forti, N.}, \bibinfo{author}{Braca, P.},
  \bibinfo{author}{Willett, P.}, \bibinfo{year}{2021}.
\newblock \bibinfo{title}{Deep learning methods for vessel trajectory
  prediction based on recurrent neural networks}.
\newblock \bibinfo{journal}{IEEE Trans. Aerosp. Electron. Syst.}
  \bibinfo{volume}{57}, \bibinfo{pages}{4329--4346}.
\newblock \DOIprefix\doi{10.1109/TAES.2021.3096873}.
\bibitem[{Cerqueira et~al.(2020)Cerqueira, Torgo and Mozeti{\v
  c}}]{cerqueira2020evaluating}
\bibinfo{author}{Cerqueira, V.}, \bibinfo{author}{Torgo, L.},
  \bibinfo{author}{Mozeti{\v c}, I.}, \bibinfo{year}{2020}.
\newblock \bibinfo{title}{Evaluating time series forecasting models: an
  empirical study on performance estimation methods}.
\newblock \bibinfo{journal}{Mach. Learn.} \bibinfo{volume}{109},
  \bibinfo{pages}{1997--2028}.
\newblock \DOIprefix\doi{10.1007/s10994-020-05910-7}.
\bibitem[{Chang et~al.(2019)Chang, Lambert, Sangkloy, Singh, Bak, Hartnett,
  Wang, Carr, Lucey, Ramanan and Hays}]{chang2019argoverse}
\bibinfo{author}{Chang, M.F.}, \bibinfo{author}{Lambert, J.},
  \bibinfo{author}{Sangkloy, P.}, \bibinfo{author}{Singh, J.},
  \bibinfo{author}{Bak, S.}, \bibinfo{author}{Hartnett, A.},
  \bibinfo{author}{Wang, D.}, \bibinfo{author}{Carr, P.},
  \bibinfo{author}{Lucey, S.}, \bibinfo{author}{Ramanan, D.},
  \bibinfo{author}{Hays, J.}, \bibinfo{year}{2019}.
\newblock \bibinfo{title}{Argoverse: 3{D} tracking and forecasting with rich
  maps}, in: \bibinfo{booktitle}{2019 {IEEE/CVF} Conference on Computer Vision
  and Pattern Recognition ({CVPR})}, pp. \bibinfo{pages}{8740--8749}.
\newblock \DOIprefix\doi{10.1109/CVPR.2019.00895}.
\bibitem[{Che et~al.(2018)Che, Purushotham, Cho, Sontag and Liu}]{che2018grud}
\bibinfo{author}{Che, Z.}, \bibinfo{author}{Purushotham, S.},
  \bibinfo{author}{Cho, K.}, \bibinfo{author}{Sontag, D.},
  \bibinfo{author}{Liu, Y.}, \bibinfo{year}{2018}.
\newblock \bibinfo{title}{Recurrent neural networks for multivariate time
  series with missing values}.
\newblock \bibinfo{journal}{Sci. Rep.} \bibinfo{volume}{8},
  \bibinfo{pages}{6085}.
\newblock \DOIprefix\doi{10.1038/s41598-018-24271-9}.
\bibitem[{{Danish Maritime Authority}(2026)}]{dma2026ais}
\bibinfo{author}{{Danish Maritime Authority}}, \bibinfo{year}{2026}.
\newblock \bibinfo{title}{{AIS} data, {D}anish waters, {M}ay 2026 [dataset]}.
\newblock \bibinfo{howpublished}{Danish Maritime Authority}.
\newblock \URLprefix \url{http://aisdata.ais.dk/}. \bibinfo{note}{accessed 21
  September 2026}.
\bibitem[{Du et~al.(2023)Du, C{\^o}t{\'e} and Liu}]{du2023saits}
\bibinfo{author}{Du, W.}, \bibinfo{author}{C{\^o}t{\'e}, D.},
  \bibinfo{author}{Liu, Y.}, \bibinfo{year}{2023}.
\newblock \bibinfo{title}{{SAITS}: self-attention-based imputation for time
  series}.
\newblock \bibinfo{journal}{Expert Syst. Appl.} \bibinfo{volume}{219},
  \bibinfo{pages}{119619}.
\newblock \DOIprefix\doi{10.1016/j.eswa.2023.119619}.
\bibitem[{Fang and Ji(2026)}]{fang2026sde}
\bibinfo{author}{Fang, Y.}, \bibinfo{author}{Ji, M.}, \bibinfo{year}{2026}.
\newblock \bibinfo{title}{Physics-informed neural stochastic differential
  equations for probabilistic multi-step vessel trajectory prediction}.
\newblock \bibinfo{journal}{Ocean Eng.} \bibinfo{volume}{365},
  \bibinfo{pages}{127015}.
\newblock \DOIprefix\doi{10.1016/j.oceaneng.2026.127015}.
\bibitem[{Ferrari~Dacrema et~al.(2019)Ferrari~Dacrema, Cremonesi and
  Jannach}]{dacrema2019progress}
\bibinfo{author}{Ferrari~Dacrema, M.}, \bibinfo{author}{Cremonesi, P.},
  \bibinfo{author}{Jannach, D.}, \bibinfo{year}{2019}.
\newblock \bibinfo{title}{Are we really making much progress? a worrying
  analysis of recent neural recommendation approaches}, in:
  \bibinfo{booktitle}{Proceedings of the 13th {ACM} Conference on Recommender
  Systems}, pp. \bibinfo{pages}{101--109}.
\newblock \DOIprefix\doi{10.1145/3298689.3347058}.
\bibitem[{Guo and Ma(2026)}]{guo2026shiptrajreview}
\bibinfo{author}{Guo, S.}, \bibinfo{author}{Ma, W.}, \bibinfo{year}{2026}.
\newblock \bibinfo{title}{A systematic review of deep learning-based methods
  for ship trajectory prediction}.
\newblock \bibinfo{journal}{Journal of Marine Science and Engineering}
  \bibinfo{volume}{14}, \bibinfo{pages}{810}.
\newblock \DOIprefix\doi{10.3390/jmse14090810}.
\bibitem[{Guo et~al.(2026)Guo, Ren, Wei, Liu, Mei and Xu}]{mgformer2026}
\bibinfo{author}{Guo, X.}, \bibinfo{author}{Ren, Y.}, \bibinfo{author}{Wei,
  Y.}, \bibinfo{author}{Liu, J.}, \bibinfo{author}{Mei, Q.},
  \bibinfo{author}{Xu, M.}, \bibinfo{year}{2026}.
\newblock \bibinfo{title}{{MGFormer}: a graph-enhanced transformer framework
  for long-range gap imputation in {AIS} trajectories}.
\newblock \bibinfo{journal}{Ocean Eng.} \bibinfo{volume}{358},
  \bibinfo{pages}{125563}.
\newblock \DOIprefix\doi{10.1016/j.oceaneng.2026.125563}.
\bibitem[{Guo et~al.(2024)Guo, Wang, Ren, Liu, Wang, Chen, Zhang and
  Xu}]{guo2024segmentation}
\bibinfo{author}{Guo, X.}, \bibinfo{author}{Wang, N.}, \bibinfo{author}{Ren,
  Y.}, \bibinfo{author}{Liu, J.}, \bibinfo{author}{Wang, H.},
  \bibinfo{author}{Chen, X.}, \bibinfo{author}{Zhang, B.}, \bibinfo{author}{Xu,
  M.}, \bibinfo{year}{2024}.
\newblock \bibinfo{title}{Ship trajectory segmentation by movement states while
  addressing uncertainty and sparsity}.
\newblock \bibinfo{journal}{Ocean Eng.} \bibinfo{volume}{312},
  \bibinfo{pages}{119218}.
\newblock \DOIprefix\doi{10.1016/j.oceaneng.2024.119218}.
\bibitem[{Gupta et~al.(2018)Gupta, Johnson, Fei-Fei, Savarese and
  Alahi}]{gupta2018socialgan}
\bibinfo{author}{Gupta, A.}, \bibinfo{author}{Johnson, J.},
  \bibinfo{author}{Fei-Fei, L.}, \bibinfo{author}{Savarese, S.},
  \bibinfo{author}{Alahi, A.}, \bibinfo{year}{2018}.
\newblock \bibinfo{title}{Social {GAN}: socially acceptable trajectories with
  generative adversarial networks}, in: \bibinfo{booktitle}{2018 {IEEE/CVF}
  Conference on Computer Vision and Pattern Recognition ({CVPR})}, pp.
  \bibinfo{pages}{2255--2264}.
\newblock \DOIprefix\doi{10.1109/CVPR.2018.00240}.
\bibitem[{Henderson et~al.(2018)Henderson, Islam, Bachman, Pineau, Precup and
  Meger}]{henderson2018drl}
\bibinfo{author}{Henderson, P.}, \bibinfo{author}{Islam, R.},
  \bibinfo{author}{Bachman, P.}, \bibinfo{author}{Pineau, J.},
  \bibinfo{author}{Precup, D.}, \bibinfo{author}{Meger, D.},
  \bibinfo{year}{2018}.
\newblock \bibinfo{title}{Deep reinforcement learning that matters}, in:
  \bibinfo{booktitle}{Proceedings of the {AAAI} Conference on Artificial
  Intelligence}.
\newblock \DOIprefix\doi{10.1609/aaai.v32i1.11694}.
\bibitem[{Jin et~al.(2025)Jin, Zhang and Tang}]{jin2025stgdpm}
\bibinfo{author}{Jin, W.}, \bibinfo{author}{Zhang, X.}, \bibinfo{author}{Tang,
  H.}, \bibinfo{year}{2025}.
\newblock \bibinfo{title}{{STGDPM}: Vessel trajectory prediction with
  spatio-temporal graph diffusion probabilistic model}, in:
  \bibinfo{booktitle}{Database Systems for Advanced Applications (DASFAA)}, pp.
  \bibinfo{pages}{571--586}.
\newblock \DOIprefix\doi{10.1007/978-981-95-3830-0_43}.
\bibitem[{Kapoor and Narayanan(2023)}]{kapoor2023leakage}
\bibinfo{author}{Kapoor, S.}, \bibinfo{author}{Narayanan, A.},
  \bibinfo{year}{2023}.
\newblock \bibinfo{title}{Leakage and the reproducibility crisis in
  machine-learning-based science}.
\newblock \bibinfo{journal}{Patterns} \bibinfo{volume}{4},
  \bibinfo{pages}{100804}.
\newblock \DOIprefix\doi{10.1016/j.patter.2023.100804}.
\bibitem[{Kaufman et~al.(2011)Kaufman, Rosset and Perlich}]{kaufman2011leakage}
\bibinfo{author}{Kaufman, S.}, \bibinfo{author}{Rosset, S.},
  \bibinfo{author}{Perlich, C.}, \bibinfo{year}{2011}.
\newblock \bibinfo{title}{Leakage in data mining: formulation, detection, and
  avoidance}, in: \bibinfo{booktitle}{Proceedings of the 17th {ACM} {SIGKDD}
  International Conference on Knowledge Discovery and Data Mining}, pp.
  \bibinfo{pages}{556--563}.
\newblock \DOIprefix\doi{10.1145/2020408.2020496}.
\bibitem[{Li et~al.(2024)Li, Gan, Lan, Cai, Liu, Lin and Liu}]{li2024diffutraj}
\bibinfo{author}{Li, C.}, \bibinfo{author}{Gan, Y.}, \bibinfo{author}{Lan, T.},
  \bibinfo{author}{Cai, Y.}, \bibinfo{author}{Liu, X.}, \bibinfo{author}{Lin,
  R.}, \bibinfo{author}{Liu, Q.}, \bibinfo{year}{2024}.
\newblock \bibinfo{title}{{DiffuTraj}: a stochastic vessel trajectory
  prediction approach via guided diffusion process}.
\newblock \bibinfo{journal}{arXiv preprint arXiv:2410.09550} .
\bibitem[{Li et~al.(2023)Li, Jiao and Yang}]{li2023shiptrajreview}
\bibinfo{author}{Li, H.}, \bibinfo{author}{Jiao, H.}, \bibinfo{author}{Yang,
  Z.}, \bibinfo{year}{2023}.
\newblock \bibinfo{title}{Ship trajectory prediction based on machine learning
  and deep learning: a systematic review and methods analysis}.
\newblock \bibinfo{journal}{Engineering Applications of Artificial
  Intelligence} \bibinfo{volume}{126}, \bibinfo{pages}{107062}.
\newblock \DOIprefix\doi{10.1016/j.engappai.2023.107062}.
\bibitem[{Liang et~al.(2022)Liang, Liu, Zhan, Li, Zhu and
  Wang}]{liang2022stmgcnflow}
\bibinfo{author}{Liang, M.}, \bibinfo{author}{Liu, R.W.},
  \bibinfo{author}{Zhan, Y.}, \bibinfo{author}{Li, H.}, \bibinfo{author}{Zhu,
  F.}, \bibinfo{author}{Wang, F.Y.}, \bibinfo{year}{2022}.
\newblock \bibinfo{title}{Fine-grained vessel traffic flow prediction with a
  spatio-temporal multigraph convolutional network}.
\newblock \bibinfo{journal}{IEEE Transactions on Intelligent Transportation
  Systems} \bibinfo{volume}{23}, \bibinfo{pages}{23694--23707}.
\newblock \DOIprefix\doi{10.1109/TITS.2022.3199160}.
\bibitem[{Liang et~al.(2024)Liang, Su, Gao, Liu and Zhan}]{liang2024traveltime}
\bibinfo{author}{Liang, M.}, \bibinfo{author}{Su, J.}, \bibinfo{author}{Gao,
  R.}, \bibinfo{author}{Liu, R.W.}, \bibinfo{author}{Zhan, Y.},
  \bibinfo{year}{2024}.
\newblock \bibinfo{title}{Estimation of vessel link-level travel time
  distribution: a directed network-driven approach}.
\newblock \bibinfo{journal}{Ocean Engineering} \bibinfo{volume}{313},
  \bibinfo{pages}{119371}.
\newblock \DOIprefix\doi{10.1016/j.oceaneng.2024.119371}.
\bibitem[{Liang et~al.(2021)Liang, Zhan and Liu}]{liang2021mvffnet}
\bibinfo{author}{Liang, M.}, \bibinfo{author}{Zhan, Y.}, \bibinfo{author}{Liu,
  R.W.}, \bibinfo{year}{2021}.
\newblock \bibinfo{title}{{MVFFNet}: multi-view feature fusion network for
  imbalanced ship classification}.
\newblock \bibinfo{journal}{Pattern Recognition Letters} \bibinfo{volume}{151},
  \bibinfo{pages}{26--32}.
\newblock \DOIprefix\doi{10.1016/j.patrec.2021.07.024}.
\bibitem[{Liu et~al.(2022)Liu, Liang, Nie, Yuan, Xiong, Yu and
  Guizani}]{liu2022stmgcntraj}
\bibinfo{author}{Liu, R.W.}, \bibinfo{author}{Liang, M.}, \bibinfo{author}{Nie,
  J.}, \bibinfo{author}{Yuan, Y.}, \bibinfo{author}{Xiong, Z.},
  \bibinfo{author}{Yu, H.}, \bibinfo{author}{Guizani, N.},
  \bibinfo{year}{2022}.
\newblock \bibinfo{title}{{STMGCN}: mobile edge computing-empowered vessel
  trajectory prediction using spatio-temporal multigraph convolutional
  network}.
\newblock \bibinfo{journal}{IEEE Transactions on Industrial Informatics}
  \bibinfo{volume}{18}, \bibinfo{pages}{7977--7987}.
\newblock \DOIprefix\doi{10.1109/TII.2022.3165886}.
\bibitem[{Lones(2024)}]{lones2024pitfalls}
\bibinfo{author}{Lones, M.A.}, \bibinfo{year}{2024}.
\newblock \bibinfo{title}{Avoiding common machine learning pitfalls}.
\newblock \bibinfo{journal}{Patterns} \bibinfo{volume}{5},
  \bibinfo{pages}{101046}.
\newblock \DOIprefix\doi{10.1016/j.patter.2024.101046}.
\bibitem[{Ma et~al.(2026)Ma, Han, Song, Yao, Wang and Wu}]{ma2026envship}
\bibinfo{author}{Ma, K.}, \bibinfo{author}{Han, Q.}, \bibinfo{author}{Song,
  C.}, \bibinfo{author}{Yao, J.}, \bibinfo{author}{Wang, H.},
  \bibinfo{author}{Wu, C.}, \bibinfo{year}{2026}.
\newblock \bibinfo{title}{{EnvShip}: a unified framework for context-aware and
  cross-region vessel trajectory forecasting}.
\newblock \bibinfo{journal}{arXiv preprint arXiv:2606.15240} .
\bibitem[{Melis et~al.(2018)Melis, Dyer and Blunsom}]{melis2018evaluation}
\bibinfo{author}{Melis, G.}, \bibinfo{author}{Dyer, C.},
  \bibinfo{author}{Blunsom, P.}, \bibinfo{year}{2018}.
\newblock \bibinfo{title}{On the state of the art of evaluation in neural
  language models}, in: \bibinfo{booktitle}{International Conference on
  Learning Representations (ICLR)}.
\newblock \href{http://arxiv.org/abs/1707.05589}{\tt arXiv:1707.05589}.
\bibitem[{Musgrave et~al.(2020)Musgrave, Belongie and Lim}]{musgrave2020metric}
\bibinfo{author}{Musgrave, K.}, \bibinfo{author}{Belongie, S.},
  \bibinfo{author}{Lim, S.N.}, \bibinfo{year}{2020}.
\newblock \bibinfo{title}{A metric learning reality check}, in:
  \bibinfo{booktitle}{Computer Vision -- {ECCV} 2020}, pp.
  \bibinfo{pages}{681--699}.
\newblock \DOIprefix\doi{10.1007/978-3-030-58595-2_41}.
\bibitem[{{National Oceanic and Atmospheric
  Administration}(2023)}]{noaa_marinecadastre}
\bibinfo{author}{{National Oceanic and Atmospheric Administration}},
  \bibinfo{year}{2023}.
\newblock \bibinfo{title}{{AIS} broadcast data, {M}arine{C}adastre.gov, {J}une
  2023 [dataset]}.
\newblock \bibinfo{howpublished}{NOAA Office for Coastal Management}.
\newblock \URLprefix \url{https://marinecadastre.gov/ais/}.
  \bibinfo{note}{accessed 21 September 2026}.
\bibitem[{Nguyen and Fablet(2024)}]{nguyen2021traisformer}
\bibinfo{author}{Nguyen, D.}, \bibinfo{author}{Fablet, R.},
  \bibinfo{year}{2024}.
\newblock \bibinfo{title}{A transformer network with sparse augmented data
  representation and cross entropy loss for {AIS}-based vessel trajectory
  prediction}.
\newblock \bibinfo{journal}{IEEE Access} \bibinfo{volume}{12},
  \bibinfo{pages}{21596--21609}.
\newblock \DOIprefix\doi{10.1109/ACCESS.2024.3349957}.
\bibitem[{Nie et~al.(2023)Nie, Nguyen, Sinthong and
  Kalagnanam}]{nie2023patchtst}
\bibinfo{author}{Nie, Y.}, \bibinfo{author}{Nguyen, N.H.},
  \bibinfo{author}{Sinthong, P.}, \bibinfo{author}{Kalagnanam, J.},
  \bibinfo{year}{2023}.
\newblock \bibinfo{title}{A time series is worth 64 words: long-term
  forecasting with transformers}, in: \bibinfo{booktitle}{International
  Conference on Learning Representations (ICLR)}.
\newblock \href{http://arxiv.org/abs/2211.14730}{\tt arXiv:2211.14730}.
\bibitem[{Oliver et~al.(2018)Oliver, Odena, Raffel, Cubuk and
  Goodfellow}]{oliver2018realistic}
\bibinfo{author}{Oliver, A.}, \bibinfo{author}{Odena, A.},
  \bibinfo{author}{Raffel, C.}, \bibinfo{author}{Cubuk, E.D.},
  \bibinfo{author}{Goodfellow, I.J.}, \bibinfo{year}{2018}.
\newblock \bibinfo{title}{Realistic evaluation of deep semi-supervised learning
  algorithms}, in: \bibinfo{booktitle}{Advances in Neural Information
  Processing Systems (NeurIPS) 31}.
\newblock \href{http://arxiv.org/abs/1804.09170}{\tt arXiv:1804.09170}.
\bibitem[{Park et~al.(2026)Park, Jung, Seo, Choi, Cho, Park and
  Choi}]{park2025aisllm}
\bibinfo{author}{Park, H.}, \bibinfo{author}{Jung, J.}, \bibinfo{author}{Seo,
  M.}, \bibinfo{author}{Choi, H.}, \bibinfo{author}{Cho, D.},
  \bibinfo{author}{Park, S.}, \bibinfo{author}{Choi, D.G.},
  \bibinfo{year}{2026}.
\newblock \bibinfo{title}{{AIS-LLM}: a unified framework for maritime
  trajectory prediction, anomaly detection, and collision risk assessment with
  explainable forecasting}.
\newblock \bibinfo{journal}{IEEE Access} \bibinfo{volume}{14},
  \bibinfo{pages}{106874--106891}.
\newblock \DOIprefix\doi{10.1109/ACCESS.2026.3705215}.
\bibitem[{Raisi and Nazarzehi~Had(2026)}]{raisi2026code}
\bibinfo{author}{Raisi, Z.}, \bibinfo{author}{Nazarzehi~Had, V.M.},
  \bibinfo{year}{2026}.
\newblock \bibinfo{title}{Code and split manifests for ``{P}rotocol before
  progress: leakage-aware evaluation of {AIS}-based vessel trajectory
  prediction'' [software]}.
\newblock \bibinfo{howpublished}{Zenodo}.
\newblock \DOIprefix\doi{10.5281/zenodo.22868234}. \bibinfo{note}{concept DOI,
  resolves to the latest version. Version used in this paper: v1.4.0 (November
  2026), doi:10.5281/zenodo.24345380, with both corpora and all revision
  experiments}.
\bibitem[{Riveiro et~al.(2018)Riveiro, Pallotta and
  Vespe}]{riveiro2018maritime}
\bibinfo{author}{Riveiro, M.}, \bibinfo{author}{Pallotta, G.},
  \bibinfo{author}{Vespe, M.}, \bibinfo{year}{2018}.
\newblock \bibinfo{title}{Maritime anomaly detection: a review}.
\newblock \bibinfo{journal}{WIREs Data Min. Knowl. Discov.}
  \bibinfo{volume}{8}, \bibinfo{pages}{e1266}.
\newblock \DOIprefix\doi{10.1002/widm.1266}.
\bibitem[{Roberts et~al.(2017)Roberts, Bahn, Ciuti, Boyce, Elith,
  Guillera-Arroita, Hauenstein, Lahoz-Monfort, Schr{\"o}der, Thuiller, Warton,
  Wintle, Hartig and Dormann}]{roberts2017blockcv}
\bibinfo{author}{Roberts, D.R.}, \bibinfo{author}{Bahn, V.},
  \bibinfo{author}{Ciuti, S.}, \bibinfo{author}{Boyce, M.S.},
  \bibinfo{author}{Elith, J.}, \bibinfo{author}{Guillera-Arroita, G.},
  \bibinfo{author}{Hauenstein, S.}, \bibinfo{author}{Lahoz-Monfort, J.J.},
  \bibinfo{author}{Schr{\"o}der, B.}, \bibinfo{author}{Thuiller, W.},
  \bibinfo{author}{Warton, D.I.}, \bibinfo{author}{Wintle, B.A.},
  \bibinfo{author}{Hartig, F.}, \bibinfo{author}{Dormann, C.F.},
  \bibinfo{year}{2017}.
\newblock \bibinfo{title}{Cross-validation strategies for data with temporal,
  spatial, hierarchical, or phylogenetic structure}.
\newblock \bibinfo{journal}{Ecography} \bibinfo{volume}{40},
  \bibinfo{pages}{913--929}.
\newblock \DOIprefix\doi{10.1111/ecog.02881}.
\bibitem[{Salzmann et~al.(2020)Salzmann, Ivanovic, Chakravarty and
  Pavone}]{salzmann2020trajectron}
\bibinfo{author}{Salzmann, T.}, \bibinfo{author}{Ivanovic, B.},
  \bibinfo{author}{Chakravarty, P.}, \bibinfo{author}{Pavone, M.},
  \bibinfo{year}{2020}.
\newblock \bibinfo{title}{{Trajectron++}: dynamically-feasible trajectory
  forecasting with heterogeneous data}, in: \bibinfo{booktitle}{Computer Vision
  -- {ECCV} 2020}, pp. \bibinfo{pages}{683--700}.
\newblock \DOIprefix\doi{10.1007/978-3-030-58523-5_40}.
\bibitem[{Tashiro et~al.(2021)Tashiro, Song, Song and Ermon}]{tashiro2021csdi}
\bibinfo{author}{Tashiro, Y.}, \bibinfo{author}{Song, J.},
  \bibinfo{author}{Song, Y.}, \bibinfo{author}{Ermon, S.},
  \bibinfo{year}{2021}.
\newblock \bibinfo{title}{{CSDI}: conditional score-based diffusion models for
  probabilistic time series imputation}, in: \bibinfo{booktitle}{Advances in
  Neural Information Processing Systems (NeurIPS) 34}.
\newblock \href{http://arxiv.org/abs/2107.03502}{\tt arXiv:2107.03502}.
\bibitem[{Thiede and Brahma(2019)}]{thiede2019variety}
\bibinfo{author}{Thiede, L.A.}, \bibinfo{author}{Brahma, P.P.},
  \bibinfo{year}{2019}.
\newblock \bibinfo{title}{Analyzing the variety loss in the context of
  probabilistic trajectory prediction}, in: \bibinfo{booktitle}{2019 {IEEE/CVF}
  International Conference on Computer Vision ({ICCV})}, pp.
  \bibinfo{pages}{9953--9962}.
\newblock \DOIprefix\doi{10.1109/ICCV.2019.01005}.
\bibitem[{Xie et~al.(2025)Xie, Tu, Fu, Yuan and Han}]{xie2025survey}
\bibinfo{author}{Xie, Z.}, \bibinfo{author}{Tu, E.}, \bibinfo{author}{Fu, X.},
  \bibinfo{author}{Yuan, G.}, \bibinfo{author}{Han, Y.}, \bibinfo{year}{2025}.
\newblock \bibinfo{title}{{AIS} data-driven maritime monitoring based on
  transformer: a comprehensive review}, in: \bibinfo{booktitle}{2025
  International Joint Conference on Neural Networks (IJCNN)}, pp.
  \bibinfo{pages}{1--8}.
\newblock \DOIprefix\doi{10.1109/IJCNN64981.2025.11228006}.
\bibitem[{Yu et~al.(2025)Yu, Yin, Geng, Chen and Yang}]{yu2025aisformer}
\bibinfo{author}{Yu, Q.}, \bibinfo{author}{Yin, X.}, \bibinfo{author}{Geng,
  X.}, \bibinfo{author}{Chen, S.}, \bibinfo{author}{Yang, J.},
  \bibinfo{year}{2025}.
\newblock \bibinfo{title}{{AISFormer} for long-term vessel trajectory
  prediction}.
\newblock \bibinfo{journal}{Ocean Eng.} \bibinfo{volume}{340},
  \bibinfo{pages}{122098}.
\newblock \DOIprefix\doi{10.1016/j.oceaneng.2025.122098}.
\bibitem[{Yuan et~al.(2025)Yuan, Liu, Wu, Yu, Xin and
  Wang}]{yuan2025gatransformer}
\bibinfo{author}{Yuan, H.}, \bibinfo{author}{Liu, K.}, \bibinfo{author}{Wu,
  X.}, \bibinfo{author}{Yu, Y.}, \bibinfo{author}{Xin, X.},
  \bibinfo{author}{Wang, W.}, \bibinfo{year}{2025}.
\newblock \bibinfo{title}{{GATransformer}: a vessel trajectory prediction
  method based on attention algorithm in complex navigable waters}.
\newblock \bibinfo{journal}{Ocean Eng.} \bibinfo{volume}{326},
  \bibinfo{pages}{120902}.
\newblock \DOIprefix\doi{10.1016/j.oceaneng.2025.120902}.
\bibitem[{Zeng et~al.(2023)Zeng, Chen, Zhang and Xu}]{zeng2023dlinear}
\bibinfo{author}{Zeng, A.}, \bibinfo{author}{Chen, M.}, \bibinfo{author}{Zhang,
  L.}, \bibinfo{author}{Xu, Q.}, \bibinfo{year}{2023}.
\newblock \bibinfo{title}{Are transformers effective for time series
  forecasting?}, in: \bibinfo{booktitle}{Proceedings of the {AAAI} Conference
  on Artificial Intelligence}, pp. \bibinfo{pages}{11121--11128}.
\newblock \DOIprefix\doi{10.1609/aaai.v37i9.26317}.
\bibitem[{Zhan et~al.(2026)Zhan, Li, Zhang, Lu and Li}]{zhan2026shiptrajr1}
\bibinfo{author}{Zhan, Y.}, \bibinfo{author}{Li, Y.}, \bibinfo{author}{Zhang,
  C.}, \bibinfo{author}{Lu, Y.}, \bibinfo{author}{Li, Y.},
  \bibinfo{year}{2026}.
\newblock \bibinfo{title}{{ShipTraj-R1}: reinforcing ship trajectory prediction
  in large language models via group relative policy optimization}, in:
  \bibinfo{booktitle}{Advances in Knowledge Discovery and Data Mining (PAKDD
  2026)}, pp. \bibinfo{pages}{56--68}.
\newblock \DOIprefix\doi{10.1007/978-981-92-1300-9_5}.
\bibitem[{Zhou et~al.(2022)Zhou, Ma, Wen, Wang, Sun and
  Jin}]{zhou2022fedformer}
\bibinfo{author}{Zhou, T.}, \bibinfo{author}{Ma, Z.}, \bibinfo{author}{Wen,
  Q.}, \bibinfo{author}{Wang, X.}, \bibinfo{author}{Sun, L.},
  \bibinfo{author}{Jin, R.}, \bibinfo{year}{2022}.
\newblock \bibinfo{title}{{FEDformer}: frequency enhanced decomposed
  transformer for long-term series forecasting}, in:
  \bibinfo{booktitle}{Proceedings of the 39th International Conference on
  Machine Learning (ICML)}.
\newblock \href{http://arxiv.org/abs/2201.12740}{\tt arXiv:2201.12740}.

\end{thebibliography}

\end{document}